\documentclass[letterpaper]{article} %
\usepackage{aaai23}  %
\usepackage{times}  %
\usepackage{helvet}  %
\usepackage{courier}  %
\usepackage[hyphens]{url}  %
\usepackage{graphicx} %
\usepackage{natbib}  %
\usepackage{caption} %
\usepackage{algorithm}
\usepackage{algorithmic}
\usepackage{xcolor}
\newcommand{\ryn}[1]{\textcolor{blue}{#1}}
\newcommand{\rynq}[1]{\textcolor{red}{#1}}
\newcommand{\qhrewrite}[1]{\textcolor{blue}{#1}}

\renewcommand{\ryn}[1]{{#1}}
\renewcommand{\rynq}[1]{{#1}}
\renewcommand{\qhrewrite}[1]{{#1}}

\usepackage{newfloat}
\usepackage{listings}
\DeclareCaptionStyle{ruled}{labelfont=normalfont,labelsep=colon,strut=off} %
\floatstyle{ruled}
\newfloat{listing}{tb}{lst}{}
\floatname{listing}{Listing}
\title{Weakly-Supervised Camouflaged Object Detection with Scribble Annotations}
\author{
    Ruozhen He\equalcontrib,
    Qihua Dong\equalcontrib,
    Jiaying Lin\thanks{Corresponding authors: Jiaying Lin and Rynson W.H. Lau},
    Rynson W.H. Lau$^\dagger$
}
\affiliations{
    Department of Computer Science, City University of Hong Kong\\
    \{ruozhenhe2-c, qihuadong2-c, jiayinlin5-c\}@my.cityu.edu.hk,
    Rynson.Lau@cityu.edu.hk

}

\usepackage{bibentry}
\usepackage{amsmath}
\usepackage{amsfonts}

\newcommand{\qihua}[1]{\textcolor{cyan}{#1}}

\renewcommand{\qihua}{}

\begin{document}

\maketitle
\begin{abstract}
Existing camouflaged object detection (COD) methods rely heavily on large-scale datasets with pixel-wise annotations. However, due to the ambiguous boundary, annotating camouflage objects pixel-wisely is very time-consuming and labor-intensive, taking $\sim$60mins to label one image.
In this paper, we propose the first weakly-supervised COD method, using scribble annotations as supervision. To achieve this, we first relabel 4,040 images in existing camouflaged object datasets with scribbles, which takes $\sim$10s to label one image. 
As scribble annotations only describe the primary structure of objects without details, for the network to learn to localize the boundaries of camouflaged objects,
we propose a novel consistency loss composed of two parts: a cross-view loss to attain reliable consistency over different images, and an inside-view loss to maintain consistency inside a single prediction map. Besides, we observe that humans use semantic information to segment regions near the boundaries of camouflaged objects. Hence, we further propose a feature-guided loss, which includes visual features directly extracted from images and semantically significant features captured by the model. Finally, we propose a novel network for COD via scribble learning on structural information and semantic relations. Our network has two novel modules: the \textit{local-context contrasted (LCC) module}, which mimics visual inhibition to enhance image contrast/sharpness and expand the scribbles into potential camouflaged regions, and the \textit{logical semantic relation (LSR) module}, which analyzes the semantic relation to determine the regions representing the camouflaged object. Experimental results show that our model outperforms relevant SOTA methods on three COD benchmarks with an average improvement of 11.0$\%$ on MAE, 3.2$\%$ on S-measure, 2.5$\%$ on E-measure, and 4.4$\%$ on weighted F-measure\footnote{The code and dataset are available at \url{https://github.com/dddraxxx/Weakly-Supervised-Camouflaged-Object-Detection-with-Scribble-Annotations}.}. 
\end{abstract}

\section{Introduction}
Camouflaged object detection (COD) aims to detect visually inconspicuous objects in their surroundings, which includes natural objects with protective coloring, small sizes, occlusion, and artificial objects with information hiding purposes. Ambiguous boundaries between objects and backgrounds make it a more challenging task than other object detection tasks. Drawing increasing attention from the computer vision community, COD have numerous promising applications, including species discovery \cite{perez2012early}, medical image segmentation (like polyp segmentation with indistinguishable lesions) \cite{fan2020inf, fan2020pranet}, and animal-search \cite{fan2021concealed}.

\begin{figure}[t] 
    \centering
    \includegraphics[width=0.15\textwidth]{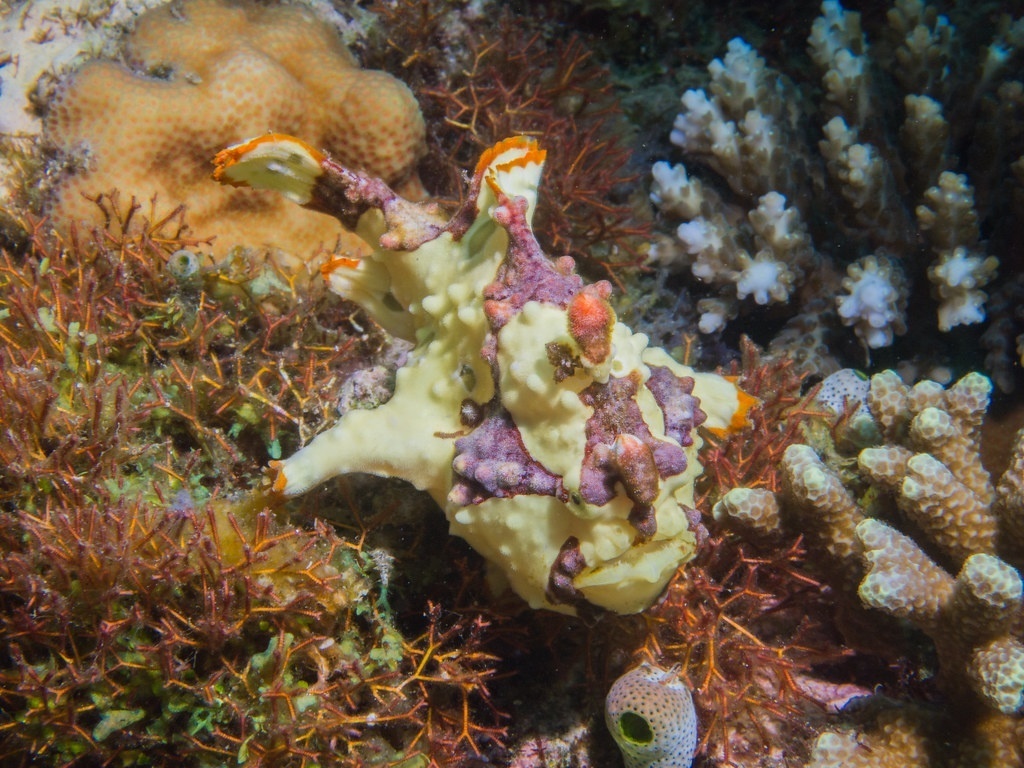}
    \includegraphics[width=0.15\textwidth]{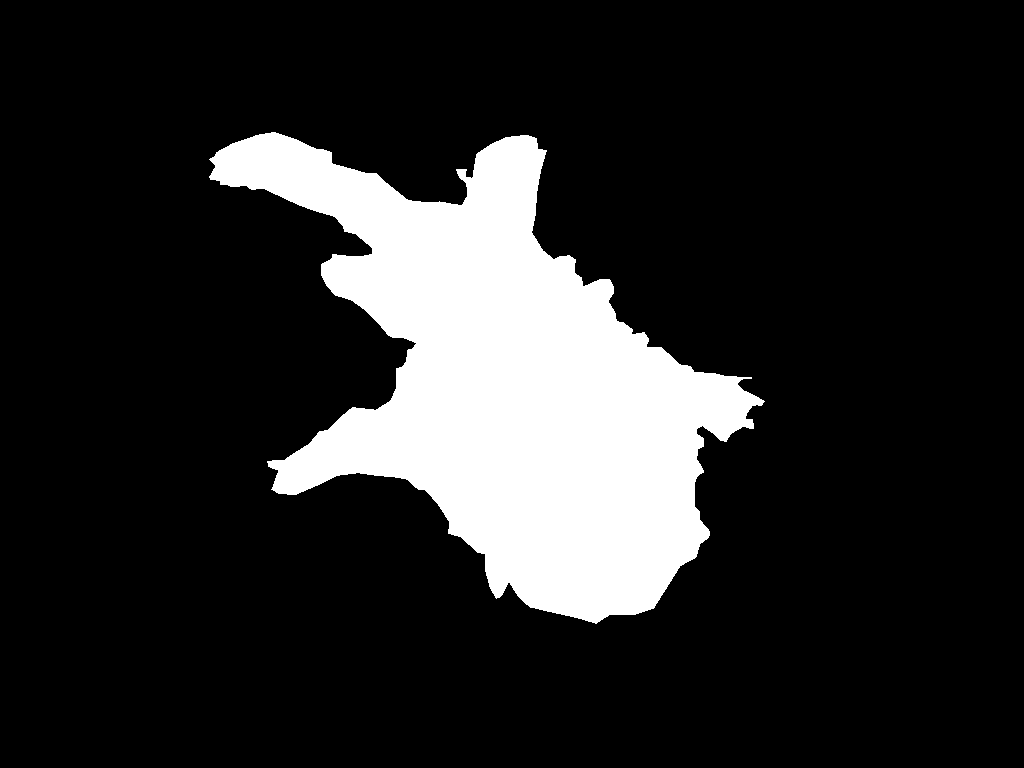}
    \includegraphics[width=0.15\textwidth]{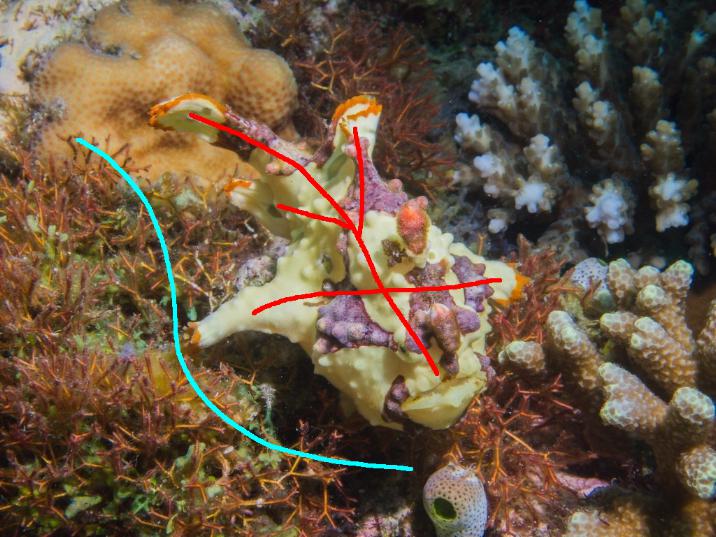}
    \\
    \makebox[0.15\textwidth]{\scriptsize Input}
    \makebox[0.15\textwidth]{\scriptsize GT (pixel-wise)}
    \makebox[0.15\textwidth]{\scriptsize Scribble}
    \\
    \includegraphics[width=0.15\textwidth]{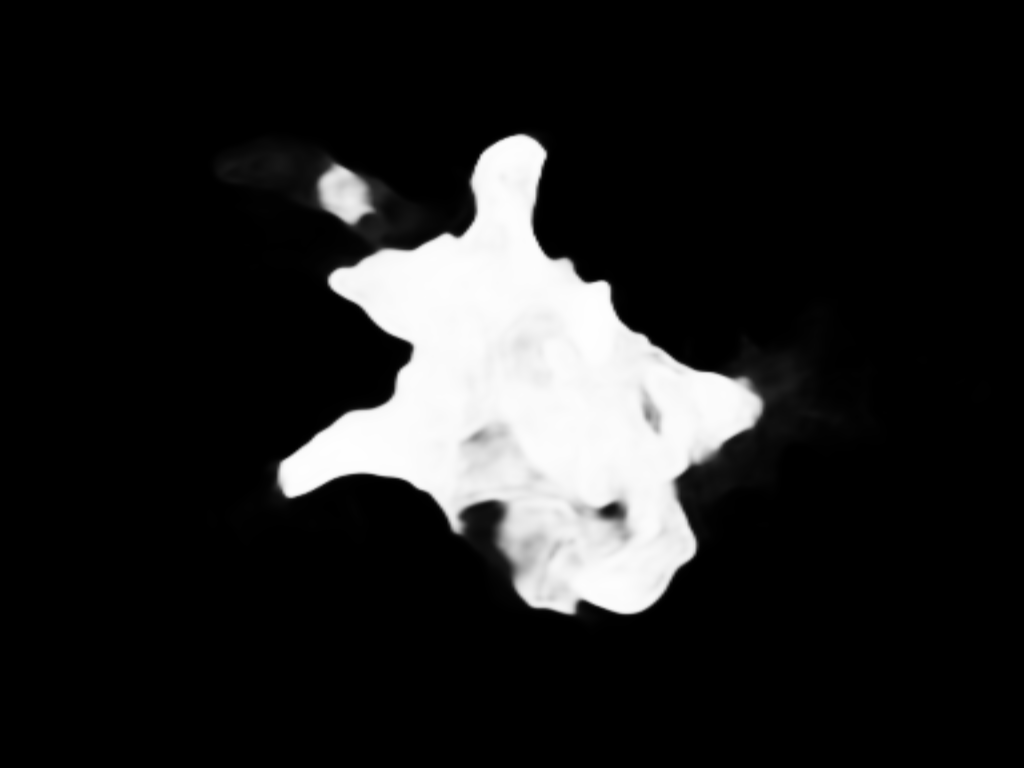}
    \includegraphics[width=0.15\textwidth]{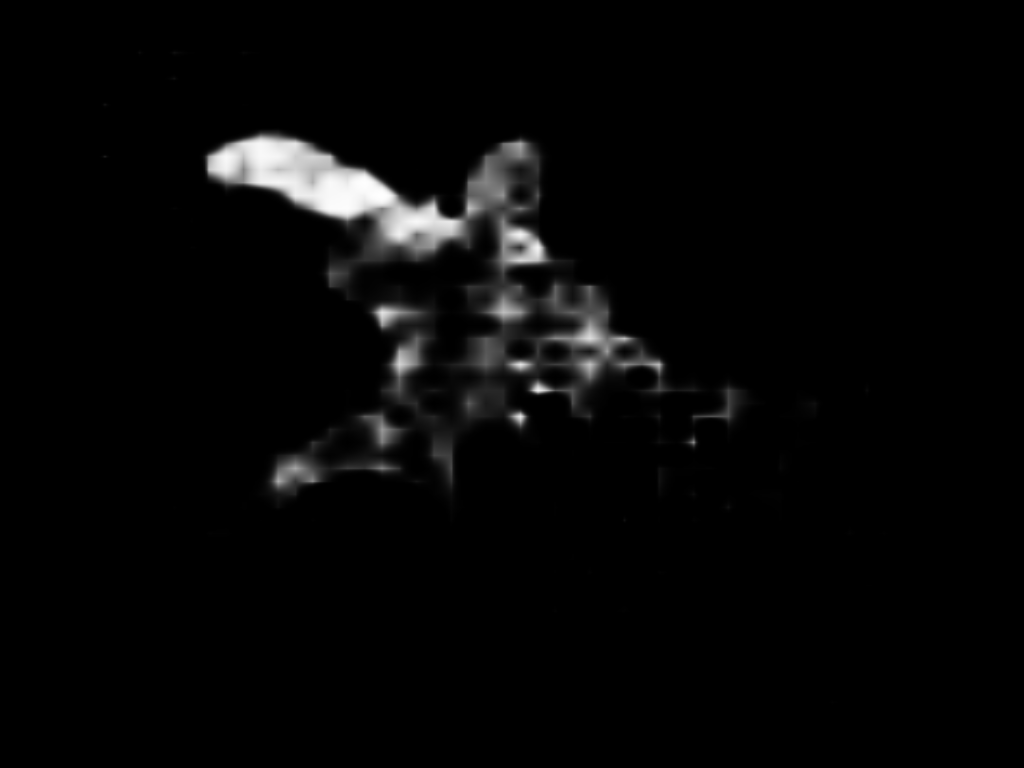}
    \includegraphics[width=0.15\textwidth]{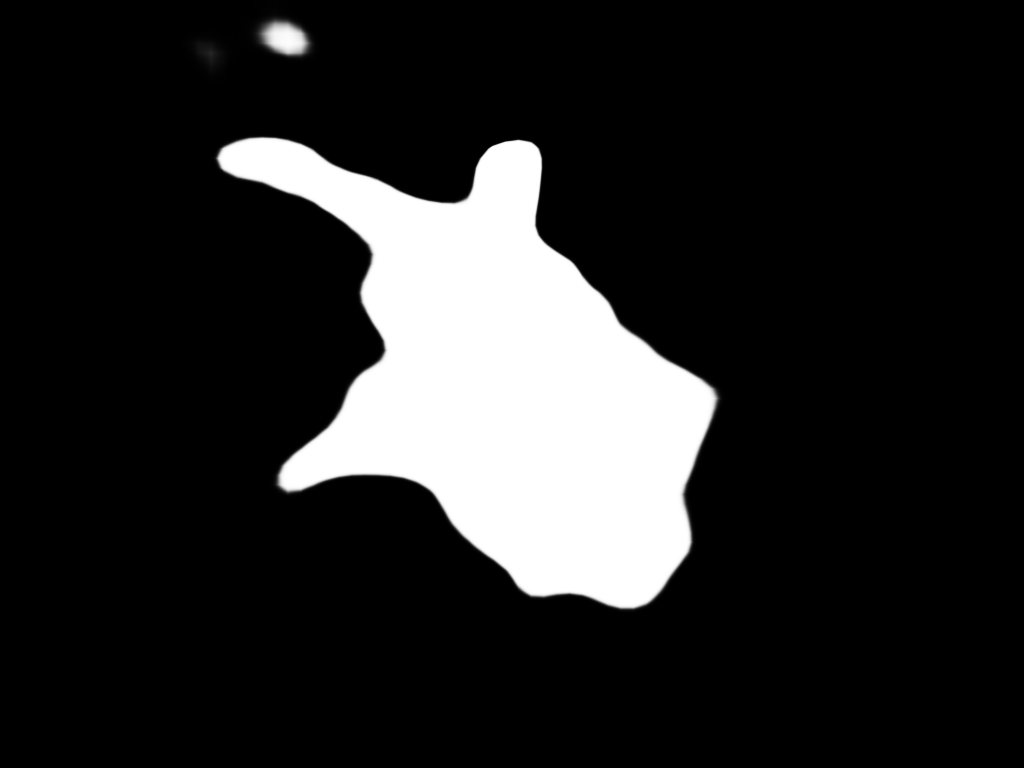}
    \\
    \makebox[0.15\textwidth]{\scriptsize ZoomNet}
    \makebox[0.15\textwidth]{\scriptsize UGTR}
    \makebox[0.15\textwidth]{\scriptsize Ours}
    \\

    \caption{\ryn{Scribbles only indicate} the primary structure of objects (cyan for background, red for foreground). Our method exploits this property to effectively learn rich semantic and structural information from the sparse labels. In some cases, it performs even better than fully-supervised models~\cite{youwei2022zoom,yang2021uncertainty}.}
    \label{fig:teaser1}
    \vspace{-4mm}
\end{figure}

\begin{figure}[t]
    \centering
    \includegraphics[width=0.4\textwidth]{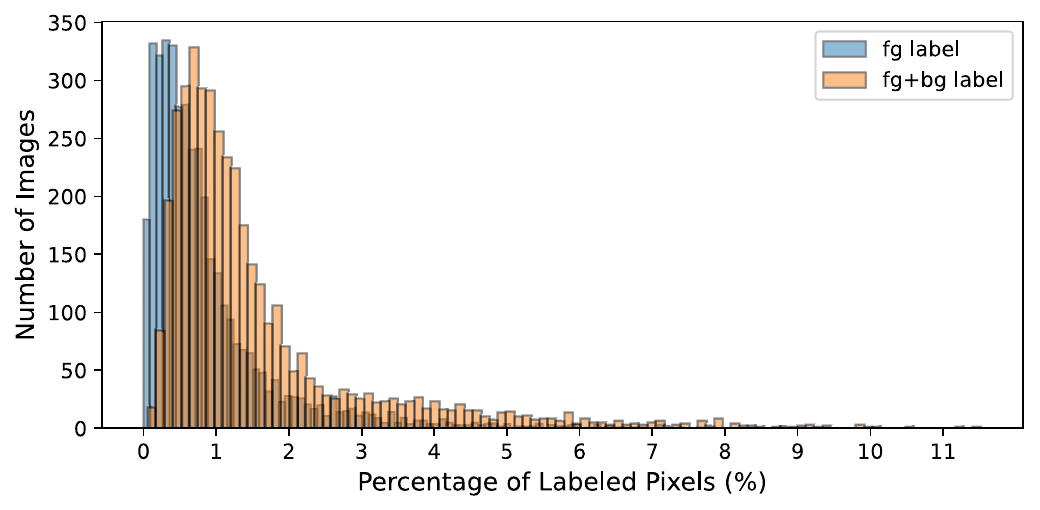}
    \caption{Percentage of labeled pixels in the S-COD dataset.}
    \label{fig:percentage}
\end{figure}

Although COD methods have already achieved excellent performances, they rely heavily on pixel-wise annotations of large-scale datasets.
There are two main weaknesses of pixel-wise annotations. First, they are time-consuming. It takes $\sim$60 minutes to annotate one image \cite{fan2020camouflaged}, which makes it very laborious to construct large-scale datasets. In contrast, according to our experience, scribble annotation only costs  $\sim$10 seconds, which is 360 times faster than pixel-wise annotation. Second, pixel-wise annotation assigns equal significance to each object pixel, which may cause the model \ryn{to fail in learning} primary structures, \ryn{as shown in the second row of Figure~\ref{fig:teaser1}}. To address these problems, we propose the first scribble-based COD dataset, named S-COD. It contains 3,040 images from the training set of COD10K~\cite{fan2020camouflaged} and 1,000 images from the training set of CAMO~\cite{le2019anabranch}. Annotators are asked to scribble the primary structure according to their first impressions without knowing the ground-truth.
Figure~\ref{fig:percentage} \ryn{shows} the percentage of annotated pixels in S-COD.
Compared to pixel-wise annotation, the labeling process of S-COD is much easier.
\ryn{Compared with other labeling approaches} (\textit{e.g.}, box and point annotation), \qihua{it provides more pixel-level guidance},
allowing semantic information to be exploited, and is comparably efficient in labeling.

\begin{figure}[t]
    \centering
    \includegraphics[width=0.08\textwidth]{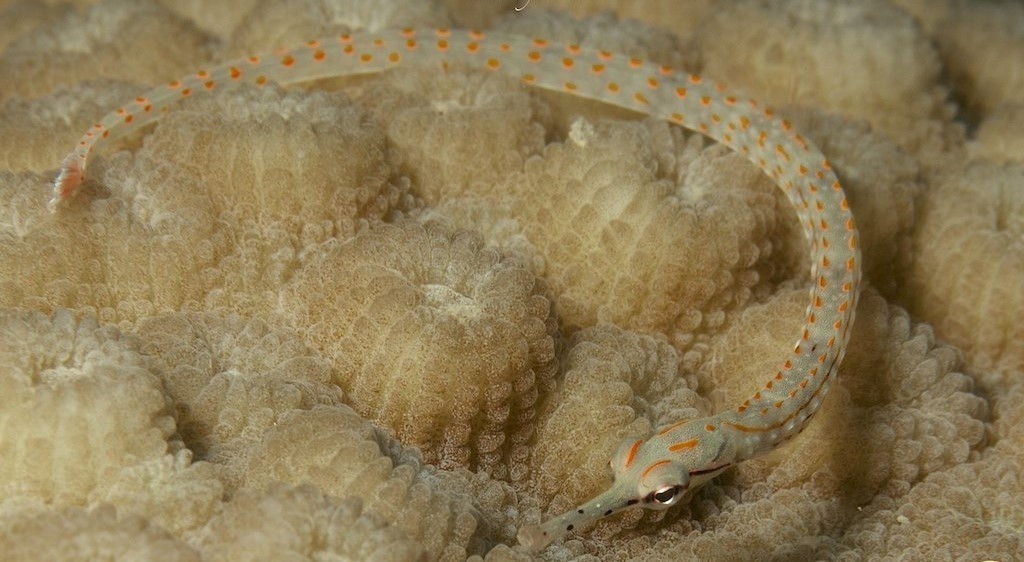}
    \includegraphics[width=0.08\textwidth]{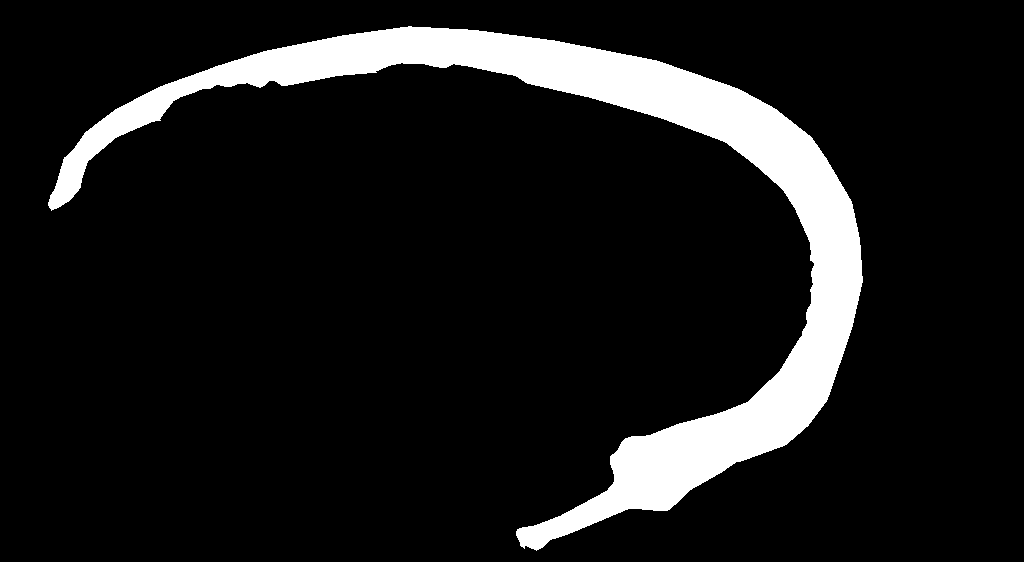}
    \includegraphics[width=0.08\textwidth]{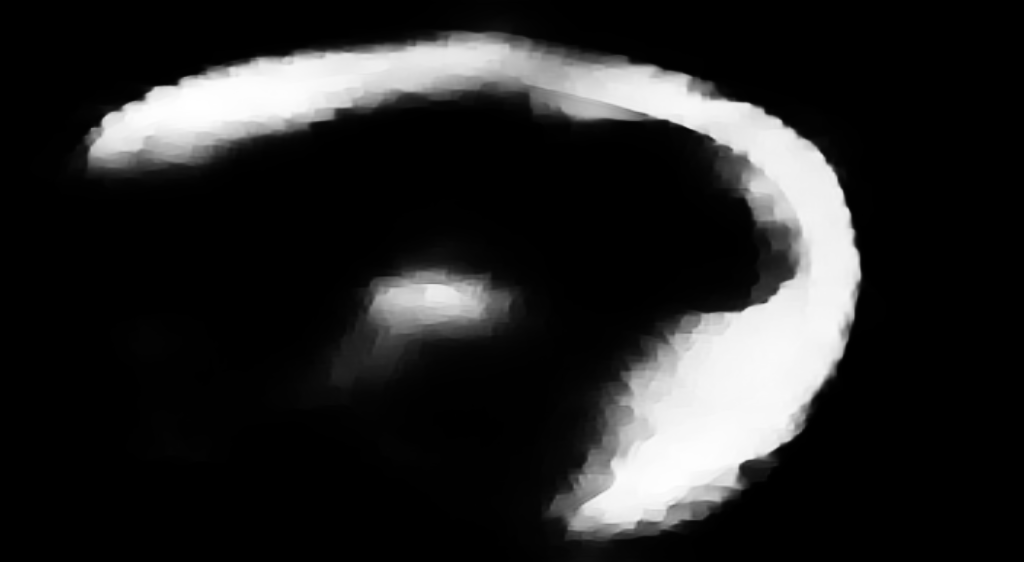}
    \includegraphics[width=0.08\textwidth]{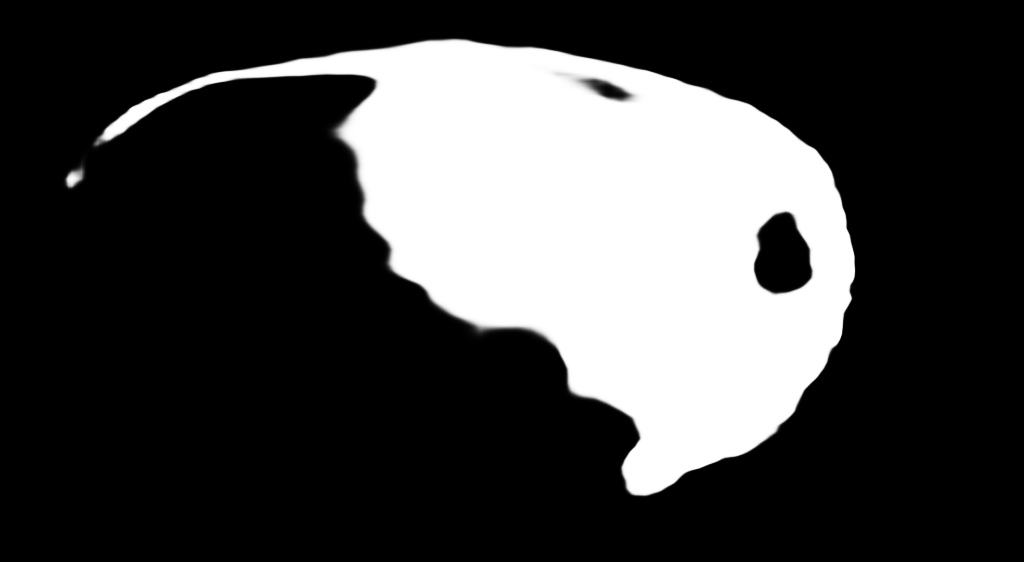}
    \includegraphics[width=0.08\textwidth]{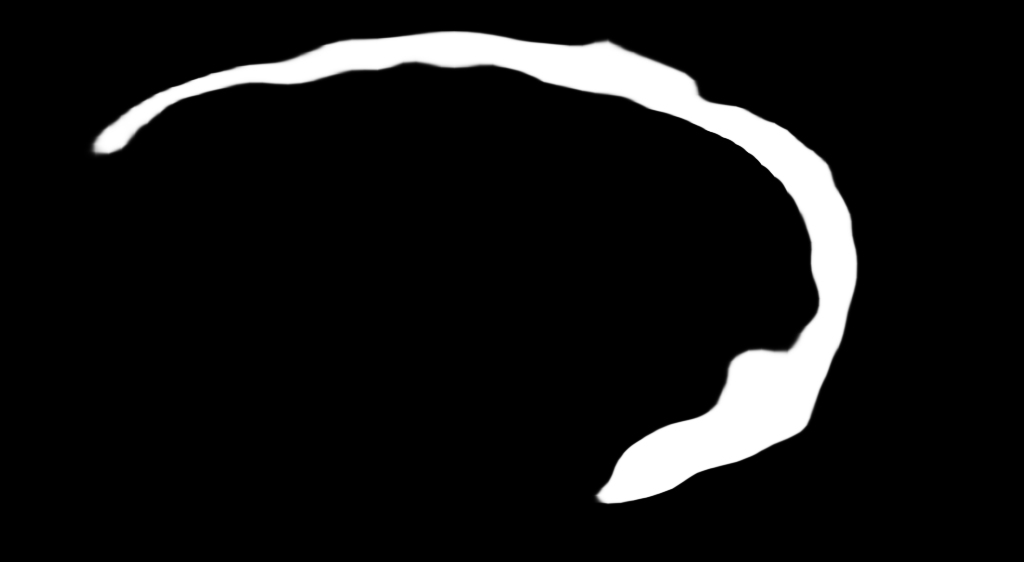}
    \\
    \includegraphics[width=0.08\textwidth]{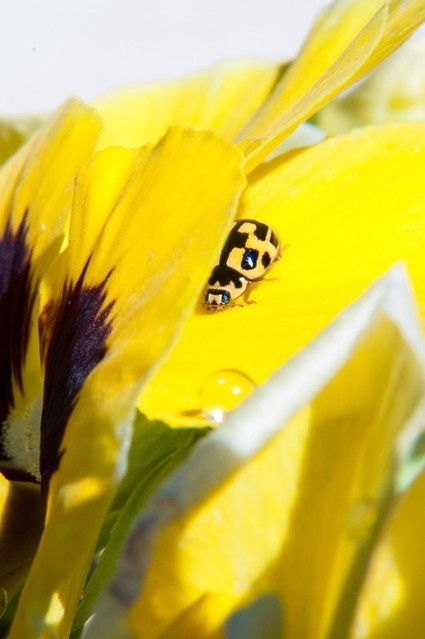}
    \includegraphics[width=0.08\textwidth]{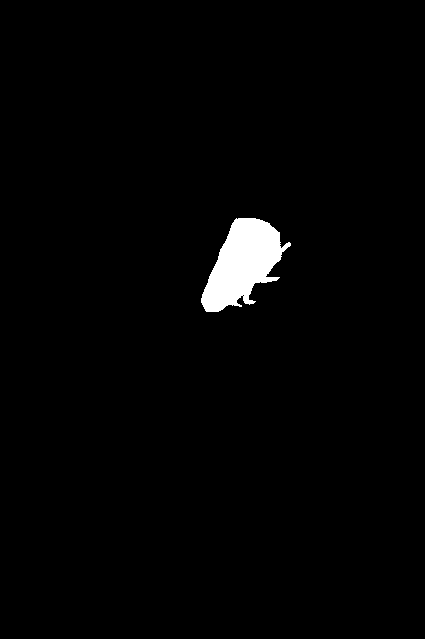}
    \includegraphics[width=0.08\textwidth]{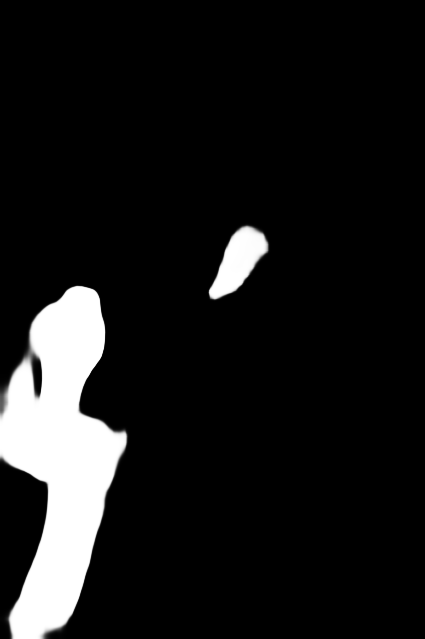}
    \includegraphics[width=0.08\textwidth]{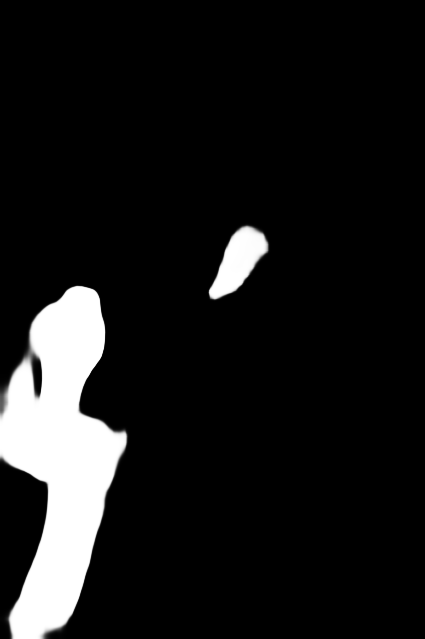}
    \includegraphics[width=0.08\textwidth]{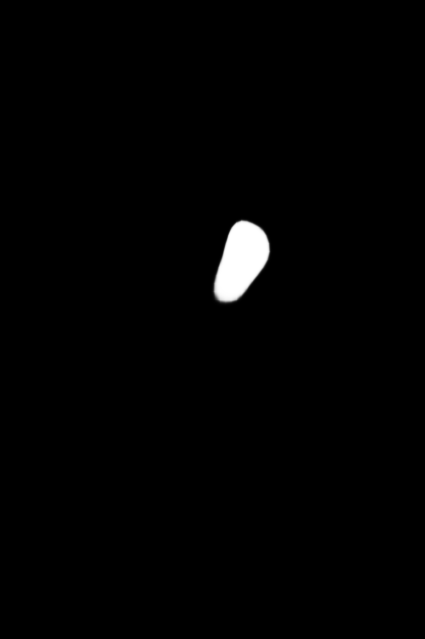}
    \\
    \makebox[0.08\textwidth]{\scriptsize Input}
    \makebox[0.08\textwidth]{\scriptsize GT}
    \makebox[0.08\textwidth]{\scriptsize SS}
    \makebox[0.08\textwidth]{\scriptsize SCWSSOD}
    \makebox[0.08\textwidth]{\scriptsize Ours}
    \\
    \caption{Two popular scenarios where existing scribble-supervised SOD methods SS~\cite{zhang2020weakly} and SCWSSOD~\cite{yu2021structure} fail \ryn{to exploit} semantic features.}
    \label{fig:teaser2}
\end{figure}

Nevertheless, how to exploit scribble annotations for COD is still under exploration. Directly applying existing scribble-based salient object detection (SOD) methods are not appropriate here since camouflaged objects are not salient.
Figure \ref{fig:teaser2} shows that two state-of-the-art scribble-based SOD methods, SS~\cite{zhang2020weakly} and SCWSSOD~\cite{yu2021structure}, fail in two common scenarios. The first row of Figure \ref{fig:teaser2} shows an object with an ambiguous boundary in the generally consistent background. Due to the similar low-level features, both SS and SCWSSOD experience difficulties recognizing the boundaries.  
The second row requires detectors to identify semantic relations of objects (\textit{e.g.,} flower stems and petals), as more than one object looks like the ``camouflaged'' foreground. 
Here, both SS and SCWSSOD mistakenly include \ryn{other objects as the foreground,} due to poor semantic information learning.

\ryn{In this paper, we present the first scribble-based COD learning framework to address the weakly-supervised COD problem with scribble annotations.}
We observe that humans would first identify possible foreground objects \cite{wald1935carotenoids} and then use semantic information to exactly segment them \cite{hubel1962receptive}. 
To incorporate this process in our \ryn{model}, we propose a feature-guided loss\ryn{, which considers} not only visual affinity but also high-level semantic features, to guide the segmentation. The high-level features are learned in an end-to-end fashion during training and do not depend on other well-trained detectors. In addition, in \ryn{our network design}, we propose the local-context contrasted (LCC) module \ryn{to mimick visual inhibition in strengthening contrast \cite{von2017sensory} in order} to find potential camouflaged regions, and the logical semantic relation (LSR) module to determine the final camouflaged object regions.
Further, we notice that current weakly-supervised methods tend to have inconsistent predictions in COD, possibly due to the ``camouflage'' characteristics. 
Hence, we design a \ryn{consistency regularization, which is stronger and more reliable than} previous weakly-supervised learning methods. Specifically, we introduce the reliability bias in the cross-view loss to improve the self-consistency mechanism. We also present the inside-view consistency loss to reduce the uncertainty of predictions. The regularization enhances the stability and quality of the prediction.

In conclusion, our main contributions are as follows:
\begin{itemize}
    \item We propose the first weakly-supervised COD dataset with scribble annotation. \ryn{Compared with pixel-wise annotation, it takes only  $\sim$10 seconds to annotate each image (360 times faster) and overcomes the limitation of assigning equal importance} to every object pixel. 
    \item We propose the first end-to-end weakly-supervised COD framework. It includes novel feature-guided loss functions and consistency loss. Imitating human perceptions, the loss functions guide the network to extract high-level features that help distinguish objects and impose stability on the predictions. 
    \item We propose a novel network for scribble learning, which utilizes low-level contrasts to expand the scribbles to wider camouflaged regions and logical semantic information to finalize the objects.
    \item Experimental results show that our framework outperforms relevant state-of-the-art methods on three COD benchmarks with an average improvement of 11.0$\%$ on MAE, 3.2$\%$ on S-measure, 2.5$\%$ on E-measure, and 4.4$\%$ on weighted F-measure. 
\end{itemize}

\section{Related Work}

\noindent\textbf{Camouflaged Object Detection.}
COD focuses on undetectable natural objects (\textit{e.g.,} hidden objects, tiny objects, objects with similar appearances to the surroundings) and undetectable artificial objects (\textit{e.g.,} synthetic pictures and artworks with information hiding purpose). 
\cite{fan2020camouflaged} proposes a COD dataset with 10K camouflaged images, which takes an average of around 60 minutes to annotate each image. 
\cite{zhai2021mutual} proposes a mutual graph learning method that splits the task into rough positioning and precise boundary locating. 
\cite{li2021uncertainty} applies joint learning on SOD and COD tasks, taking advantage of both tasks to meet a balance of global/local information. 
\cite{mei2021camouflaged} proposes a focus module to detect and remove false-positive and false-negative predictions. 
\cite{yang2021uncertainty} proposes a transformer-based probabilistic representational model to learn context information to solve uncertainty-guided ambiguity.
\cite{lin2022frequency} proposes a frequency-aware COD method.
\cite{youwei2022zoom} proposes a multi-scale network that employs the zoom strategy to learn mixed-scale semantics for accurate segmentation.

However, these methods highly rely on per-pixel ground-truth with full supervision, which is time-consuming and labor-intensive. To overcome these limitations, we propose scribble annotations to construct COD datasets, the first weakly-supervised dataset for COD task to our knowledge.

\section{Methodology}

The training dataset is defined as $D = \{x_n, y_n\}_{n=1}^{N_{img}}$, where $x_n$ is the input, $y_n$ is the annotation map, and $N_{img}$ is the total number of training images. In our task, $y_n$ is in scribble-form, in which 1 is foreground, 2 is background, and 0 is unknown pixels.

\begin{figure*}[h]
    \centering
    \includegraphics[width=1\textwidth]{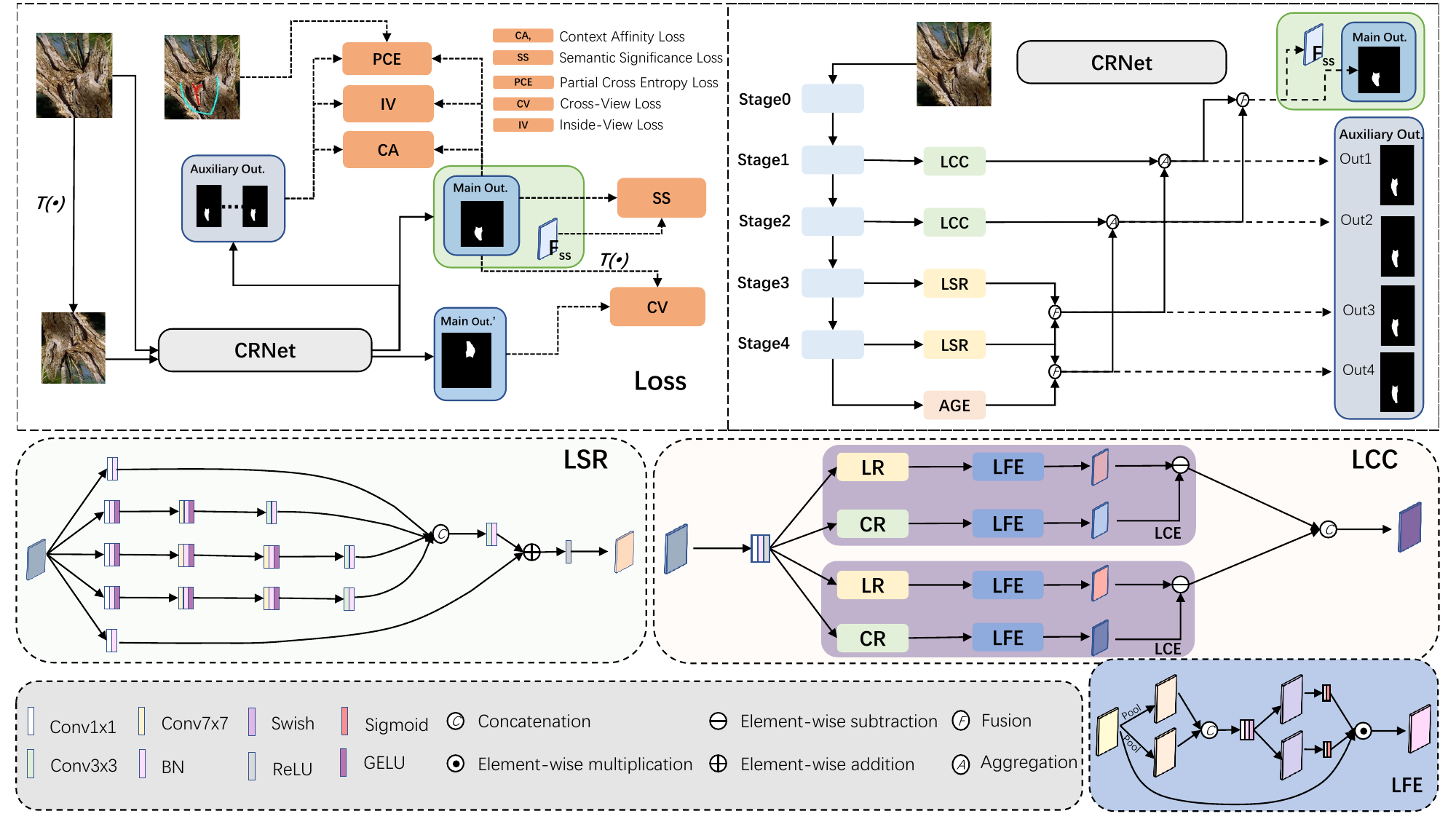}
    \caption{
    \qhrewrite{An overview of our proposed method. The top left figure illustrates the training process while the others show the CRNet architecture. During training, a random transform $T(\cdot)$ is applied to the input. Both the input and the transform are fed into the network, resulting in two outputs, and the CV loss is then computed based on them. $F_{ss}$ is the feature map of the input extracted from the CRNet and used to compute SS loss. The PCE, IV, CA loss are computed on auxiliary outputs and main outputs. Our contrast and relation network (CRNet) applies local-context contrasted (LCC) modules at the second and third stages, logical semantic relation (LSR) modules at the last two stages, and the auxiliary global extractor (AGE) at the last stage.}
    }
    \label{fig:network}
\end{figure*}

\subsection{Overall Structure}
The overall framework, including the proposed \textbf{C}ontrast and \textbf{R}elation \textbf{Net}work (CRNet) and loss functions, is shown in Figure~\ref{fig:network}. 
We first feed the input to the backbone ResNet-50 \cite{he2016deep} to obtain multi-scale input features $f_{i}$, where $i \in \{ x | 0 \leq x \leq 4, x \in \mathbb{N} \}$ denotes the stages of the backbone. CRNet uses local-context contrasted (LCC) modules for low-level features $F_{1}$, $F_{2}$ to extract contrasted features $F_{c}^{0}$, $F_{c}^{1}$, and logical semantic relation (LSR) modules for high-level input features $F_{3}$, $F_{4}$ to learn logical semantic information $F_{s}^{0}$, $F_{s}^{1}$. In addition, we design an auxiliary global extractor (AGE), a pyramid pooling module \cite{zhao2017pyramid} with GELU activation functions, to further acquire global semantic information $F_{s}^{g}$. 
Following the multiplication fusion and cross aggregation strategy \cite{zhao2021complementary},
we then fuse $F_{s}^{1}$ with $F_{s}^{0}$ and $F_{s}^{g}$, and integrate logical semantic information in $F_{l}^{0}$ and $F_{l}^{1}$, respectively. After aggregating $F_{c}^{0}$ with $F_{l}^{0}$ to $F_{out}^{0}$, $F_{c}^{1}$ with $F_{l}^{1}$ to $F_{out}^{1}$, CRNet further processes $F_{out}^{0}$ and $F_{out}^{1}$ and outputs multi-level segmentation maps (main output $P$ and auxiliary output $P_1 \text{ to } P_4$). We also extract an intermediate feature map ($F_{ss}$) for loss computation. 
During training, feature-guided loss (context affinity (CA) loss and semantic significance (SS) loss) are applied to guide the segmentation, while consistency loss (cross-view (CV) and inside-view (IV) loss) ensures the consistency of predictions.

\subsection{Local-Context Contrasted \ryn{(LCC)} Module} 
\rynq{As camouflaged objects usually share different low-level features (\textit{e.g.,} texture, color, intensity) with backgrounds, it is not easy to notice the inconspicuous differences. 
}
Visual inhibition on the mammalian retina enhances the sharpness and contrast in visual response by inhibiting \ryn{the activities of neighbor cells} \cite{von2017sensory}. Inspired by this, we propose a local-context contrasted (LCC) module to capture \ryn{and strengthen low-level differences}.
Here, a low-level contrasted extractor (LCE) uses two low-level feature extractors (LFE) with different receptive fields to represent local and context features (\textit{i.e.}, neighbors), and computes the difference to increase low-level contrast and sharpness. Furthermore, We stack two LCEs in LCC to further strengthen the low-level contrasts.
The contrast information learned by LCC helps expand scribbles to potential camouflaged regions, allowing our method to better command the object's primary structure and potential boundary.

LCC processes the input low-level features $F_{in}$, which contain informative texture, color, and intensity characteristics through two branches of low-level contrasted extractors with different receptive fields. We first reduce $F_{in}$'s channel number to 64 by a 1$\times$1 convolutional layer with batch normalization and ReLU, and then take the obtained $F_{low} \in 64 \times H \times W$ to two low-level contrasted extractors (LCEs) focusing on different sizes of fields. An LCE consists of a local receptor (LR), a context receptor (CR), and two low-level feature extractors (LFE). $F_{low}$ go through an LR, which is a 3$\times$3 convolutional layer with 1 dilation rate and an LFE to obtain $F_{local}$. Meanwhile, \rynq{$F_{low}$ 
}are also extracted by a CR, which is a 3$\times$3 convolutional layer with dilation $d_{context}$, and further by an LFE for $F_{context}$. We take the subtraction of $F_{local}$ and $F_{context}$ into batch normalization and ReLU to get one level contrasted features $F_{contrast}$. We set $d_{context}$ to 4 and 8 for two levels of LCE, extracting low-level contrasted features $F_{contrast}^{1}$ and $F_{contrast}^{2}$ concentrating on different sizes of receptive fields. The final output is a concatenation of $F_{contrast}^{1}$ and $F_{contrast}^{2}$. \textit{Refer to the Supplemental for LCC implementation details.}

\subsection{Logical Semantic Relation \ryn{(LSR)} Module} 
Scribble annotation may only annotate a part of the background. When the background consists of many low-level contrasted parts (\textit{e.g.,} green leaves and brown branches, yellow petals, and green stems), we need logical semantic relation information to identify the real foreground and background. Hence, we propose \ryn{the LSR module} to extract semantic features from 4 branches. Each branch contains a sequence of convolution layers with different kernel sizes and dilation rates, representing different receptive fields.  \ryn{We then integrate information from all branches to exploit comprehensive semantic information with a wider receptive field to determine the real foreground and background. 
} 
\textit{Refer to the Supplemental for LSR implementation details.}

\subsection{Feature-guided Loss}
\begin{figure}[t] \centering
    \includegraphics[width=0.1\textwidth]{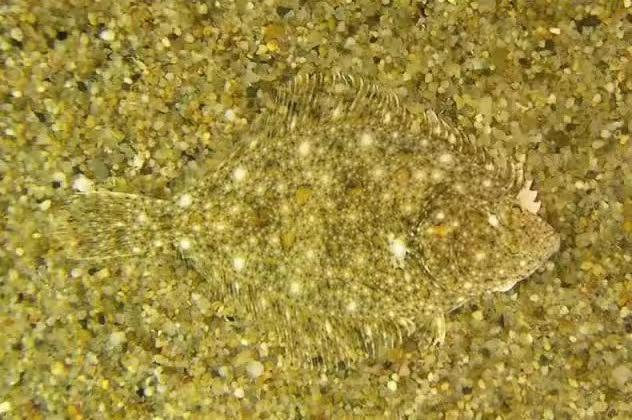}    
    \includegraphics[width=0.1\textwidth]{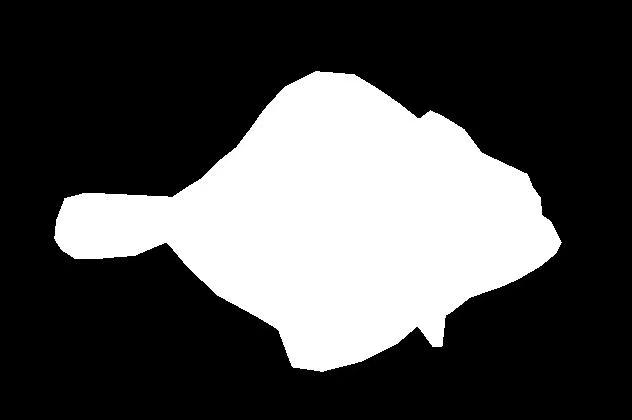}  
    \includegraphics[width=0.1\textwidth]{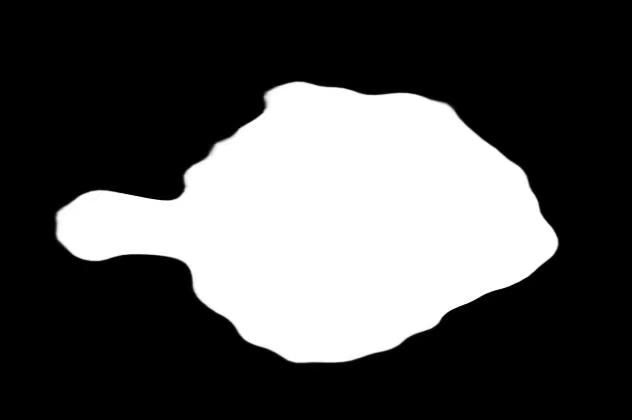}  
    \includegraphics[width=0.1\textwidth]{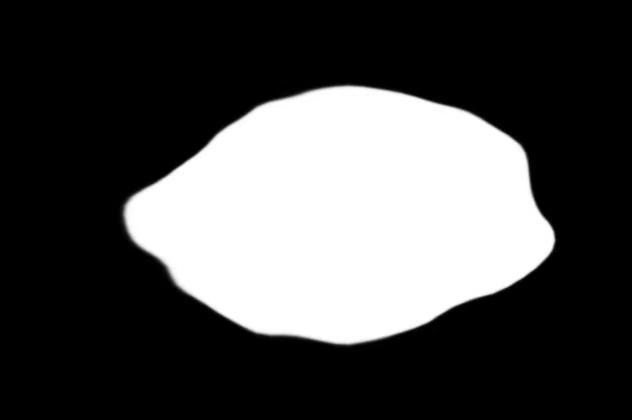}  
    \\
    \makebox[0.1\textwidth]{\scriptsize Input}
    \makebox[0.1\textwidth]{\scriptsize  GT}
    \makebox[0.1\textwidth]{\scriptsize  Pred.}
    \makebox[0.1\textwidth]{\scriptsize  Pred.$'$}
    \\
    \makebox[0.4\textwidth]{\scriptsize (a) Predictions on input (Pred.) and its transform (Pred.$'$)}
    \\  
    \makebox[0.1\textwidth]{}\\
    
    \includegraphics[width=0.1\textwidth]{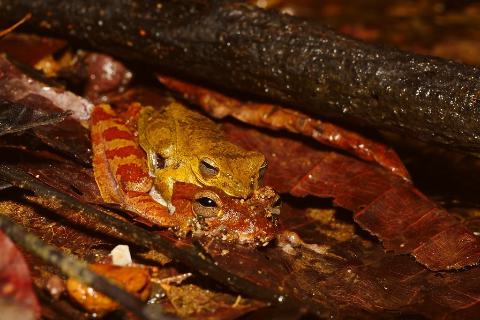}
    \includegraphics[width=0.1\textwidth]{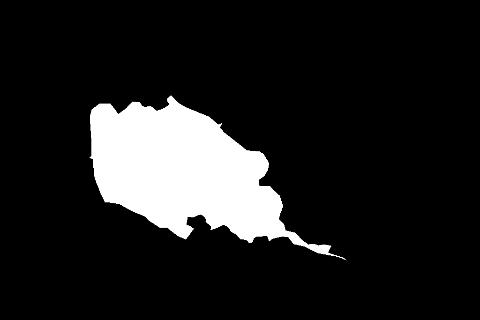}
    \includegraphics[width=0.1\textwidth]{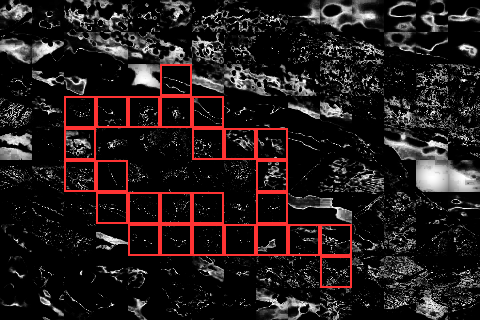}
    \includegraphics[width=0.1\textwidth]{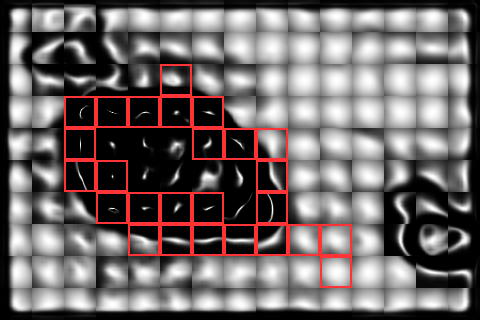}
    \\
    \includegraphics[width=0.1\textwidth]{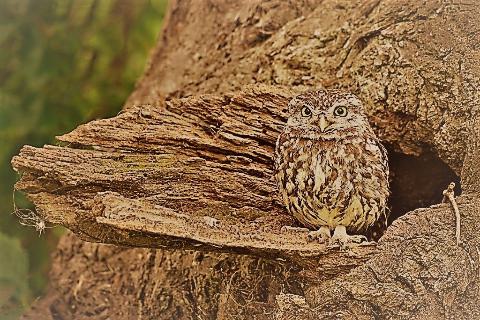}
    \includegraphics[width=0.1\textwidth]{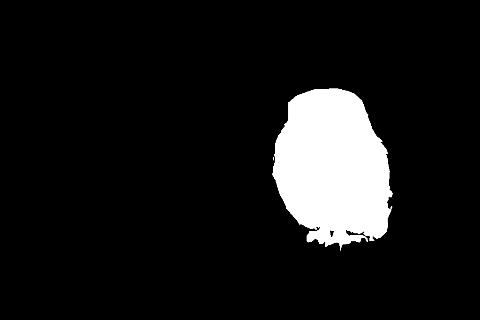}
    \includegraphics[width=0.1\textwidth]{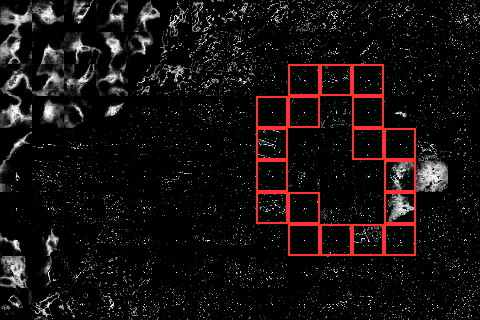}
    \includegraphics[width=0.1\textwidth]{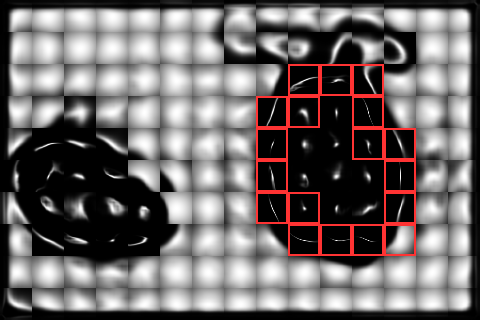}
    \\
    \makebox[0.1\textwidth]{\scriptsize Input}
    \makebox[0.1\textwidth]{\scriptsize  GT}
    \makebox[0.1\textwidth]{\scriptsize  VF}
    \makebox[0.1\textwidth]{\scriptsize  SS}
    \\
    \makebox[0.4\textwidth]{\scriptsize (b) Visualization of the visual featured (VF) and semantic featured (SS) kernels}
    \\
    \caption{(a) shows that prediction on normal input is more accurate than on its transform. The \ryn{design of the CV loss} considers this reliability bias; (b) visualizes kernels of the visual features (VF) in $K_{vis}$ and learnt semantic features (SS) in $K_{sem}$. Images are divided into $32\times 32$ blocks (red blocks means boundary regions). We calculate the kernels with respect to the center pixels (anchors) inside blocks. White indicates high energy when the pixel label differs from the anchor.}
    \label{fig:reliable&ss}
\end{figure}

Scribble-based methods often suffer from the lack of object information provided by the limited labeled data. Previous methods \cite{zhang2020weakly, yu2021structure}  exploit the information by using the pixel features of images, like colors and positions, assuming that foreground objects have visually distinctive features from backgrounds. However, in COD, such features are no longer a strong cue for boundary regions. It usually requires semantic information to decide the exact boundaries. Therefore, we design feature-guided loss based on both simple visual features (context affinity loss) and complex semantic features (semantic significance loss). As shown in Figure~\ref{fig:reliable&ss}(b), semantic features extracted from the model respond actively to camouflaged boundaries and provide valuable guidance in these regions.

\noindent\textbf{Context Affinity Loss.}
Nearby pixels with similar features tend to have the same class. Following previous methods \cite{obukhov2019gated, yu2021structure}, we adopt the kernel method to measure the visual feature similarity (colors and positions), which is defined as:
\begin{align}
    K_{vis}(i,j) &= \exp(- \frac{||S(i) - S(j)||^2}{2\sigma_S^2} - \frac{||C(i) - C(j)||^2}{2\sigma_C^2}),
    \label{eq:vf}
\end{align}
where $S(i), C(i)$ are the position $(x_i,y_i)$ and colors $(r_i,g_i,b_i)$ of pixel $i$. $\sigma_S, \sigma_C$ are hyperparameters. $D(i,j)$ calculates the probability of pixel $i,j$ having different classes ($P_{i,j}$ is the probability of positive labels for pixel $i,j$), and thus the context affinity loss $L_{ca}$ encourages visually dissimilar pixels \rynq{to have different labels} or vice versa: 
\begin{gather}
    D(i, j) = 1-P_i P_j-(1-P_i)(1-P_j), \\
    L_{ca} = \frac{1}{M} \sum_{i} \frac{1}{K_d(i)} \sum_{j \in K_d(i)} K_{vis}(i,j) D(i,j),
\end{gather}
where $K_d(i)$ is a neighbor $n\times n$ regions ($n$ is set to 5 in our experiments) \ryn{of center} pixel $i$. Through context affinity loss, the model can quickly learn from the unlabeled pixels.

\noindent\textbf{Semantic Significance Loss.}
In COD, pixels near boundaries usually resemble each other visually, and semantic features, especially those that distinguish segmented objects (thus significant), become crucial for the exact predictions. In this case, we design the semantic significance (SS) loss that utilizes significant features to refine the predictions of boundary regions.

Here, the SS loss is computed inside small boundary regions (in practice, we divide an image to $8\times 8$ region blocks ($R_{1,...,r}$) with a step size of 4). A valid boundary region is defined as \ryn{an area} with at least 30\% of the pixels being confidently classified as foreground \ryn{or} background (pixels with scribble annotation or model prediction above 0.8 is confidently classified). The design has two benefits. First, in non-boundary regions, low-level visual features suffice to provide good guidance. Second, it reduces the computation cost greatly. 
The semantic feature map $F_{ss} \in \mathbb{R}^{H\times W\times C}$ is extracted before the final prediction layer 
\footnote{For example, if the final layer is a $3\times 3$ convolution layer with \rynq{64 input channels }, \rynq{1 output channel} (1 since it is binary segmentation), it can be seen as first achieving $F_{ss}$ through a $3\times3$ conv layer with 64 input channels, 64 output channels in 64 groups, and then getting $P$ by a sum pooling on each channel, \textit{i.e.} $P_i=\sum_c^n F_{ss_{i,c}}$, where $i,n$ is the pixel index and channel number.} and its gradient is stopped (like the detach operation in Pytorch).
The significance of a featured channel is determined by its covariance with confidently classified predictions:
\begin{align}
    {Sig}_i = cov'(F_{ss_i}, P), i \in \{1, ..., C\},
\end{align}
where $F_{ss_i}$ is the feature map of the i-th channel and $cov'$ means covariance, computed only on confidently classified pixels. The reason behind is that the above correlations roughly show how well the features distinguish the foreground and background. Low-significance features are unwanted since they may include the “camouflaged” parts of the object and confuse the model. 

\ryn{We then} take the top $N$ channels ordered by ${Sig}$ to form significant feature map $\hat{F_{ss}} \in \mathbb{R}^{H\times W\times N}$. In this task, we set $N$ to 16 to balance between performance and computation cost. The semantic significance loss has a similar formulation to context affinity loss:
\begin{align}
    \label{eq:ss}
    K_{sem} &= \exp(- \frac{||S(i) - S(j)||^2}{2\sigma_S^2} - \frac{||\hat{F_{ss}}(i) - \hat{F_{ss}}(j)||^2}{2\sigma_C^2}), \\
    L_{ss} &= w_{ss} \frac{1}{M} \sum_k \frac{1}{|R_k|} \sum_{i,j \in R_k} K_{sem}(i,j)D(i,j),
\end{align}
where $S(i)$ is the position of the pixel, $R_k$ are valid boundary regions, and $w_{ss}$ is set to increase with the epoch number (exponential ramp-up to 0.15 in practice) since the model has not learned well-represented features at the beginning.  

In conclusion, the feature loss $L_{ft}$ can be written as the sum of both loss in $L_{ft} = L_{ca} + L_{ss}$.

\subsection{Consistency Loss}
Weakly-supervised methods often suffer from inconsistent predictions. Similar to self-consistency mechanisms in \ryn{self-supervision and weakly-supervision}~\cite{laine2016temporal, mittal2019semi, yu2021structure, pan2021scribble}, we propose the cross-view (CV) consistency loss to alleviate the problem by minimizing the difference between the predictions of the input and its transform. Compared to others, the CV loss excels in that it considers the reliable difference. As shown in Figure~\ref{fig:reliable&ss}(a), we observe that the model has more reliable output with normal input than transformed input, which is plausible considering more loss functions are computed on normal input. The proposed CV loss pushes the predictions to the reliable one and leads to a solid improvement in performance. In addition, the predictions tend to be uncertain due to visual similarity between background and foreground in COD, and we design an inside-view consistency loss to improve the stability of predictions.

\begin{table*}[t]
\centering
\caption{Quantitative comparison with state-of-the-arts on three benchmarks. ``F'', ``U'', and ``W'' denote fully-supervised, unsupervised, and weakly-supervised \ryn{methods}, respectively.}
\scalebox{0.7}{
\begin{tabular}{l|c|rrrr|rrrr|rrrr}
\cline{1-14}
\multicolumn{1}{c|}{}        &      & \multicolumn{4}{c|}{CAMO}                                                                              & \multicolumn{4}{c|}{CHAMELEON}                                                                       & \multicolumn{4}{c}{COD10K}                                                                        \\
\multicolumn{1}{c|}{Methods} & Sup. & \multicolumn{1}{c}{MAE$\downarrow$}   & \multicolumn{1}{c}{S$_{m}\uparrow$} & \multicolumn{1}{c}{E$_{m}\uparrow$} & \multicolumn{1}{c|}{F$_{\beta}^{w}\uparrow$} & \multicolumn{1}{c}{MAE$\downarrow$} & \multicolumn{1}{c}{S$_{m}\uparrow$} & \multicolumn{1}{c}{E$_{m}\uparrow$} & \multicolumn{1}{c|}{F$_{\beta}^{w}\uparrow$} & \multicolumn{1}{c}{MAE$\downarrow$} & \multicolumn{1}{c}{S$_{m}\uparrow$} & \multicolumn{1}{c}{E$_{m}\uparrow$} & \multicolumn{1}{c}{F$_{\beta}^{w}\uparrow$}       \\ \cline{1-14}
NLDF~\cite{luo2017non}                         & F    & 0.123                     & 0.665                  & 0.664                  & 0.495                    & 0.063                   & 0.798                  & 0.809                  & 0.652                    & 0.059                   & 0.701                  & 0.709                  & 0.473                         \\
PiCANet~\cite{liu2018picanet}                      & F    & 0.125                     & 0.701                  & 0.716                  & 0.510                    & 0.085                   & 0.765                  & 0.778                  & 0.552                    & 0.081                   & 0.696                  & 0.712                  & 0.415       \\

CPD~\cite{wu2019cascaded}                          & F    & 0.113                     & 0.716                  & 0.723                  & 0.556                    & 0.048                   & 0.857                  & 0.874                  & 0.731                    & 0.053                   & 0.750                  & 0.776                  & 0.531                    \\
EGNet~\cite{zhao2019egnet}                        & F    & 0.109                     & 0.732                  & 0.800                  & 0.604                    & 0.065                   & 0.797                  & 0.860                  & 0.649                    & 0.061                   & 0.736                  & 0.810                  & 0.517                      \\
PoolNet~\cite{liu2019simple}                      & F    & 0.105 & 0.730                  & 0.746                  & 0.575                    & 0.054                   & 0.845                  & 0.863                  & 0.690                    & 0.056                   & 0.740                  & 0.776                  & 0.506            \\
SCRN~\cite{wu2019stacked}                         & F    & 0.090                     & 0.779                  & 0.797                  & 0.643                    & 0.042                   & 0.876                  & 0.889                  & 0.741                    & 0.047                   & 0.789                  & 0.817                  & 0.575              \\
F3Net~\cite{wei2020f3net}                        & F    & 0.109                     & 0.711                  & 0.741                  & 0.564                    & 0.047                   & 0.848                  & 0.894                  & 0.744                    & 0.051                   & 0.739                  & 0.795                  & 0.544                \\

CSNet~\cite{gao2020highly}                        & F    & 0.092                     & 0.771                  & 0.795                  & 0.641                    & 0.047                   & 0.856                  & 0.868                  & 0.718                    & 0.047                   & 0.778                  & 0.809                  & 0.569                    \\
ITSD~\cite{zhou2020interactive}                         & F    & 0.102                     & 0.750                  & 0.779                  & 0.610                    & 0.057                   & 0.814                  & 0.844                  & 0.662                    & 0.051                   & 0.767                  & 0.808                  & 0.557                 \\
MINet~\cite{pang2020multi}                        & F    & 0.090                     & 0.748                  & 0.791                  & 0.637                    & 0.036                   & 0.855                  & 0.914                  & 0.771                    & 0.042                   & 0.770                  & 0.832                  & 0.608                \\

PraNet~\cite{fan2020pranet}                       & F    & 0.094                     & 0.769                  & 0.825                  & 0.663                    & 0.044                   & 0.860                  & 0.907                  & 0.763                    & 0.045                   & 0.789                  & 0.861                  & 0.629              \\

UCNet~\cite{zhang2020uc}                        & F    & 0.094                     & 0.739                  & 0.787                  & 0.640                    & 0.036                   & 0.880                  & 0.930                  & 0.817                    & 0.042                   & 0.776                  & 0.857                  & 0.633              \\ \cline{1-14}
SINet~\cite{fan2020camouflaged}                        & F    & 0.092                     & 0.745                  & 0.804                  & 0.644                    & 0.034                   & 0.872                  & 0.936                  & 0.806                    & 0.043                   & 0.776                  & 0.864                  & 0.631        \\
SLSR~\cite{lv2021simultaneously}                         & F    & 0.080                     & 0.787                  & 0.838                  & 0.696                    & 0.030                   & 0.890                  & 0.935                  & 0.822                    & 0.037                   & 0.804                  & 0.880                  & 0.673           \\
MGL-R~\cite{zhai2021mutual}                        & F    & 0.088                     & 0.775                  & 0.812                  & 0.673                    & 0.031                   & 0.893                  & 0.917                  & 0.812                    & 0.035                   & 0.814                  & 0.851                  & 0.666               \\
PFNet~\cite{mei2021camouflaged}                        & F    & 0.085                     & 0.782                  & 0.841                  & 0.695                    & 0.033                   & 0.882                  & 0.931                  & 0.810                    & 0.040                   & 0.800                  & 0.877                  & 0.660         \\
UJSC~\cite{li2021uncertainty}                         & F    & 0.073                     & 0.800                  & 0.859                  & 0.728                    & 0.030                    & 0.891                  & 0.945                  & 0.833                    & 0.035                   & 0.809                  & 0.884                  & 0.684        \\
C2FNet~\cite{sun2021context}                       & F    & 0.080                     & 0.796                  & 0.854                  & 0.719                    & 0.032                   & 0.888                  & 0.935                  & 0.828                    & 0.036                   & 0.813                  & 0.890                  & 0.686             \\
UGTR~\cite{yang2021uncertainty}                         & F    & 0.086                     & 0.784                  & 0.822                  & 0.684                    & 0.031                   & 0.888                  & 0.910                  & 0.794                    & 0.036                   & 0.817                  & 0.852                  & 0.666         \\ 
ZoomNet~\cite{youwei2022zoom}   & F & 0.066 &  0.820 & 0.892 & 0.752 &
0.023 & 0.902 & 0.958 & 0.845 &
0.029 & 0.838 & 0.911 & 0.729     \\ 
\cline{1-14}
DUSD~\cite{zhang2018deep}                     & U    & 0.166                     & 0.551             &  0.594                 & 0.308                    &  0.129                  & 0.578                  &  0.634               & 0.316                   & 0.107                   & 0.580              & 0.646                  & 0.276              \\
USPS~\cite{nguyen2019deepusps}                             & U    &       0.207                    & 0.568                       & 0.641                       & 0.399                         & 0.188                        & 0.573                       & 0.631                    & 0.380                         & 0.196                        & 0.519                       & 0.536                       & 0.265                                \\
SS~\cite{zhang2020weakly}                           & W    & 0.118                     & 0.696                  & 0.786                  & 0.562                    & 0.067                   & 0.782                  & 0.860                   & 0.654                    & 0.071                   & 0.684                  & 0.770                  & 0.461             \\
SCWSSOD~\cite{yu2021structure}                      & W    & 0.102                     & 0.713                  & 0.795                  & 0.618                    & 0.053                   & 0.792                  & 0.881                  & 0.714                    & 0.055                   & 0.710                  & 0.805                  & 0.546     \\ \cline{1-14}
Ours                         & W    & \textbf{0.092}                     & \textbf{0.735}                  & \textbf{0.815}                  & \textbf{0.641}                    & \textbf{0.046}                   & \textbf{0.818}                  & \textbf{0.897}                  & \textbf{0.744}                    & \textbf{0.049}                   & \textbf{0.733}                 & \textbf{0.832}                  & \textbf{0.576}               \\ \cline{1-14}
\end{tabular}
}
\label{tab:eval}
\end{table*}

\noindent\textbf{Cross-View Consistency Loss.}
For a neural network function $f_{\theta}(\cdot)$ with parameter $\theta$, some transformations $T(\cdot)$, input $x$, the ideal situation is $f_\theta(T(x)) = T(f_\theta(x))$. Here, the transform includes combinations of resizing, flipping, translation and cropping, and is randomly chosen. The choice of it is explored in the ablation study.
As regularization, we use the similar consistency loss $L_{cv'}$ \cite{yu2021structure} to push them towards each other.
\begin{gather}
    Sm(p_1, p_2)= {1-{SSIM}(p_1, p_2) \over 2},
\\
   L_{cv'}(P_1, P_2) = {1\over M} \sum_{i} (1-\alpha) \cdot Sm(P_{1_i}, P_{2_i}) + \alpha \vert P_{1_i}-P_{2_i}\vert,
\end{gather}
where SSIM is single scale SSIM \cite{godard2017unsupervised}. $p_1, p_2$ are two pixels. $\alpha$ is 0.85. $P_1, P_2$ are prediction maps of the input and its transform. $M$ is the total number of pixels and $i$ is a pixel index. 

Considering the above-mentioned reliability bias, we \ryn{aim for the predictions of the transform $\hat{P}$ to be} pushed more than that of the normal input $P$. The key here is to weight their backward gradient differently, and the proposed cross-view consistency loss can be written as:
\begin{align}
    L_{cv\ } &= (1+\gamma) L_{cv'}(P^d, \hat{P}) + (1-\gamma) L_{cv'}(P,\hat{P}^d),
\end{align}
where \rynq{$P^d, \hat{P}^d$ have the same values as $P, \hat{P}$ yet the gradient on them will be ignored during back-propagation (like the detach operation in PyTorch)}. If $\gamma=0$, it is the original loss $L_{cv'}$; if $\gamma>0$, the backward gradient that pushes $\hat{P}$ to $P$ is greater than the other way around, and thus the goal is reached. In practice, $\gamma$ is set to 0.3.

\noindent\textbf{Inside-view Consistency Loss.}
\ryn{We note} that uncertain predictions are likely to be inconsistent. Therefore, we present the inside-view consistency (IV) loss which "looks" inside the output and encourages predictions with high certainty by minimizing their entropy. We also use a soft indicator to filter out noisy predictions: when the entropy is above a certain threshold, the prediction result is not sure and it is malicious to increase the certainty of the model in this case. 
The inside-view consistency loss is as below.
\begin{align}
    L_{iv} &= w_{iv} \cdot {1 \over \vert I - \mathcal{B} \vert}\sum_{(i) \in {I - \mathcal{B}}} -P_{i}\log P_{i} - (1-P_{i})\log (1-P_{i}),
\end{align}
where $I, \mathcal{B}$ are the set of all pixels and noisy pixels. $i$ is the pixel index. $w_{iv}$ is the weight of this loss and set to 0.05 in practice. The entropy threshold for the near-boundary pixel is set to 0.5 \ryn{empirically}. The loss is added in the late stage of training when predictions are relatively accurate.

Combined with all the consistency \ryn{losses}, we have the final consistency loss: $L_{cst} = L_{cv} + L_{iv}$.

\subsection{Objective Function}
Below is PCE loss, where $\tilde{P}$ is the set of labeled pixels in the scribble map, $y_i$ is the true class of pixel $i$, and $\hat{y_i}$ are the predictions on pixel $i$:
\begin{align}
    L_{pce} &= {1\over N}\sum_{i\in \tilde{P}} -y_{i} \log{\hat{y}_i} - (1-y_i)\log({1-\hat{y}_i}),
\end{align}
We compute all losses on main output $P$ while for the auxiliary outputs ($P_{1...4}$), we compute only the PCE loss, inside-view consistency, and context affinity loss. $L_{aux}^i = L_{pce}^i + L_{ca}^i + L_{iv}^i (i=1,2,3,4)$, where $L_{aux}^i$ is the loss function applied to \ryn{the} i-th auxiliary output. Here, we do not include the other two losses for their \ryn{small} improvement, possibly because they require high-level feature representations or accurate segmentation to guide the model. Every output is up-sampled by linear interpolation to the same size as the input. Finally, the total objective function of our output is: 
$L = L_{cst} + L_{ft} + L_{pce} + \sum_{i=1}^{4} \beta_i L_{aux}^i$, where $\beta_i={1-0.2i}$.

\section{Experiments}

\subsection{Datasets and Implementation Details}
\noindent\textbf{Datasets.} Our experiments are conducted on three COD benchmarks, CAMO\cite{le2019anabranch}, CHAMELEON\cite{skurowski2018animal}, and COD10K\cite{fan2020camouflaged}. Following previous studies \cite{fan2020camouflaged, mei2021camouflaged, zhai2021mutual}, we relabel 4,040 images (3,040 from COD10K, 1,000 from CAMO) and propose the S-COD dataset for training. The remaining is for testing. 

\noindent\textbf{Evaluation Metrics.} We adopt four evaluation metrics: Mean Absolute Error (MAE), S-measure $S_{m}$ \cite{fan2017structure}, E-measure ($E_{m}$) \cite{fan2018enhanced}, and weighted F-measure $F_{\beta}^{w}$ \cite{margolin2014evaluate}.

\noindent\textbf{Implementation Details.} We implement our method with PyTorch and conduct experiments on a GeForce RTX2080Ti GPU. In the training phase, input images are resized to 320$\times$320 with horizontal flips. We use the stochastic gradient descent (SGD) optimizer with a momentum of 0.9, a weight decay of 5e-4, and triangle learning rate schedule with maximum learning rate 1e-3. The batch size is 16, and the training epoch is 150. It takes around 5 hours to train. As for the inference process, input images are only resized to 320$\times$320. We then directly predict the final maps without any post-processing (\textit{e.g.}, CRF).

\subsection{Analysis}
\begin{figure}[t] \centering
    \includegraphics[width=0.05\textwidth]{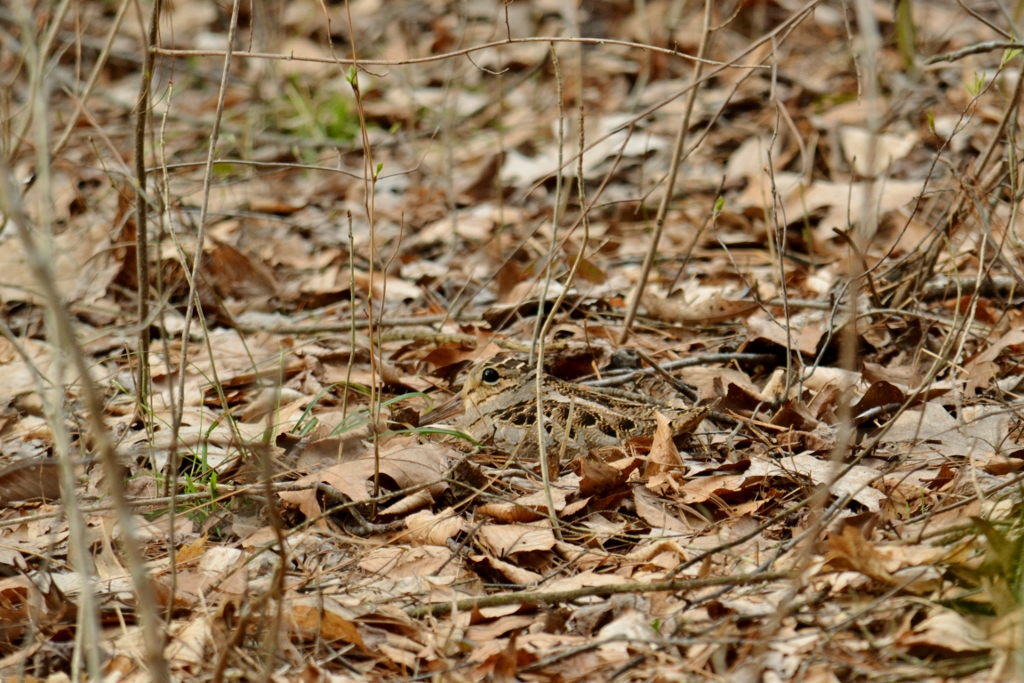}
    \includegraphics[width=0.05\textwidth]{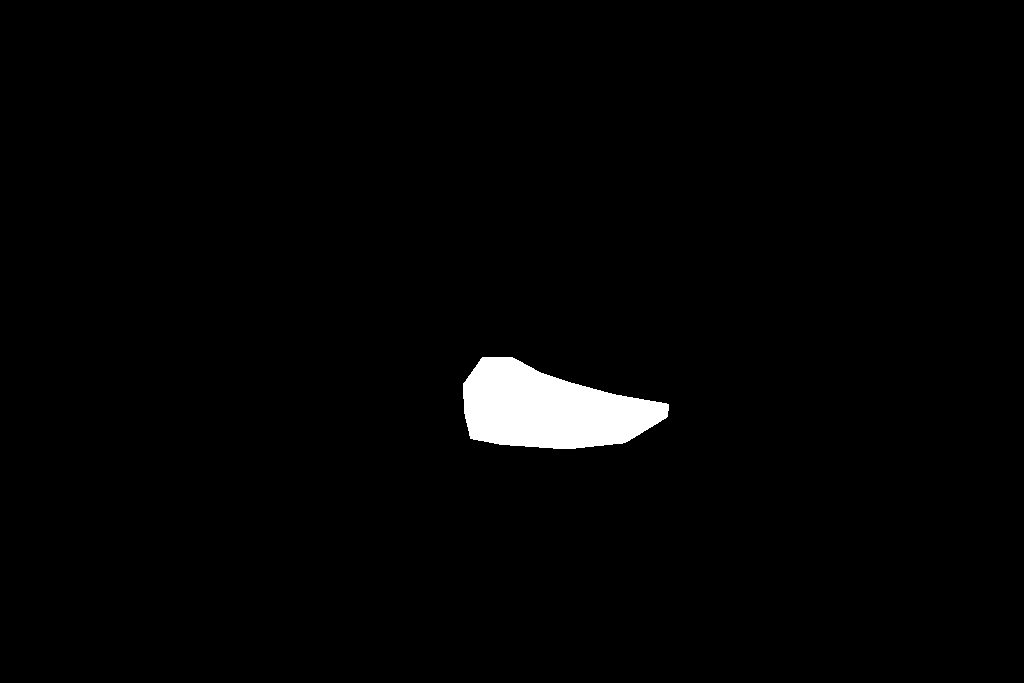}
    \includegraphics[width=0.05\textwidth]{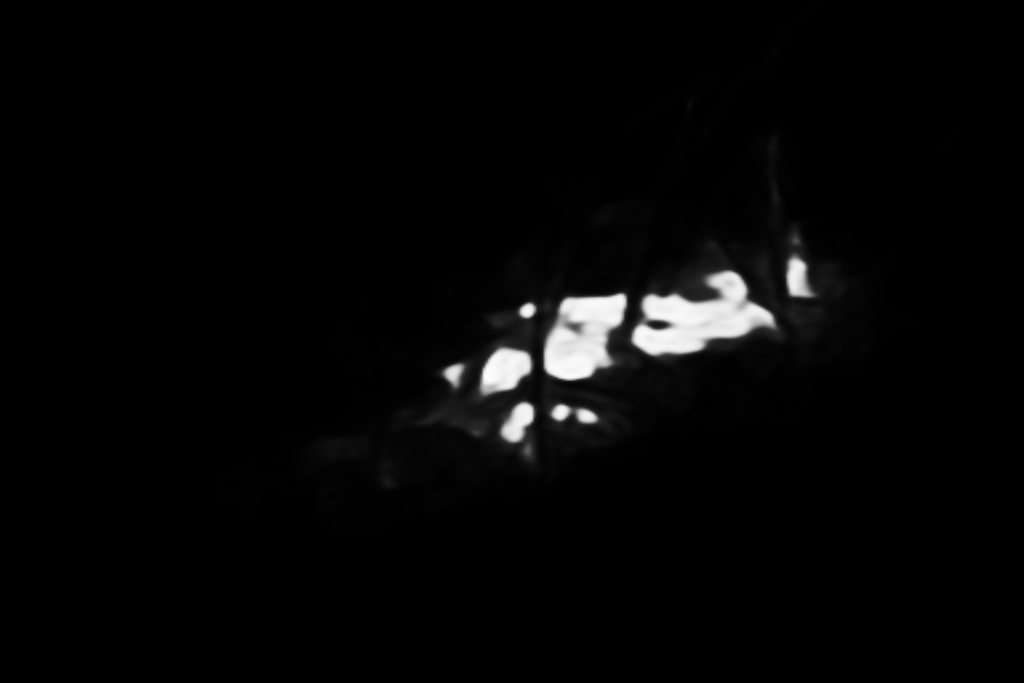}
    \includegraphics[width=0.05\textwidth]{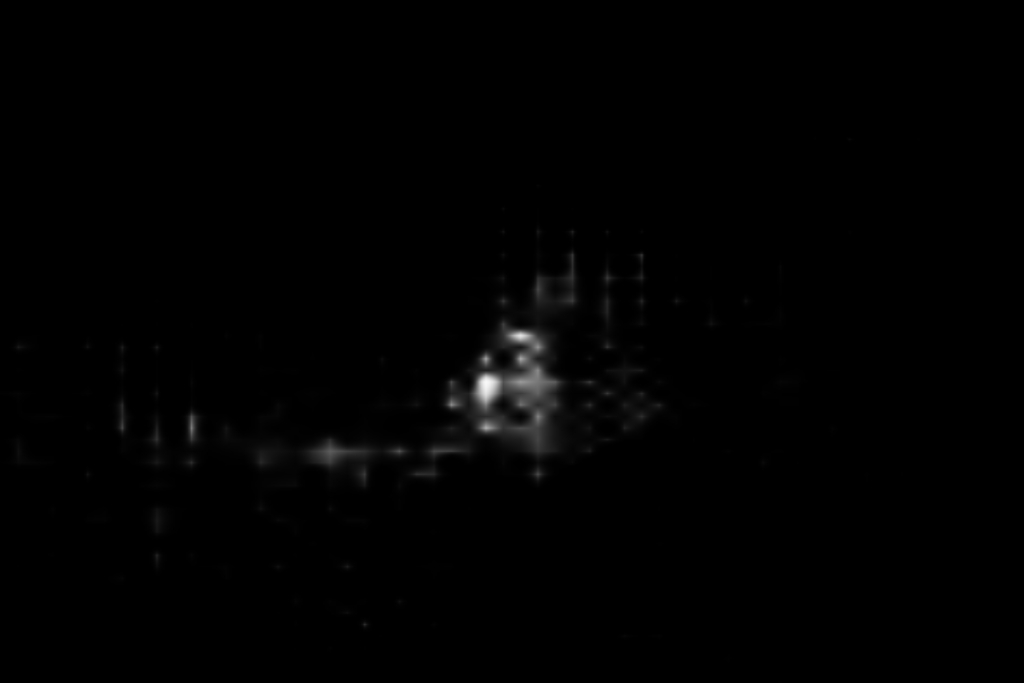}
    \includegraphics[width=0.05\textwidth]{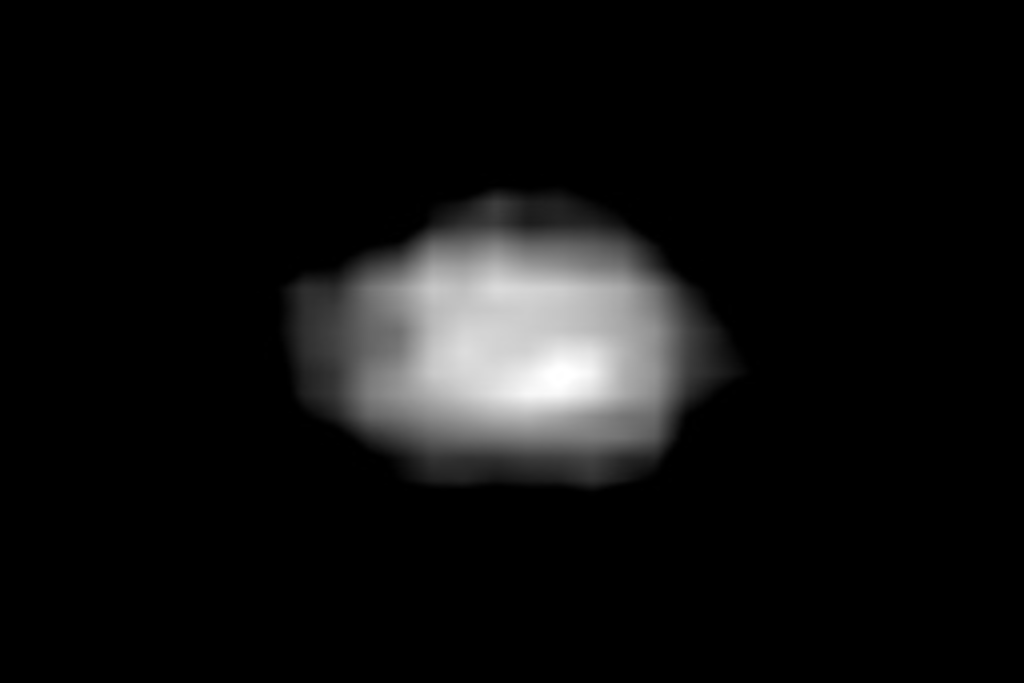}
    \includegraphics[width=0.05\textwidth]{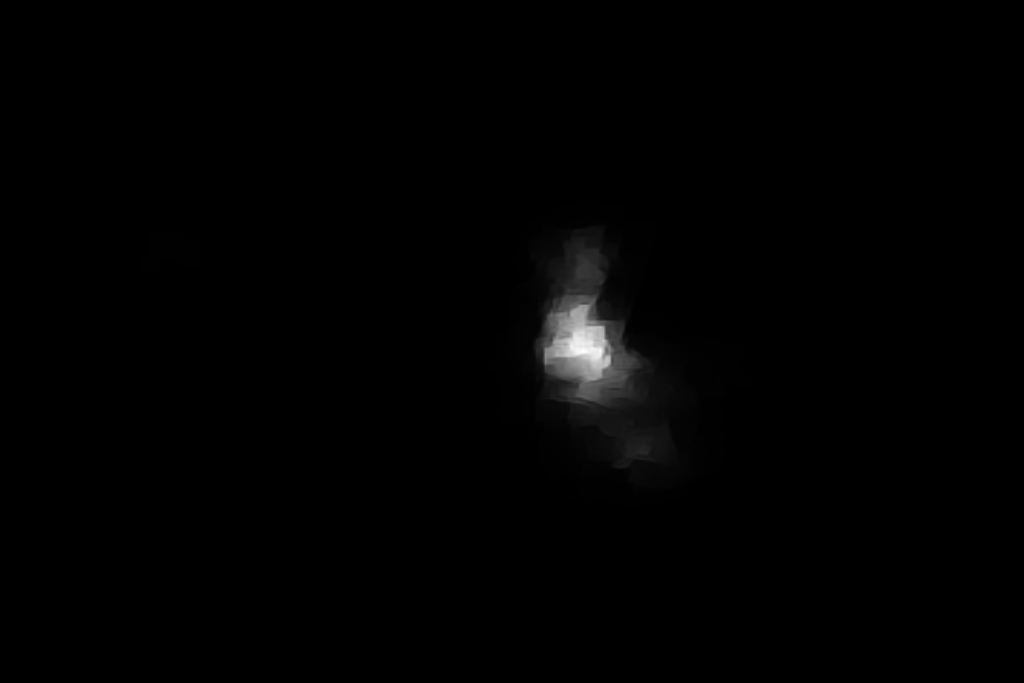}
    \includegraphics[width=0.05\textwidth]{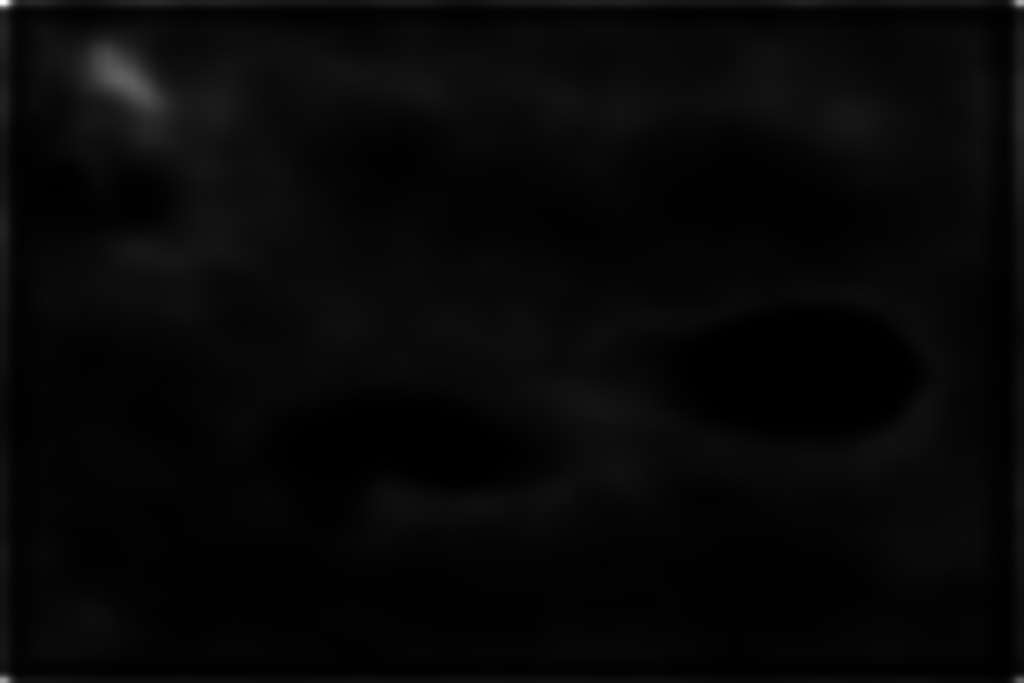}
    \includegraphics[width=0.05\textwidth]{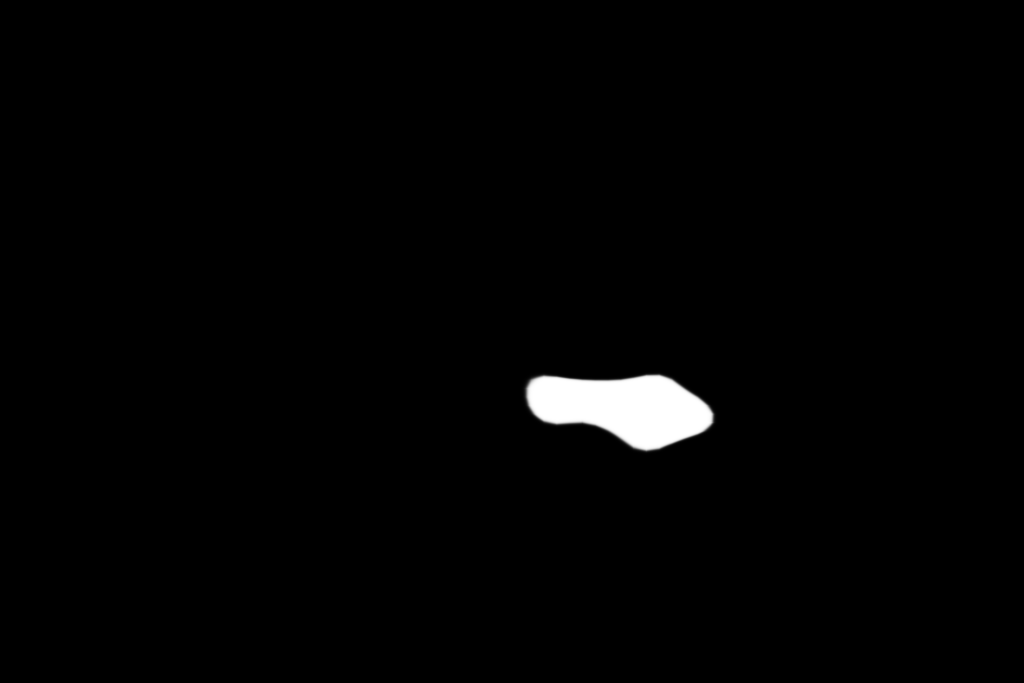}
    \\

    \includegraphics[width=0.05\textwidth]{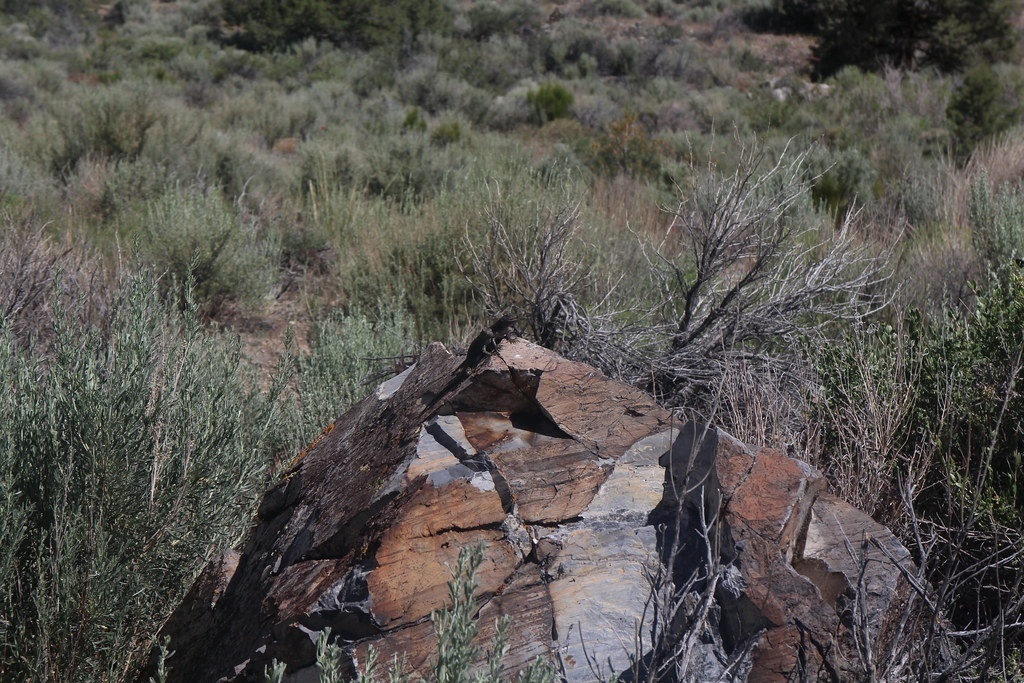}
    \includegraphics[width=0.05\textwidth]{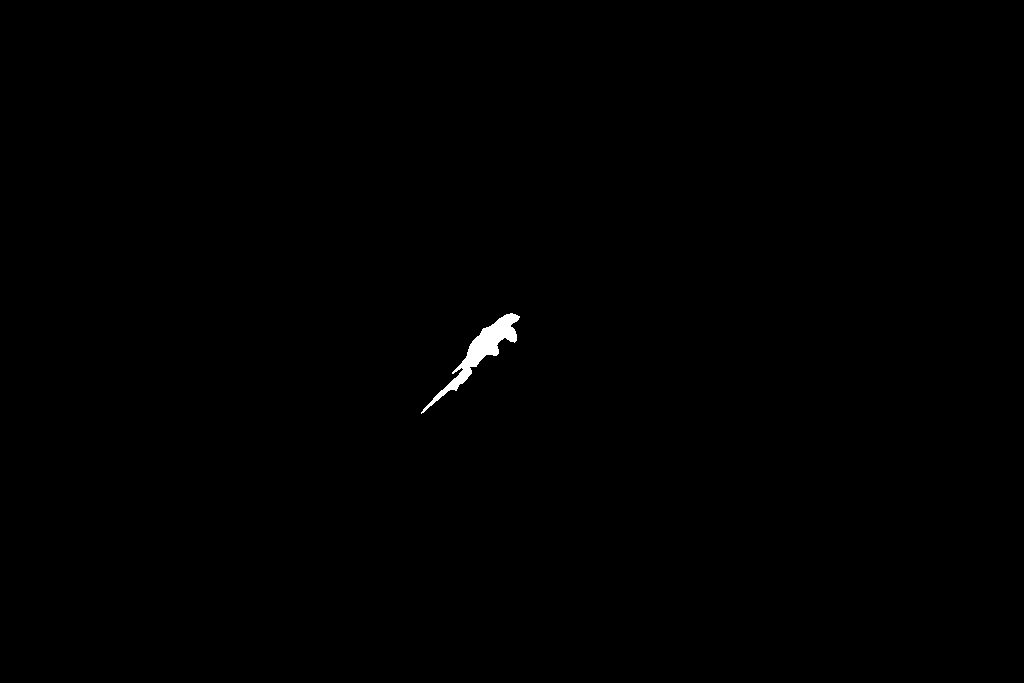}
    \includegraphics[width=0.05\textwidth]{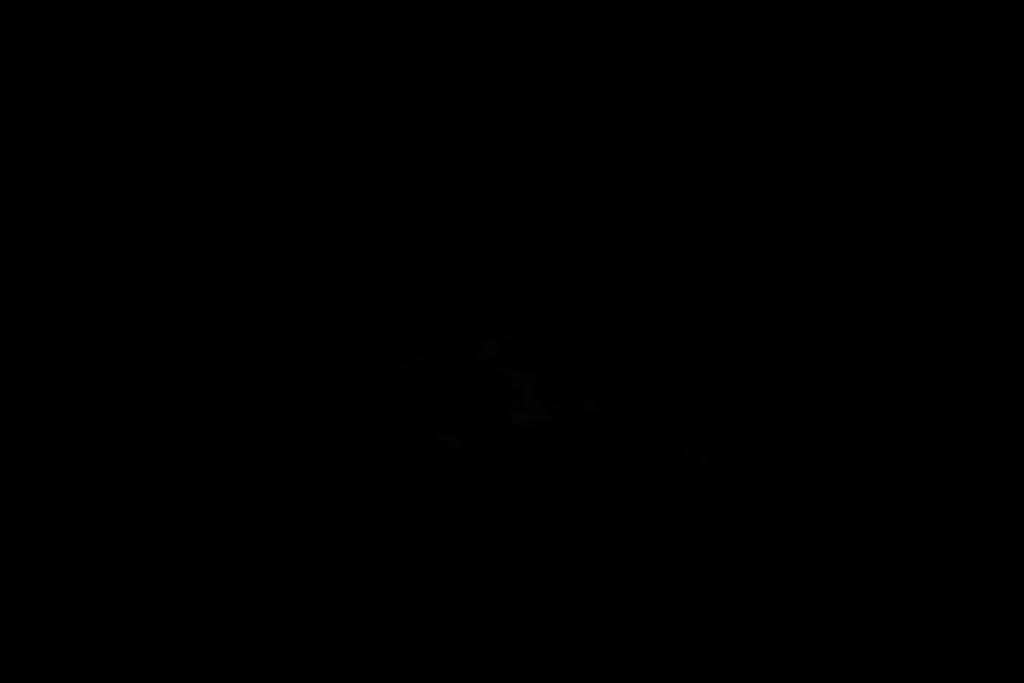}
    \includegraphics[width=0.05\textwidth]{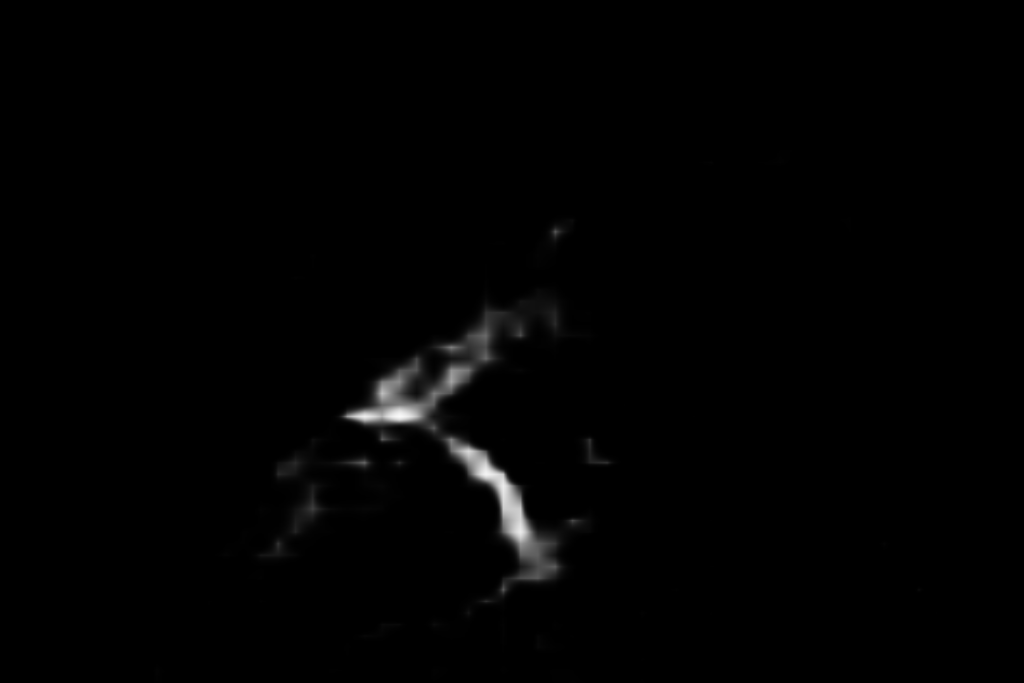}
    \includegraphics[width=0.05\textwidth]{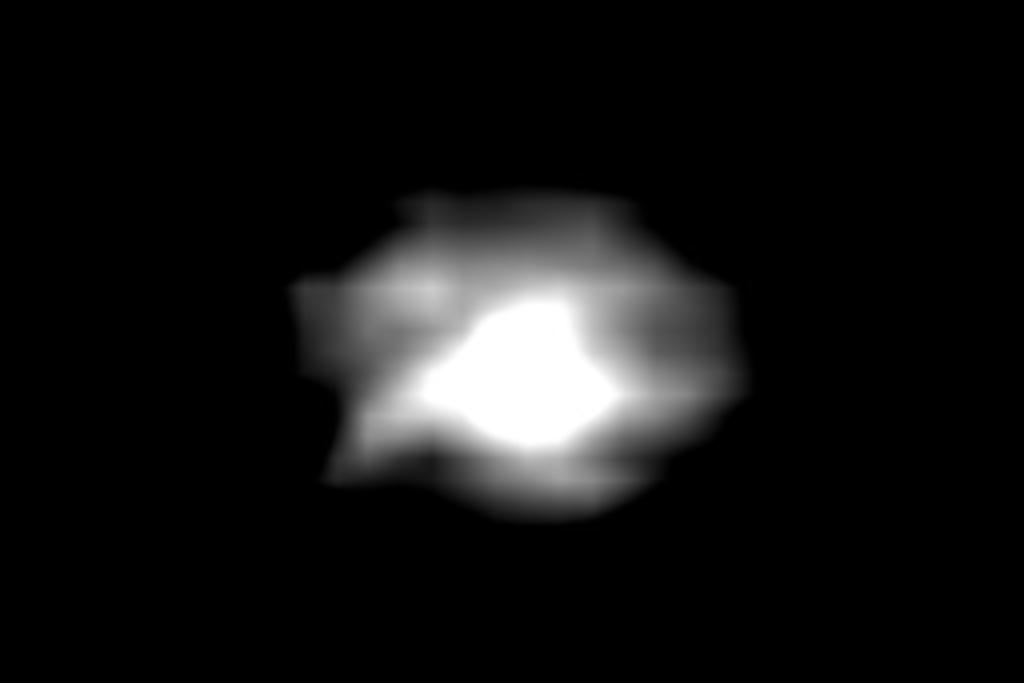}
    \includegraphics[width=0.05\textwidth]{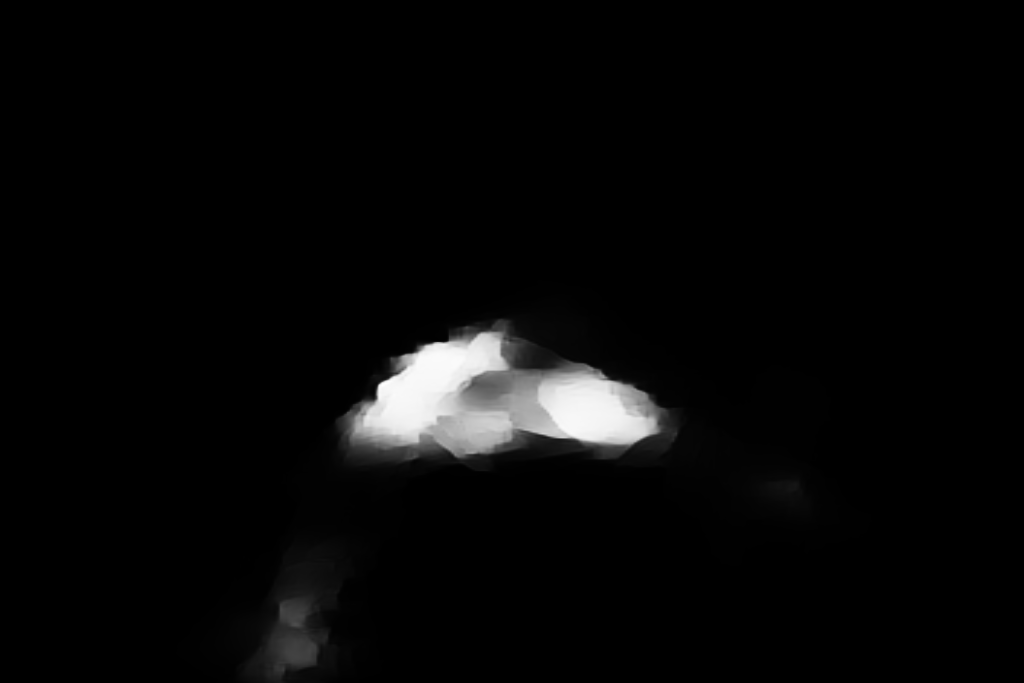}
    \includegraphics[width=0.05\textwidth]{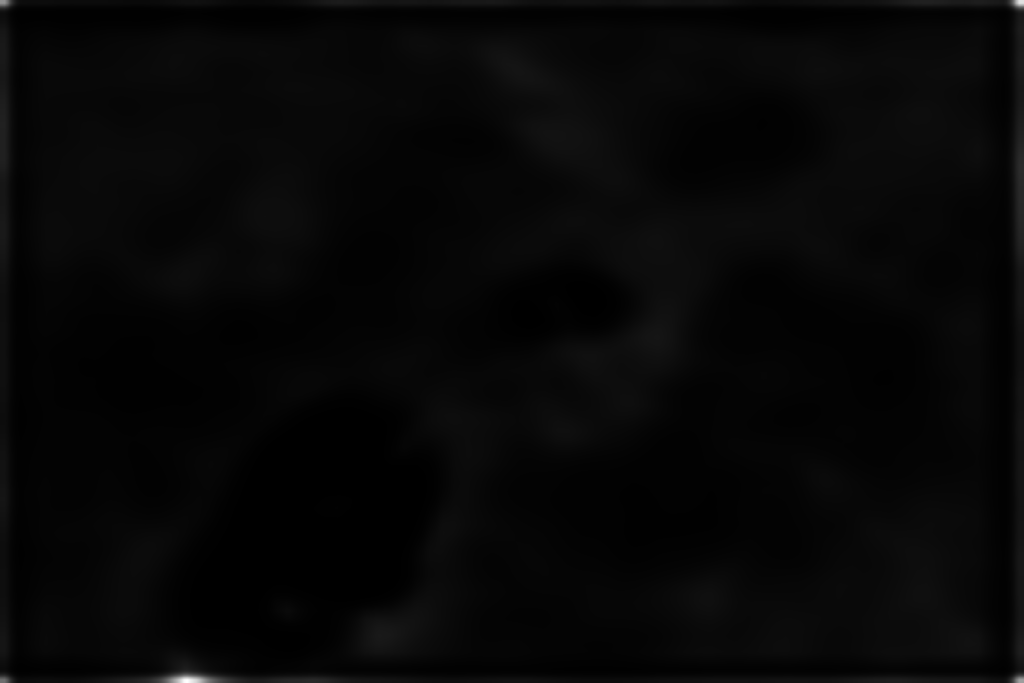}
    \includegraphics[width=0.05\textwidth]{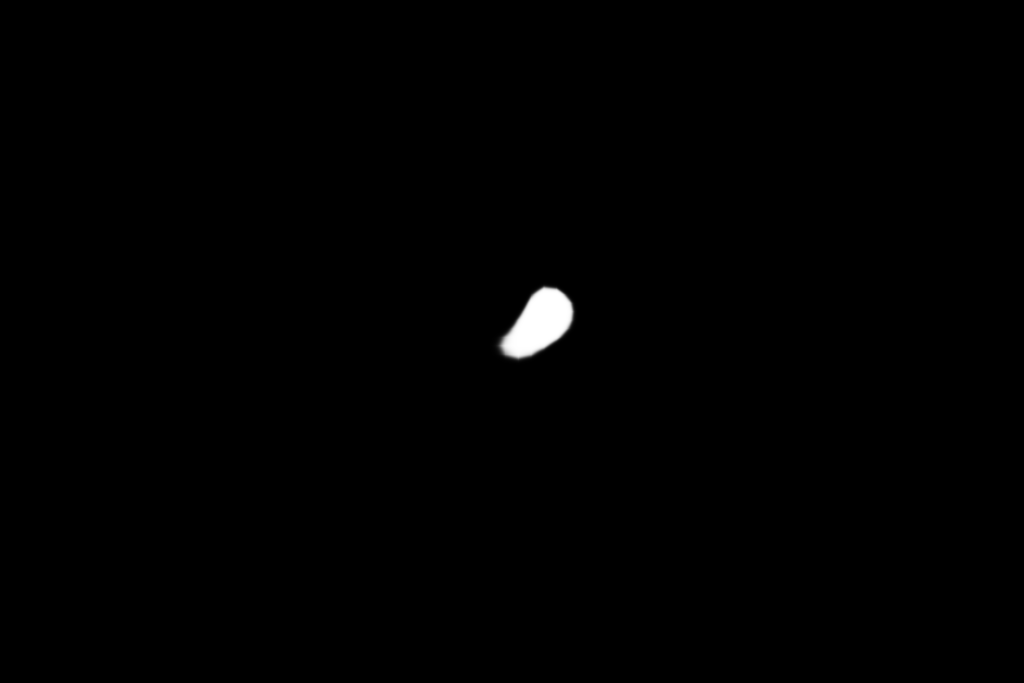}
    \\

    \includegraphics[width=0.05\textwidth]{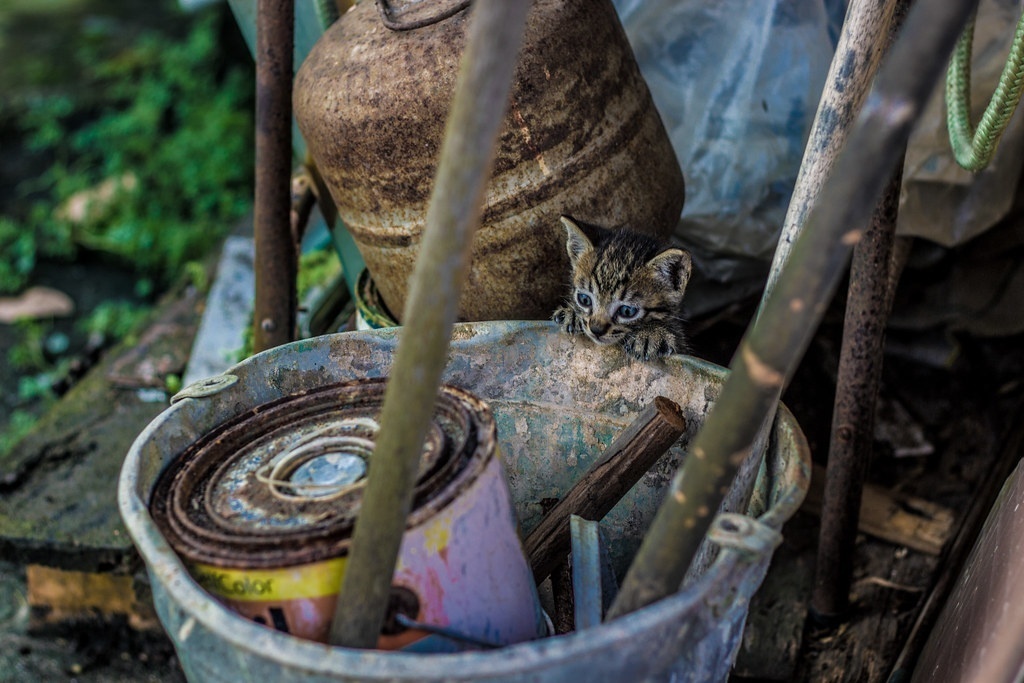}
    \includegraphics[width=0.05\textwidth]{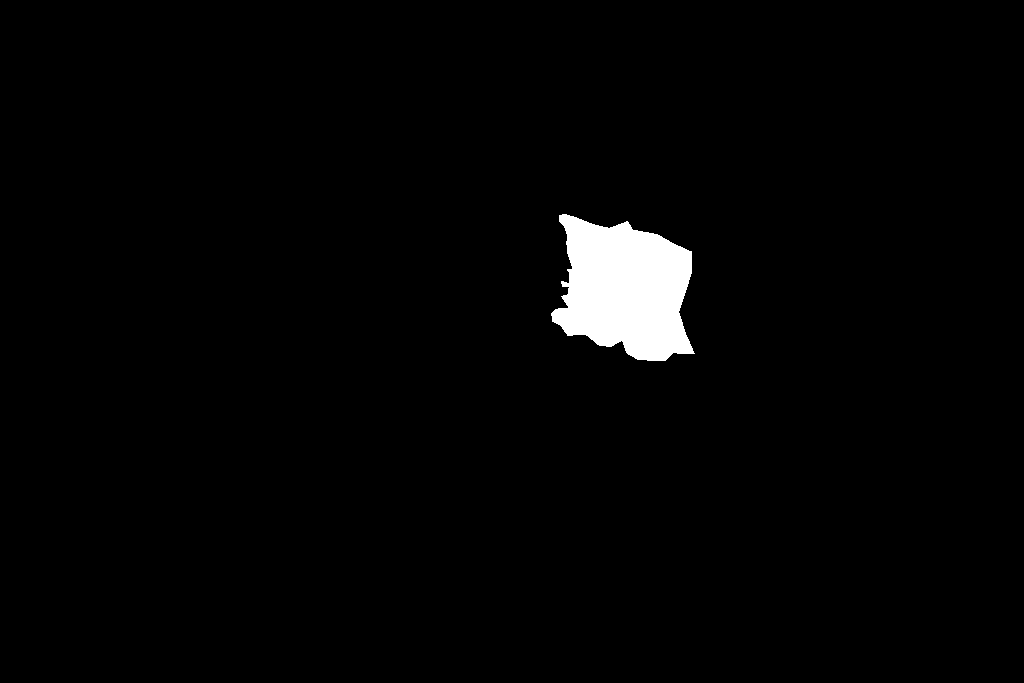}
    \includegraphics[width=0.05\textwidth]{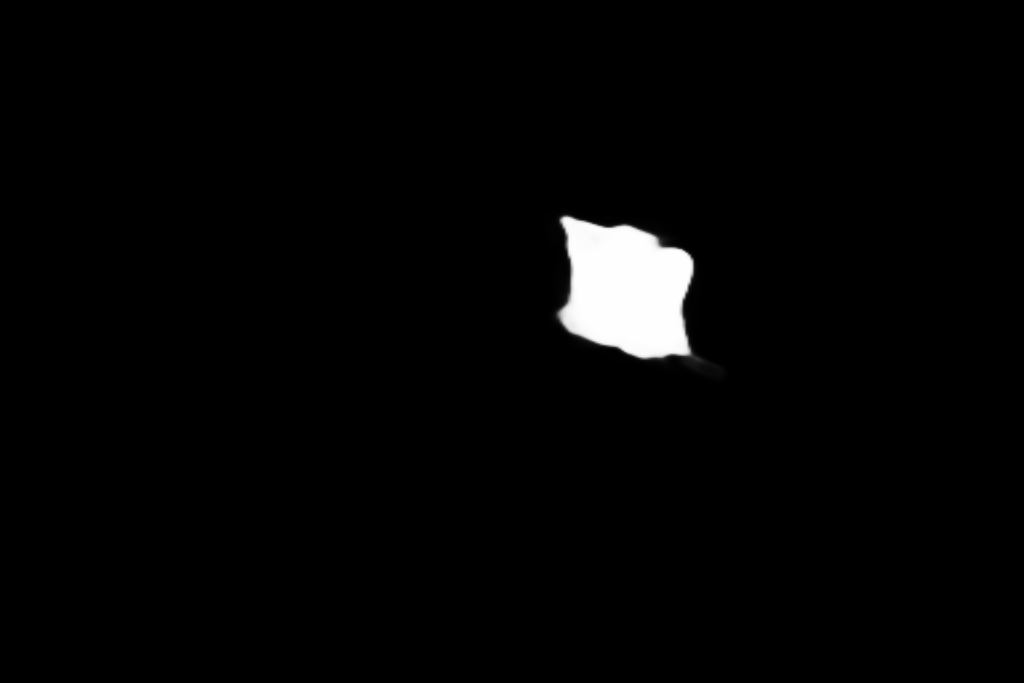}
    \includegraphics[width=0.05\textwidth]{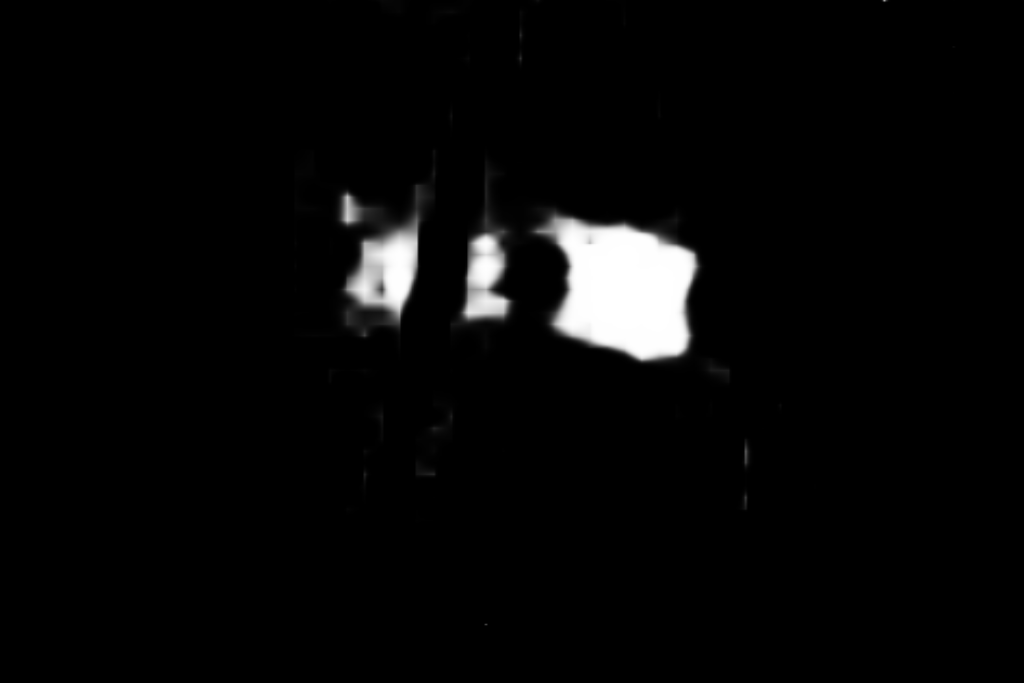}
    \includegraphics[width=0.05\textwidth]{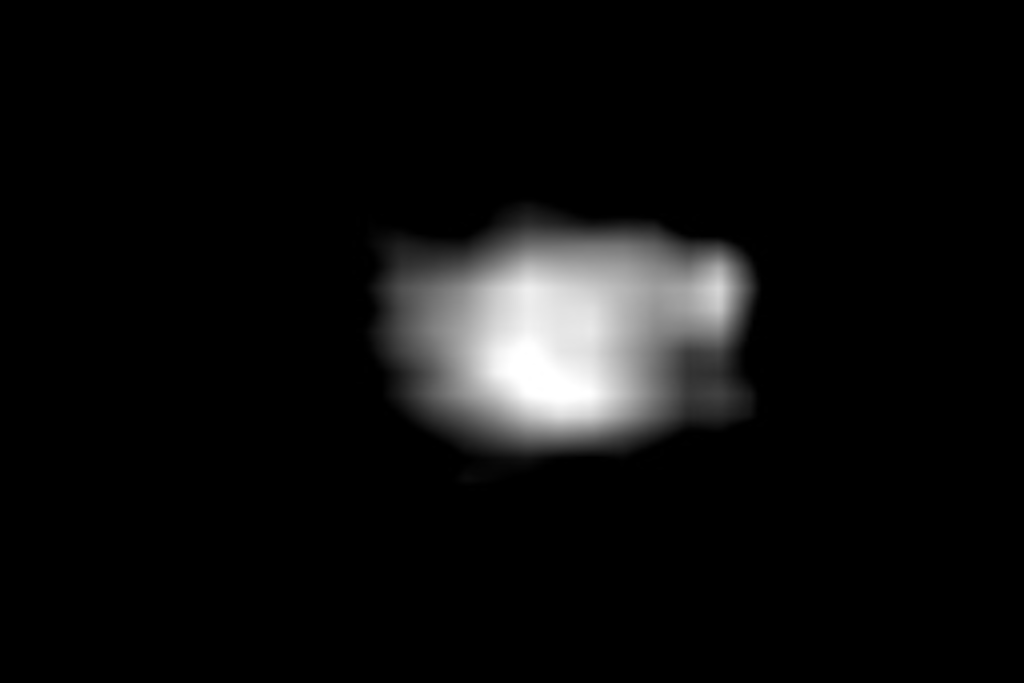}
    \includegraphics[width=0.05\textwidth]{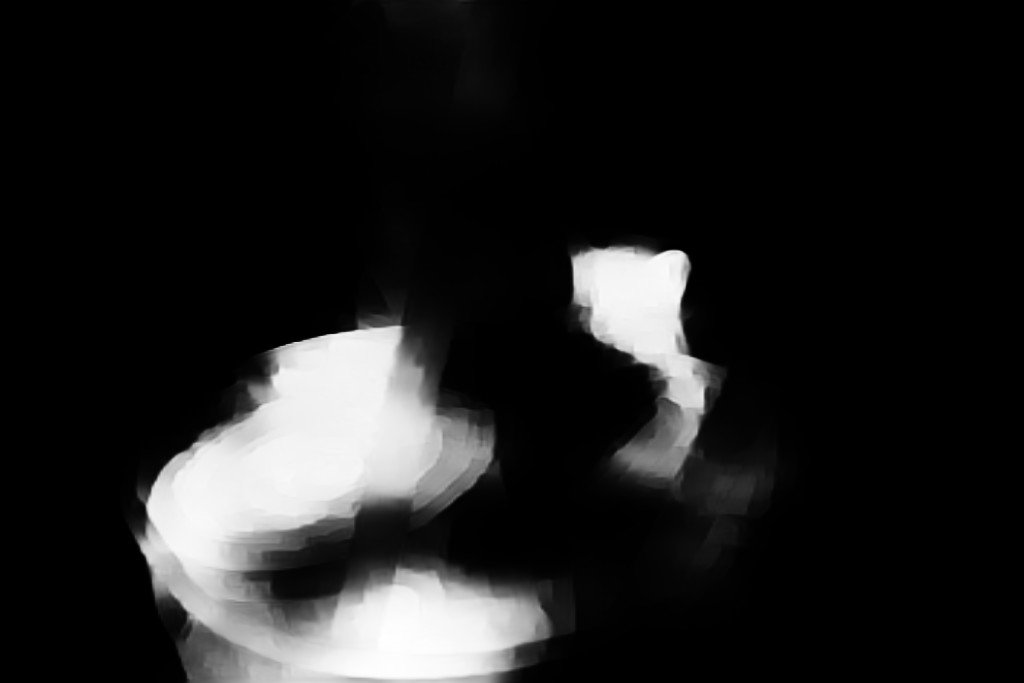}
    \includegraphics[width=0.05\textwidth]{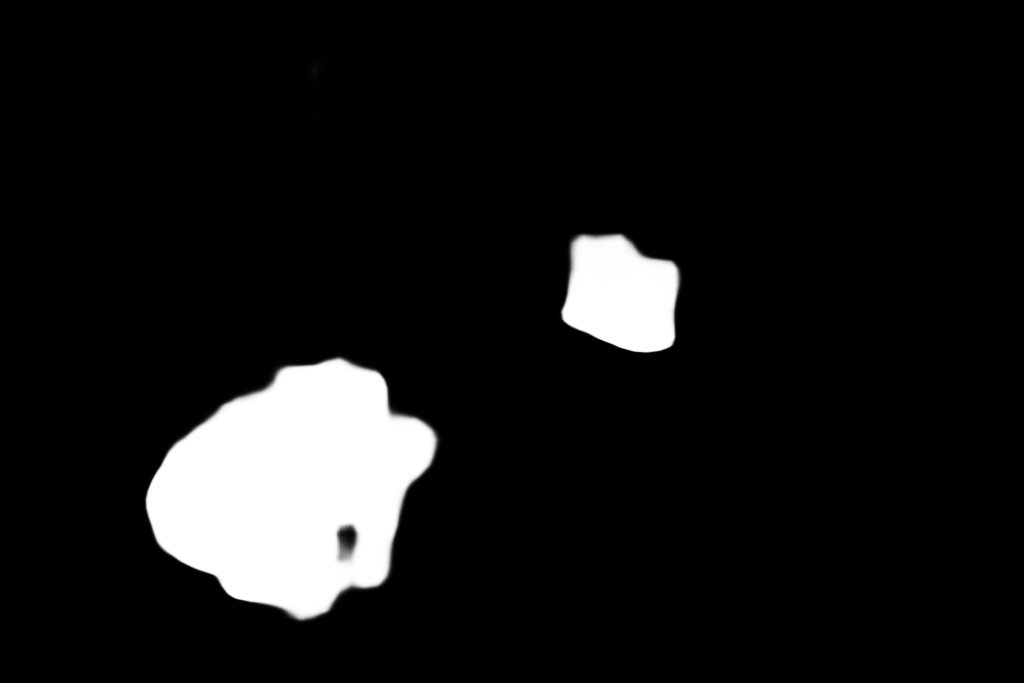}
    \includegraphics[width=0.05\textwidth]{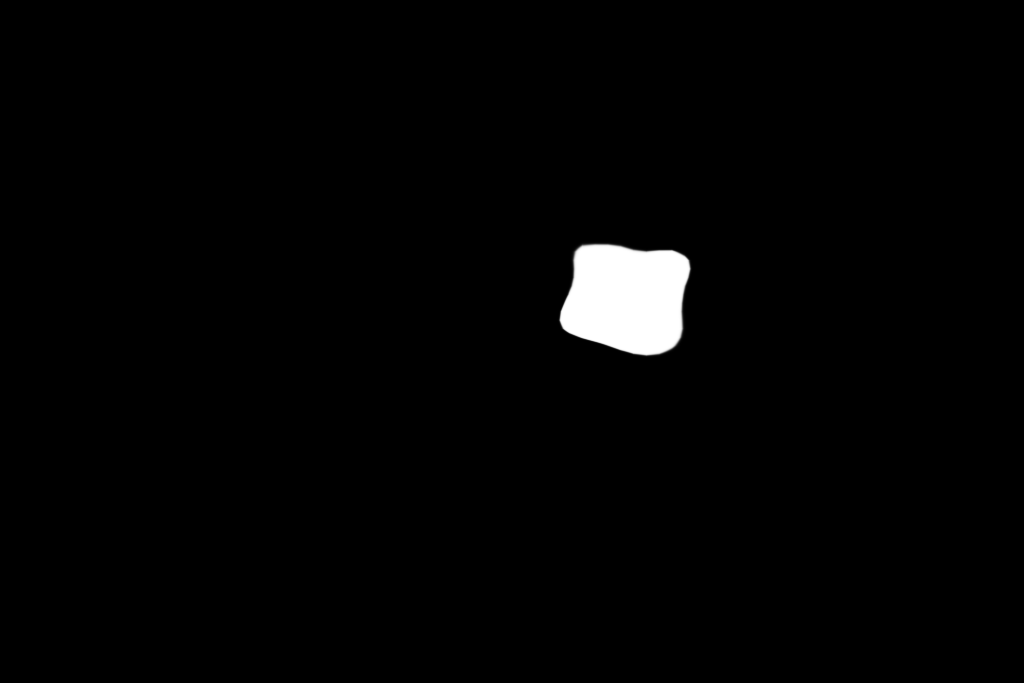}
    \\

    \includegraphics[width=0.05\textwidth]{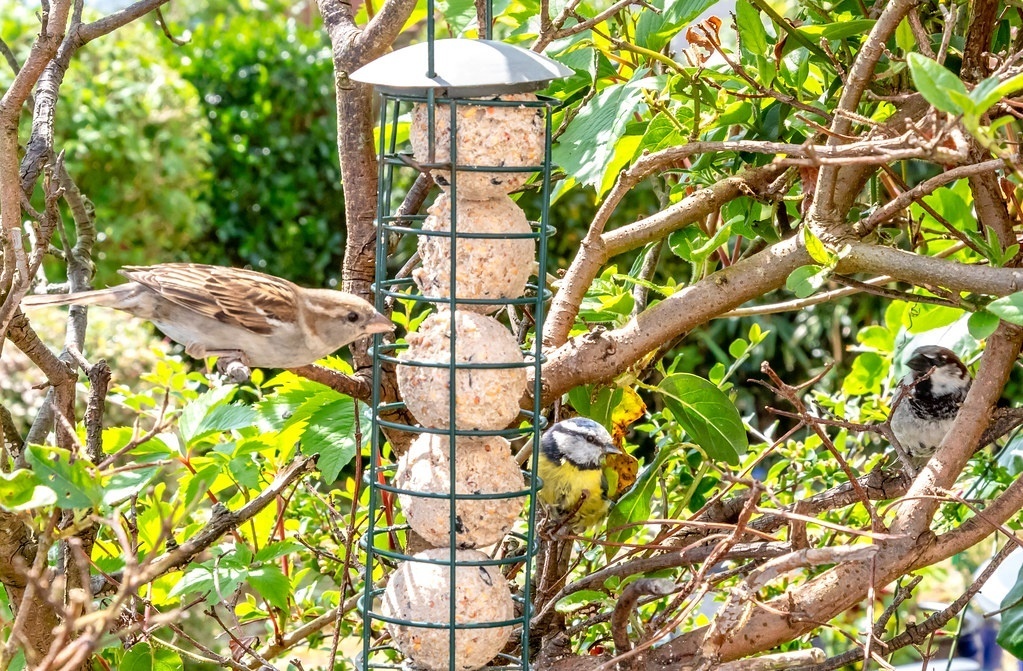}
    \includegraphics[width=0.05\textwidth]{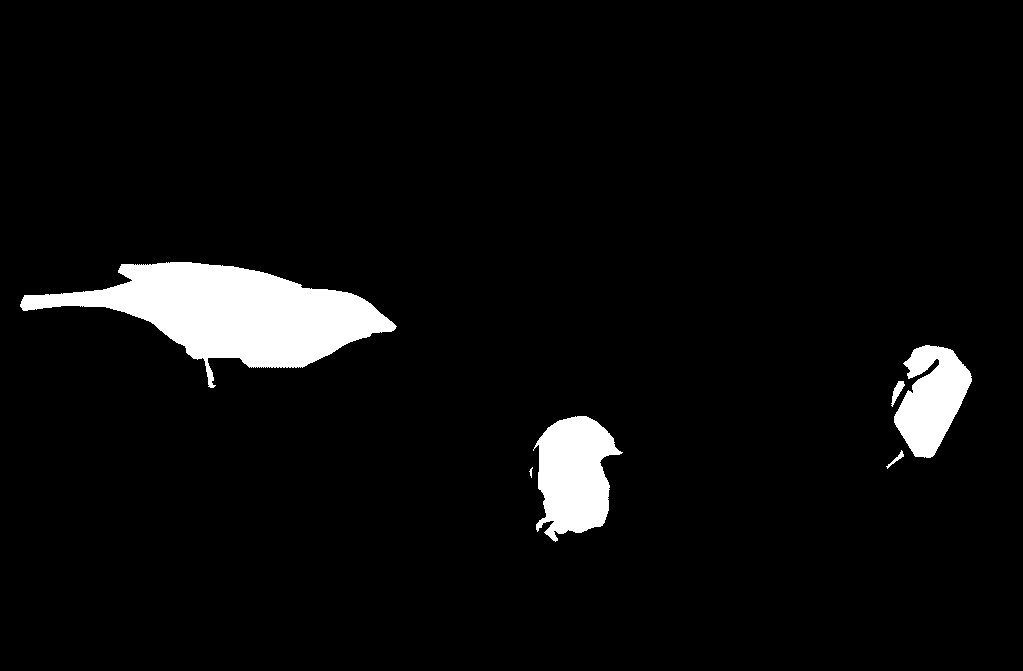}
    \includegraphics[width=0.05\textwidth]{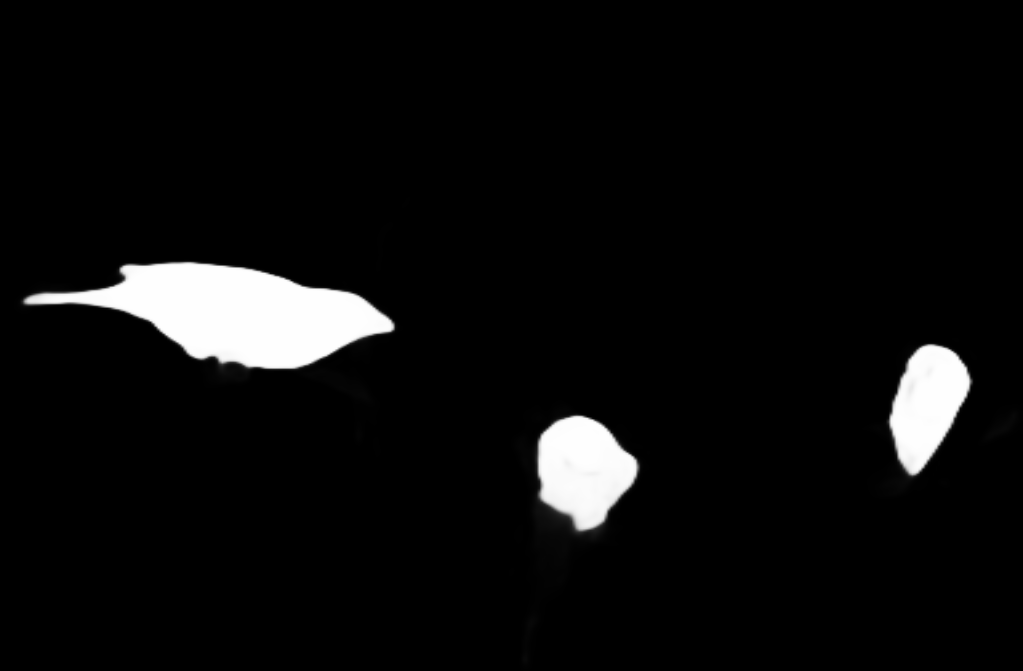} 
    \includegraphics[width=0.05\textwidth]{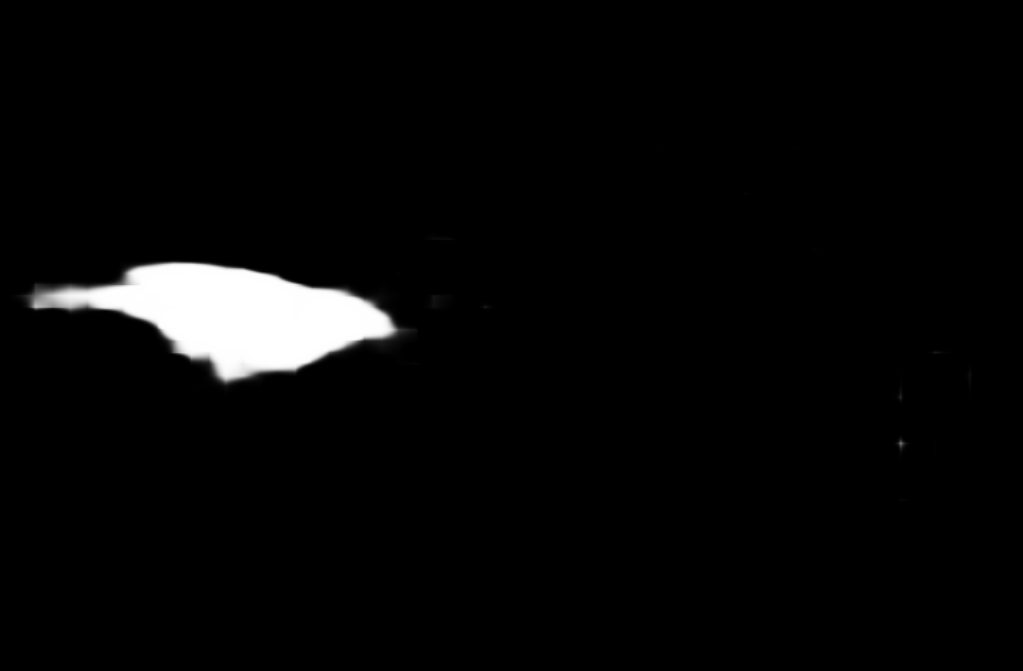}
    \includegraphics[width=0.05\textwidth]{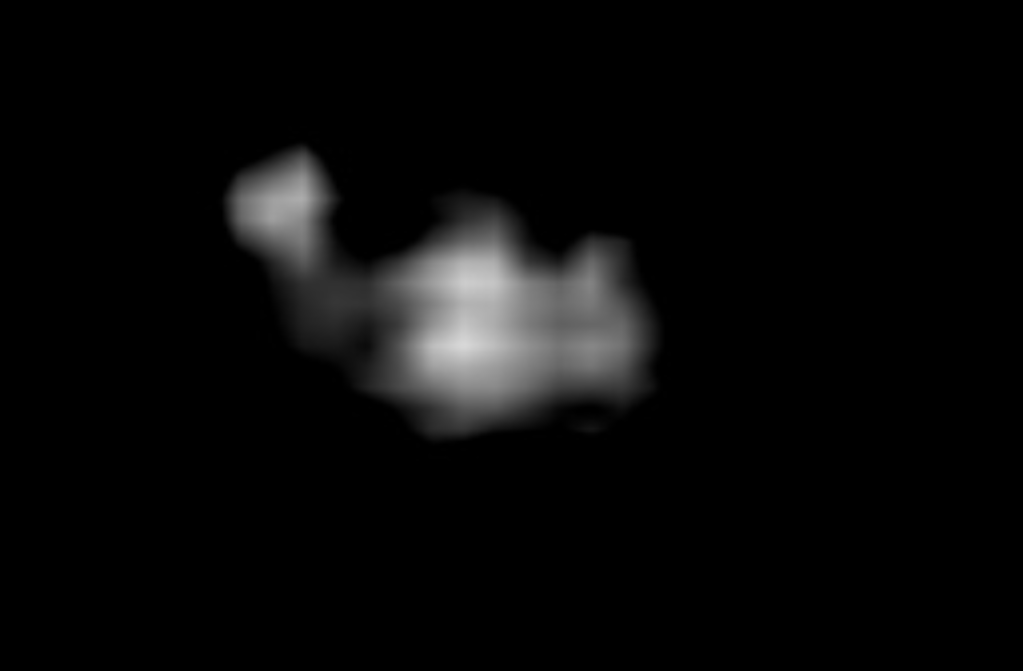}
    \includegraphics[width=0.05\textwidth]{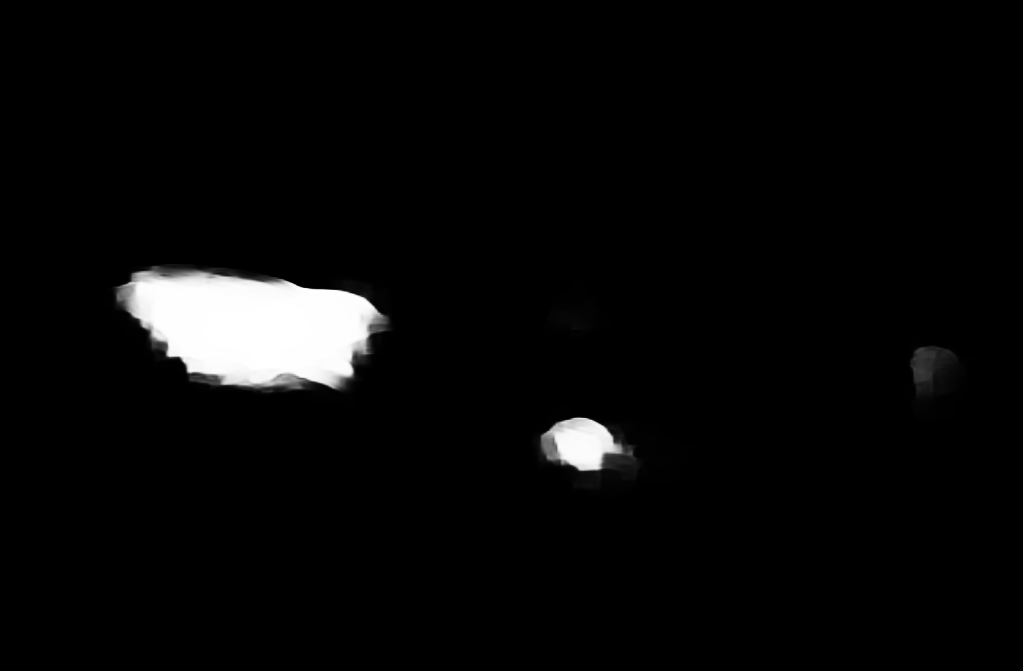}
    \includegraphics[width=0.05\textwidth]{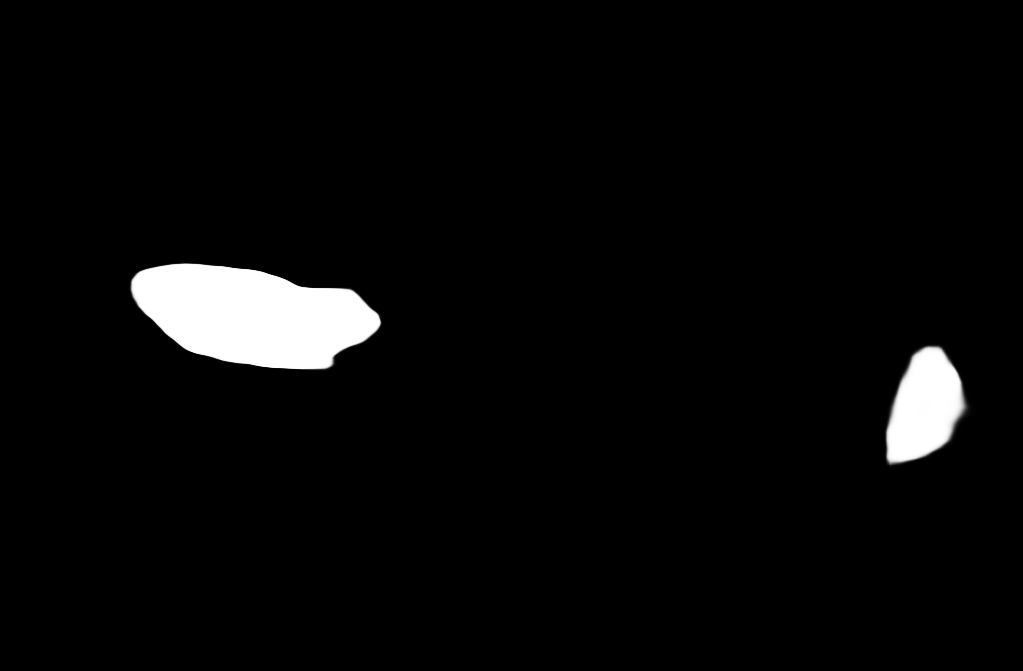}
    \includegraphics[width=0.05\textwidth]{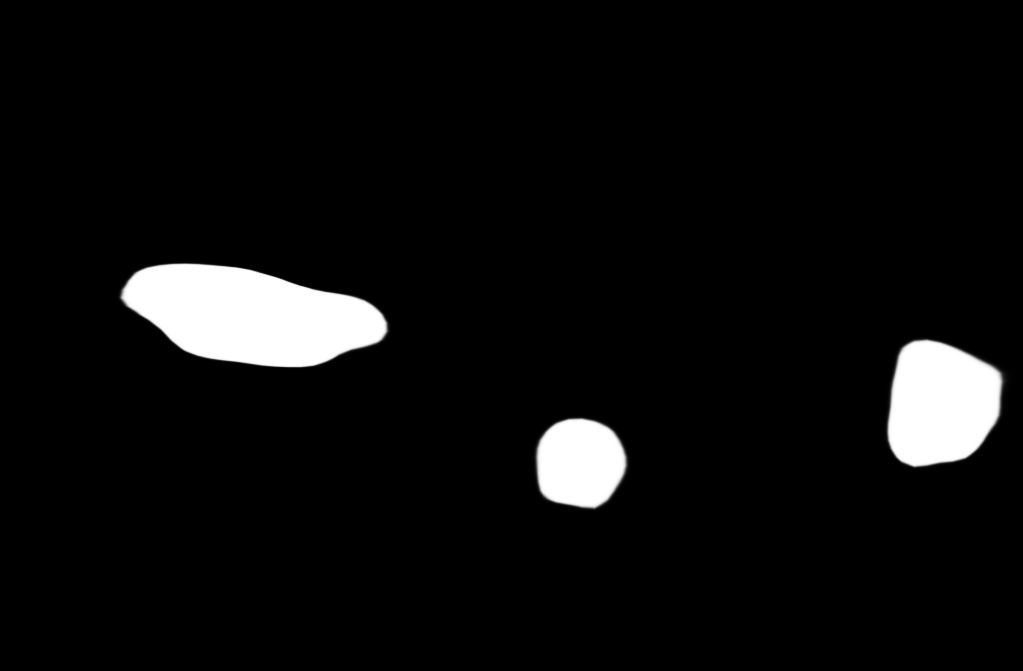}
    \\

    \makebox[0.05\textwidth]{\scriptsize Input}
    \makebox[0.05\textwidth]{\scriptsize GT}
    \makebox[0.05\textwidth]{\scriptsize ZoomNet}
    \makebox[0.05\textwidth]{\scriptsize UGTR}
    \makebox[0.05\textwidth]{\scriptsize DUSD}
    \makebox[0.05\textwidth]{\scriptsize SS}
    \makebox[0.05\textwidth]{\scriptsize SCWSSOD}
    \makebox[0.05\textwidth]{\scriptsize Ours}
    \\
    \caption{Qualitative comparison of our method with state-of-the-arts fully-supervised, unsupervised, and scribble-based weakly-supervised methods in challenging scenarios.}
    \label{fig:comparison}
\end{figure}

\noindent\textbf{Comparison with State-of-the-arts.} As we propose the first weakly-supervised method, we introduce 2 scribble-based weakly and 2 unsupervised SOD \ryn{methods} for comparison. We also provide the results of fully-supervised 8 COD and 12 SOD methods for reference.
Quantitative comparisons are demonstrated in Table~\ref{tab:eval}. Our method performs the best under four metrics on three benchmarks among weakly or unsupervised methods. It achieves an average enhancement of 11.0$\%$ on MAE, 3.2$\%$ on S-measure, 2.5$\%$ on E-measure, and 4.4$\%$ on weighted F-measure than the state-of-the-art method SCWSSOD \cite{yu2021structure}. In addition, it outperforms 7 fully-supervised methods~\cite{luo2017non,liu2018picanet,wu2019cascaded,zhao2019egnet,liu2019simple,wei2020f3net,zhou2020interactive}.
We also find that our method has the largest improvement in CAMO (outperforms nearly all fully-supervised SOD methods and is close to COD methods), which is the most challenging one among all of the 3 COD datasets (worst metric value). This shows that our method is indeed better at discovering hard camouflage objects than others.
Figure \ref{fig:comparison} shows that our method performs well in various challenging scenarios, including high intrinsic similarities (row 1), tiny objects (row 2), complex backgrounds (row 3), and multiple objects (row 4).

\begin{figure}[h] \centering
    \includegraphics[width=0.06\textwidth]{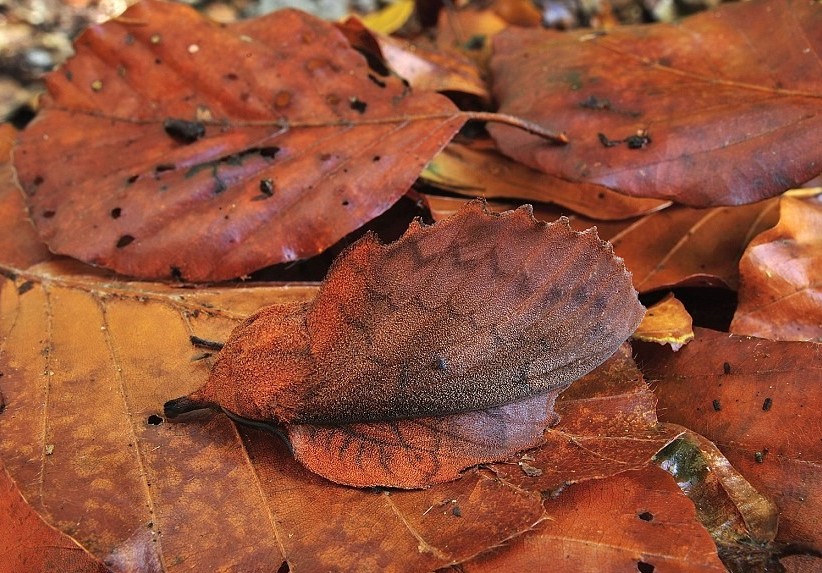}
    \includegraphics[width=0.06\textwidth]{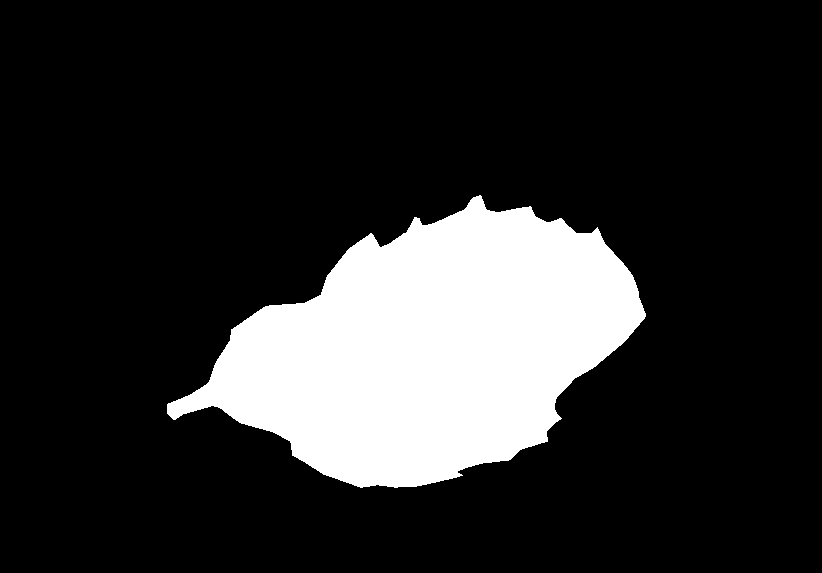}
    \includegraphics[width=0.06\textwidth]{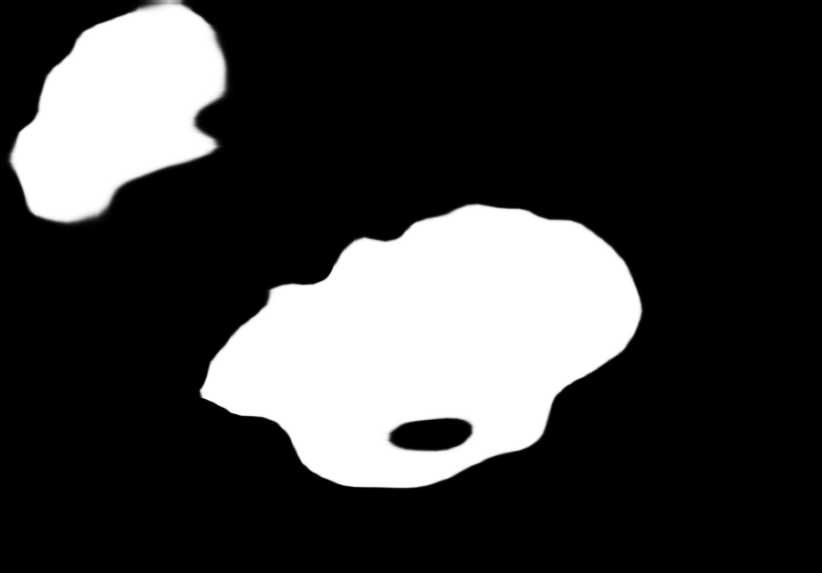}
    \includegraphics[width=0.06\textwidth]{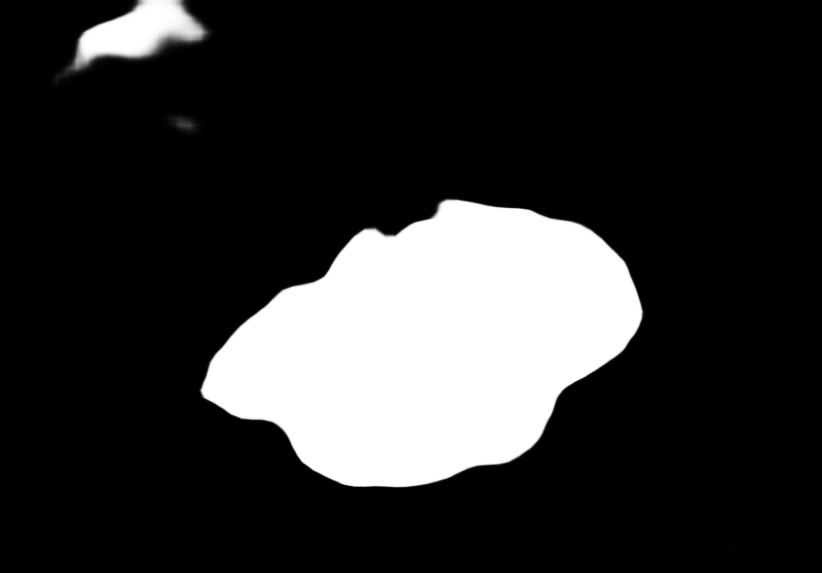}
    \includegraphics[width=0.06\textwidth]{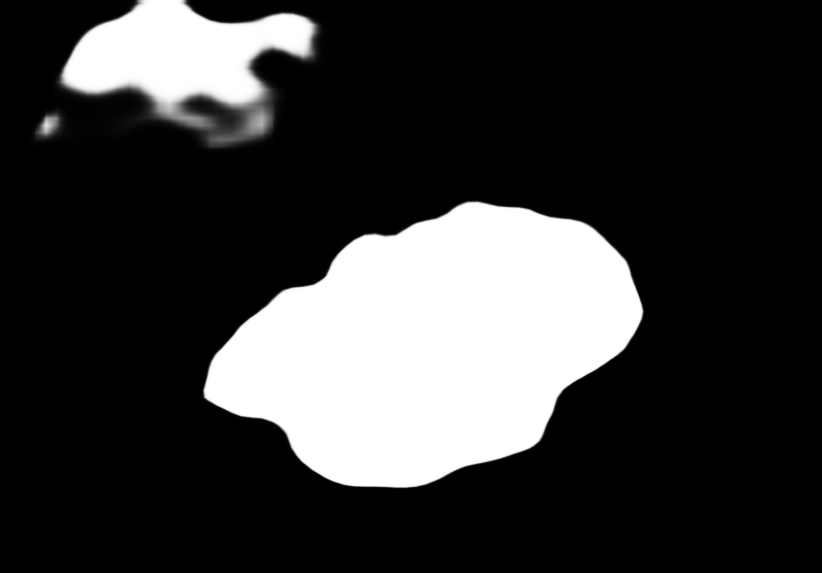}
    \includegraphics[width=0.06\textwidth]{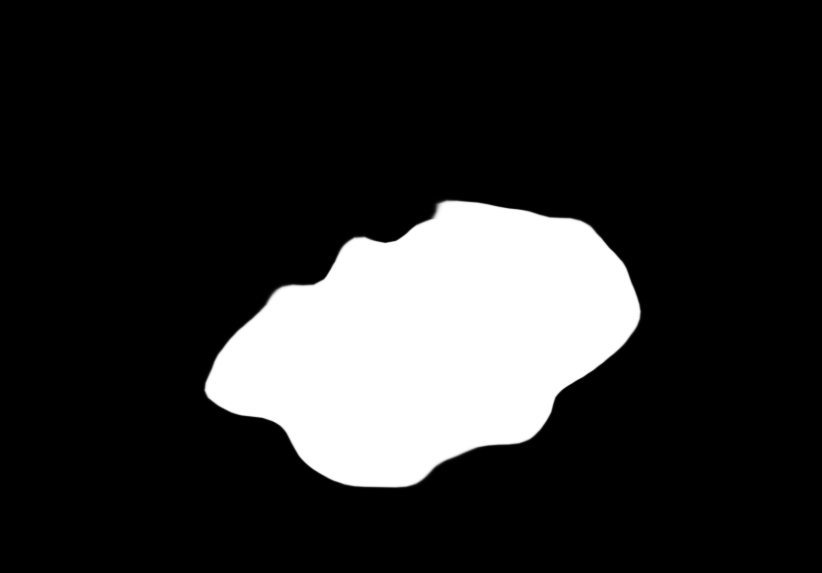}
    \includegraphics[width=0.06\textwidth]{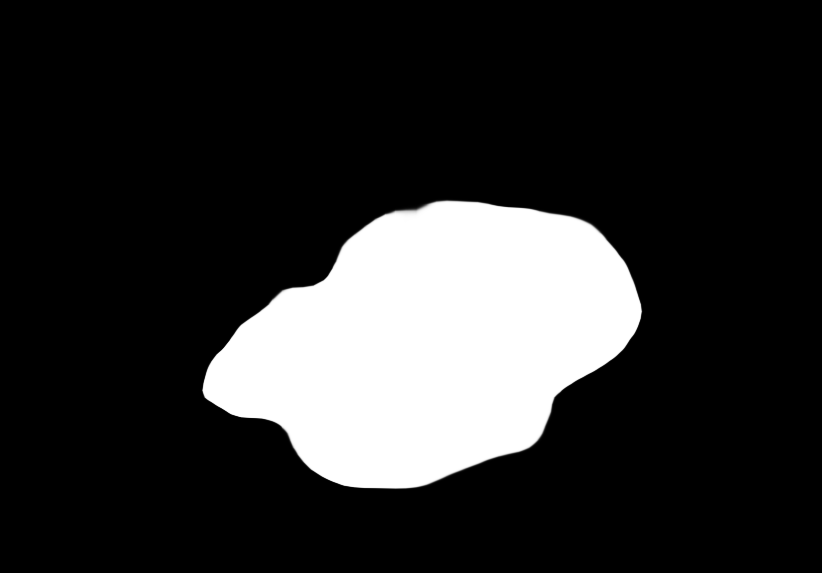}
    \\
    \makebox[0.06\textwidth]{\scriptsize Input}
    \makebox[0.06\textwidth]{\scriptsize GT}
    \makebox[0.06\textwidth]{\scriptsize I}
    \makebox[0.06\textwidth]{\scriptsize II}
    \makebox[0.06\textwidth]{\scriptsize III}
    \makebox[0.06\textwidth]{\scriptsize IV}
    \makebox[0.06\textwidth]{\scriptsize Ours}
    \caption{A visual example of the component ablation study.} 
    \label{fig:componentAblation}
\end{figure}

\noindent\textbf{Ablation Studies on Modules.} 
To verify the effectiveness of our modules, we conduct ablation studies on challenging dataset CAMO \cite{le2019anabranch}, where the methods obtain the worst scores according to Table \ref{tab:eval}. Table \ref{tab:componentAblation} shows only using a backbone (BB) performs worst (\textit{i.e.,} Ablation I), while adding LCC or LSR improves performances on different metrics. As shown in Figure \ref{fig:componentAblation}, LCC finds potential camouflaged regions with low-level contrasts, but it may be confused by complex background (\textit{e.g.,} many distinct leaves). Meanwhile, LSR analyzes logical semantic relations between different parts, but it may segment inaccurate boundaries. When LCC and LSR cooperate to detect camouflaged objects (Ours), the performance is enhanced dramatically from the single module usage (III, IV). It shows the effectiveness of CRNet design.
\begin{table}[t]
\centering
\caption{The ablation study results of components on CAMO~\cite{le2019anabranch}. 
}
\scalebox{0.7}{
\begin{tabular}{ccccccccc}
\hline
\multicolumn{1}{c|}{Methods}                  &     BB             & AGE                  & LCC                  & \multicolumn{1}{l|}{LSR}                  & \multicolumn{1}{c}{MAE$\downarrow$}                           & \multicolumn{1}{c}{S$_{m}\uparrow$}                            & \multicolumn{1}{c}{E$_{m}\uparrow$}                            & \multicolumn{1}{c}{F$_{\beta}^{w}\uparrow$}                           \\ \hline
\multicolumn{1}{c|}{Ablation I}                  & $\surd$ &                      &                      & \multicolumn{1}{l|}{}                     & 0.104                     & 0.701                     & 0.774                     & 0.598                     \\
\multicolumn{1}{c|}{Ablation II}                  &$\surd$ & $\surd$ &                      & \multicolumn{1}{l|}{}                     & 0.100                                             & 0.716                                             & 0.799                                             & 0.615                                             \\
\multicolumn{1}{c|}{Ablation III}                  &$\surd$ & $\surd$ & $\surd$ & \multicolumn{1}{l|}{}                     & 0.099                                             & 0.721                                             & 0.806                                             & 0.626                                             \\
\multicolumn{1}{c|}{Ablation IV}                  &$\surd$ & $\surd$ &                      & \multicolumn{1}{l|}{$\surd$} & 0.098                                             & 0.713                                             & 0.783                                             & 0.612                                             \\
\multicolumn{1}{c|}{Ours}                  &$\surd$ & $\surd$ & $\surd$ & \multicolumn{1}{l|}{$\surd$} & 0.092 & 0.735 & 0.815 & 0.641 \\ \hline
                     &                      &                      &                                           & \multicolumn{1}{c}{}      & \multicolumn{1}{c}{}      & \multicolumn{1}{c}{}      & \multicolumn{1}{c}{}     

\label{tab:componentAblation}
\end{tabular}
}
\end{table}

\noindent\textbf{Ablation Studies on Loss Functions.}\label{sssec:ablation on loss} 
A detailed ablation study for loss functions is also conducted in Table~\ref{tab:lossAblation}. 
We first explore various combinations of transformation operations in cross-view consistency. It is shown that flipping, translating, and cropping upgrade the performance significantly. The second group, the ablation of consistency loss, shows improvements on all metrics except MAE. This indicates the benefit of the proposed consistency mechanism. The third group ablates our feature-guided loss. The final group is the overall component ablation of consistency loss and feature-guided loss. We see that both losses provide tremendous improvement in the test dataset.

\begin{table}[htb]
\centering
\caption{The ablation study for our loss functions on CAMO~\cite{le2019anabranch}. Groups correspond to ablations on transformations in cross-view consistency, on consistency loss, on feature loss, and on all losses. Here, pce stands for partial cross-entropy; \ryn{ft and cs stand} for feature-guided loss and consistency loss (cs=cv+iv, ft=ca+ss); \ryn{cv and iv stand} for cross-view and inside-view consistency loss; cv' means cross-view consistency without reliability bias; cv($\cdot$) specifies the transforms used in computing cv; R,F,T,C are resizing, flipping, translation and cropping 
.
}
\scalebox{0.85}{
\begin{tabular}{ll|rrrr}
\hline
\multicolumn{1}{c}{Basic setting} & \multicolumn{1}{c|}{Loss} & \multicolumn{1}{c}{MAE$\downarrow$} & \multicolumn{1}{c}{S$_{m}\uparrow$} & \multicolumn{1}{c}{E$_{m}\uparrow$} & \multicolumn{1}{c}{F$_{\beta}^{w}\uparrow$} 
\\ \hline
w/ pce                             & Baseline                  & 0.215                               & 0.612                               & 0.633                               & 0.387
\\ \hline
w/ ft, iv                       & w/o cv                    & 0.105                               & 0.721                               & 0.786                               & 0.600
\\
                                  & w/ cv(R)                  & 0.097                               & 0.727                               & 0.807                               & 0.629                                     
\\
                                  & w/ cv(R, F)               & 0.094                               & 0.730                               & 0.812                               & 0.638                                                                
\\
                                  & w/ cv(R, F, T)            & 0.094                               & 0.730                               & 0.808                               & 0.637                         
\\
                                  & w/ cv(R, F, T, C)         & 0.092                               & 0.735                               & 0.815                               & 0.641
\\ \hline
w/ ft                              & w/ cv'                     & 0.095                               & 0.723                               & 0.801                               & 0.624
\\
                                  & w/ cv                  & 0.095                               & 0.726                               & 0.804                               & 0.632
\\
\multicolumn{1}{r}{}              & w/ cs              & 0.092                               & 0.735                               & 0.815                               & 0.641
\\ \hline
w/ cs                             & w/ ca                     & 0.095                               & 0.727                               & 0.807                               & 0.631
\\
                                  & w/ ft                 & 0.092                               & 0.735                               & 0.815                               & 0.641
\\ \hline
w/ pce                             & w/ cs                     & 0.096                               & 0.731                               & 0.821                               & 0.641
\\
& w/ ft                     & 0.107                               & 0.720                               & 0.785                               & 0.592  
\\
& w/ cs, ft                 & 0.092                               & 0.735                               & 0.815                               & 0.641                                     
\\ \hline
\label{tab:lossAblation}
\end{tabular}
}
\end{table}

\section{Conclusion}
In this paper, we propose the first weakly-supervised COD dataset with scribble annotation, which \ryn{takes} \textasciitilde 10 seconds \ryn{to label an} image (360 times faster than pixel-wise annotation).
To overcome the weaknesses of current weakly-supervised learning and their application to COD, we propose a novel framework consisting of two loss functions and a novel network: a consistency loss, including consistency inside and cross images, regulates the model to have coherent predictions, and incline them to more reliable ones; a feature-guided loss locates the "hidden" foreground by comparing both manually computed visual features and learned semantic features of each pixel. The proposed network learns low-level contrast to expand scribbles to wider potential regions first and then analyzes logical semantic relation information to determine the real foreground and background. Experimental results show our method outperforms unsupervised and weakly-supervised state-of-the-arts with improvement, and is even competitive with the fully-supervised methods.

\bibliography{egbib}

\end{document}


\maketitle

\section{Related Work}
\textbf{Scribble-based Weakly-Supervised Learning.}
Scribble-based learning has been widely used in computer vision tasks such as semantic segmentation~\cite{lee2021railroad_sem, pan2021scribble} and salient object detection~\cite{zhang2020weakly, yu2021structure}.  
\cite{zhang2020weakly} adopts a multi-stage framework with an auxiliary edge detection module for scribble learning. 
\cite{yu2021structure} proposes a simple yet efficient scribble-based SOD framework with a local coherence loss using color features to propagate scribble regions, and saliency structure consistency loss to predict consistent maps. 
However, semantic models focus on categories that pixels belong to, which could be up to tens or hundreds, while the COD task only cares about separating foreground and background objects.
\cite{zhang2020weakly} needs to use an extra edge detector and perform training in an iterative way, and the consistency constraints used in \cite{yu2021structure} are not fully exploited, and its local coherence in camouflage data can easily mislead the model, since camouflage objects are mostly visually ``unnoticeable'' (\textit{e.g.,} tiny, occluded, similar appearance with backgrounds).

Hence, these scribble-learning methods for SOD are generally not suitable for 
COD scribble-based learning.

\subsection{Weakly Supervised Salient Object Detection}
Most of the weakly supervised learning methods in salient object detection utilized two types of labels, namely image-level and point-level labels. Methods based on image-level labels try to attain pseudo labels using techniques like CAM \cite{zhou2016learning} and then build the model on it \cite{li2018weakly, zeng2019multi}. Although image-level annotation is abundant and out-of-the-box, it suffers from producing accurate maps because CAM only distinguishes the most distinctive part of a class and other hand-crafted like methods cannot exploit rich information of data. 

On the other hand, methods with point-level labels can train models in either a similar two-stage \cite{vernaza2017learning, piao2022noise} or one-stage fashion \cite{gao2022weakly, yu2021structure, zhang2020weakly}. One stage approaches train the model directly from partial annotations and regularization under inductive bias. It has become popular since the appearance of gated CRF loss \cite{obukhov2019gated}, which has enabled one-stage methods to perform comparably to two-stage's, for its nature of simplicity and efficiency. For point-level labels, scribble is preferred since it is nearly as easy to generate as single-point annotations and produces better results.

\section{S-COD Dataset}
\begin{figure*}[!h] 
    \centering
    
    \includegraphics[width=0.1\textwidth]{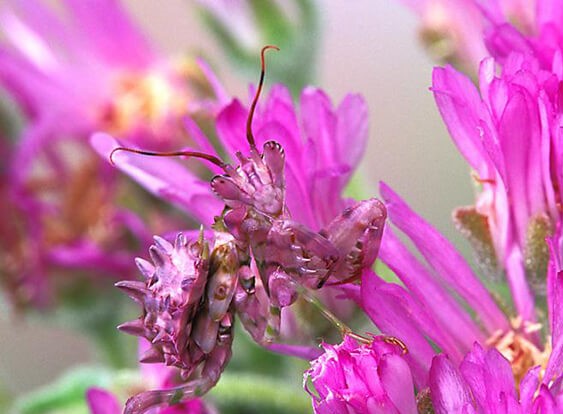}
    \includegraphics[width=0.1\textwidth]{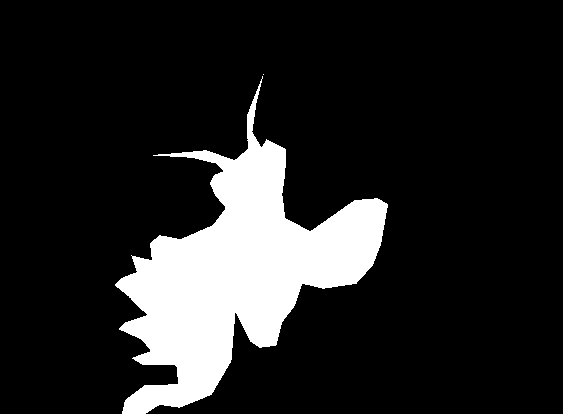}
    \includegraphics[width=0.1\textwidth]{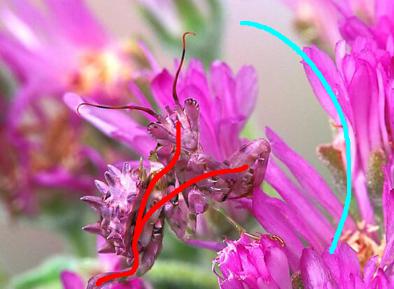}
    \hspace{0.01\textwidth}
    \includegraphics[width=0.1\textwidth]{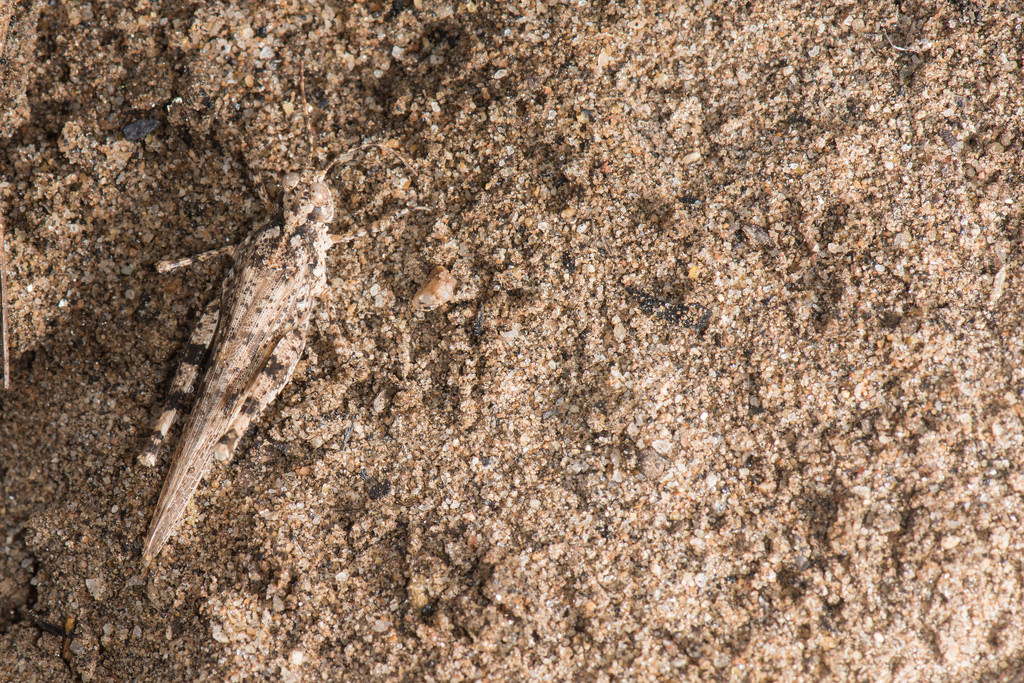}
    \includegraphics[width=0.1\textwidth]{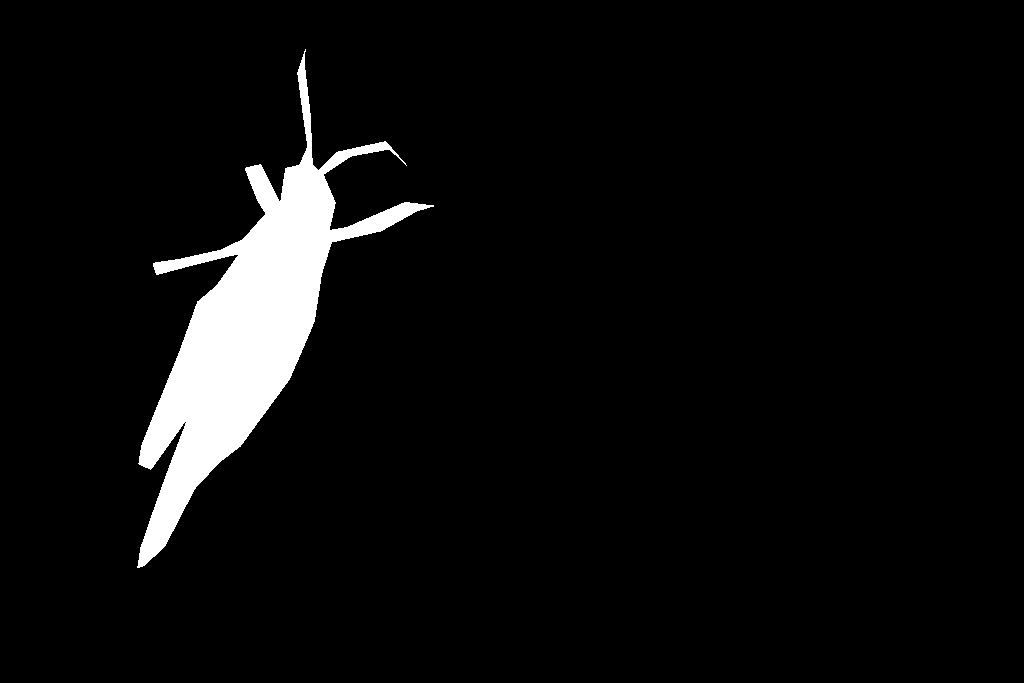}
    \includegraphics[width=0.1\textwidth]{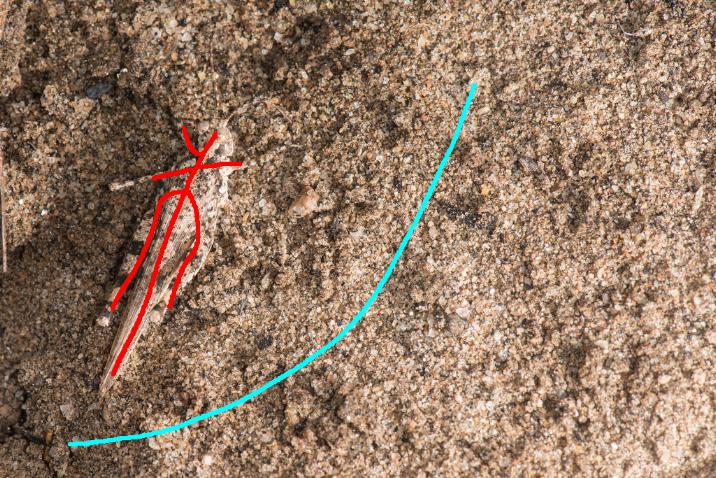}
    \hspace{0.01\textwidth}
    \includegraphics[width=0.1\textwidth]{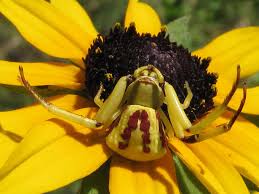}
    \includegraphics[width=0.1\textwidth]{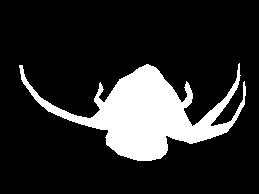}
    \includegraphics[width=0.1\textwidth]{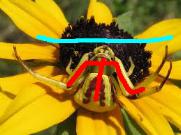}
    \hspace{0.01\textwidth}
    \\
    
    \includegraphics[width=0.1\textwidth]{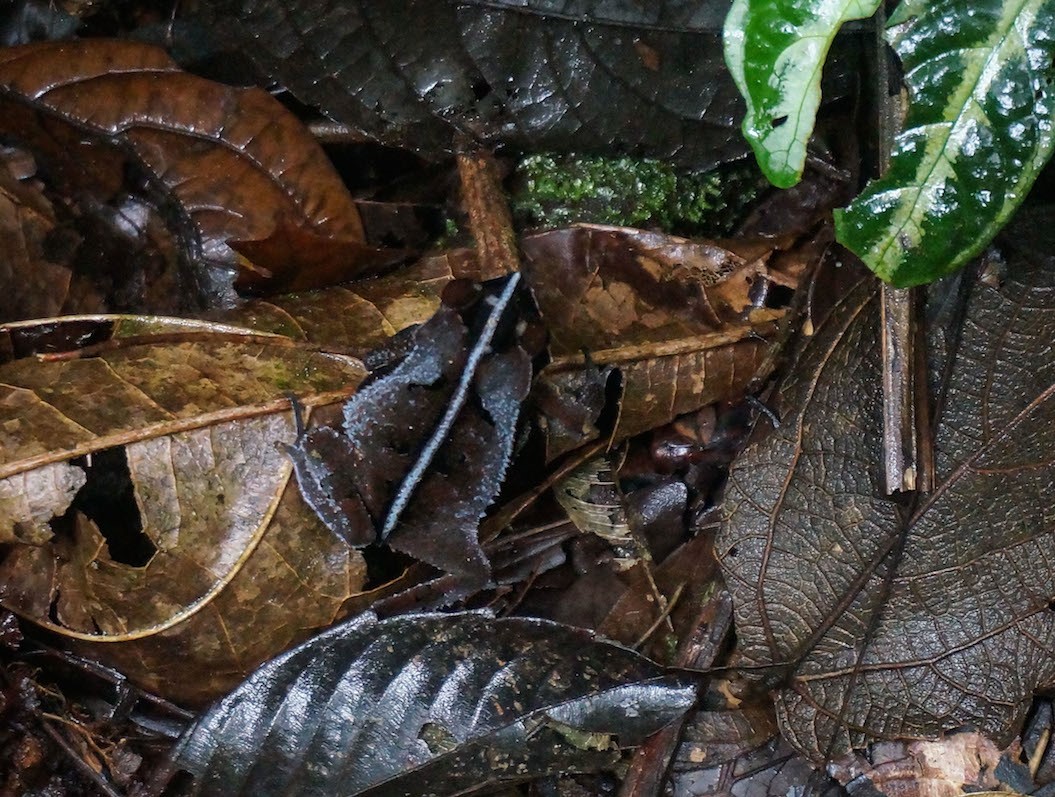}
    \includegraphics[width=0.1\textwidth]{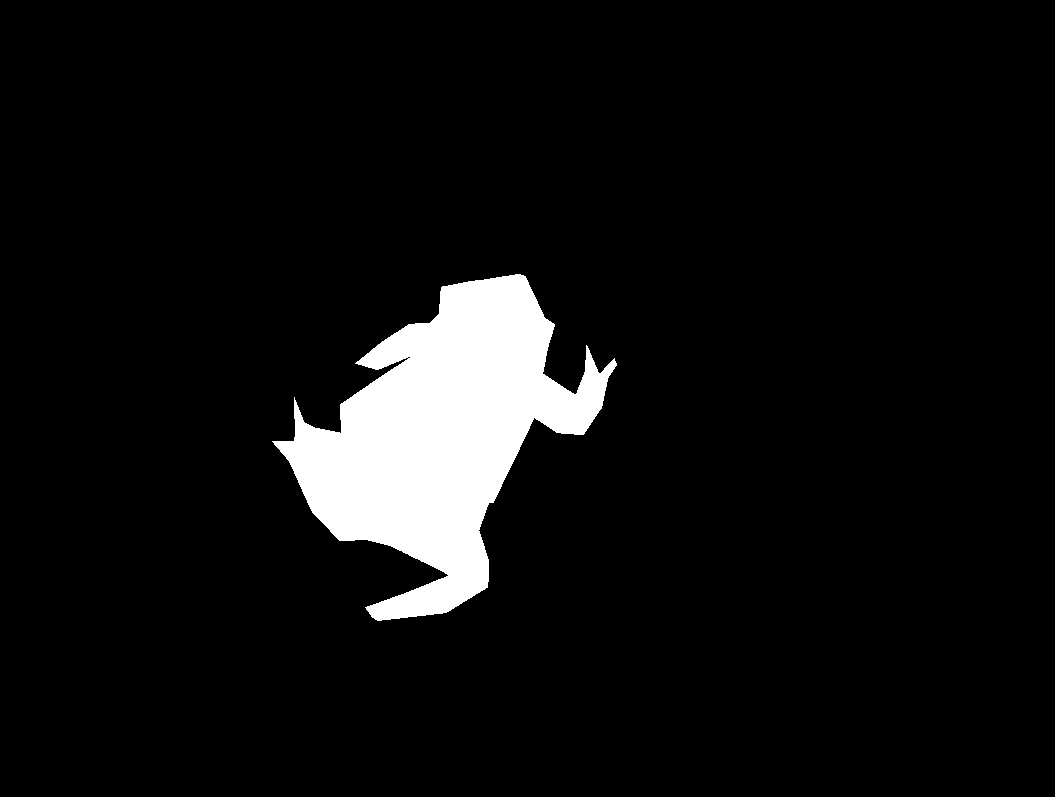}
    \includegraphics[width=0.1\textwidth]{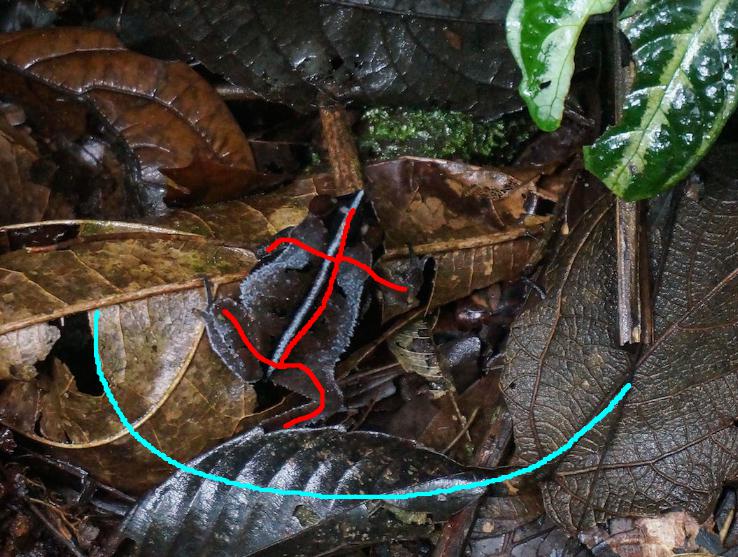}
    \hspace{0.01\textwidth}
    \includegraphics[width=0.1\textwidth]{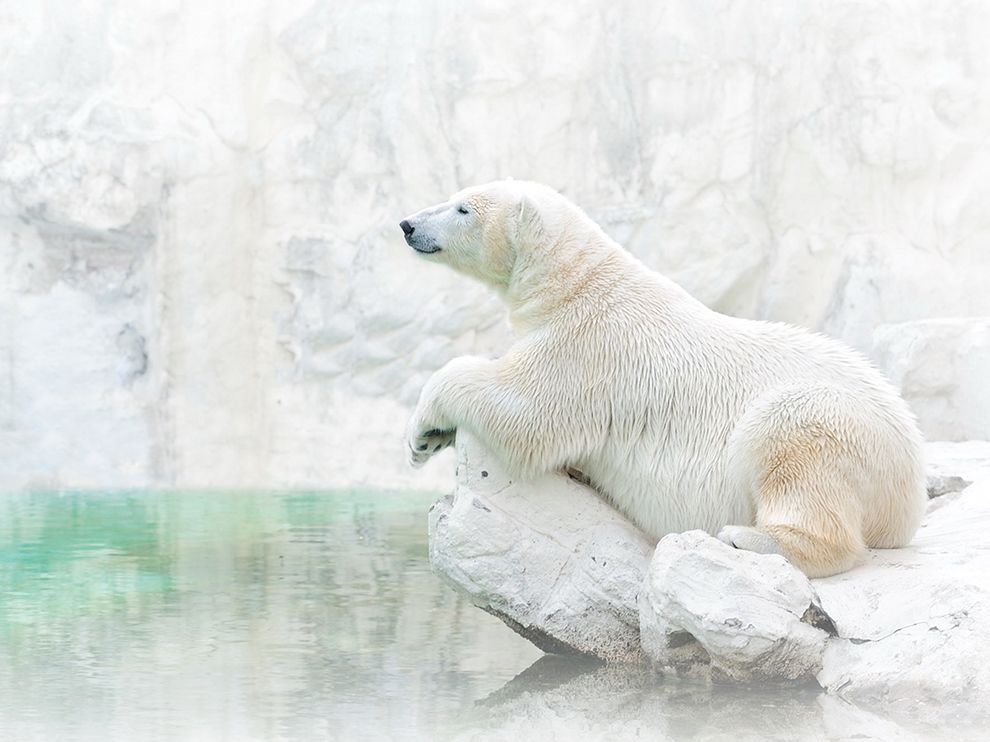}
    \includegraphics[width=0.1\textwidth]{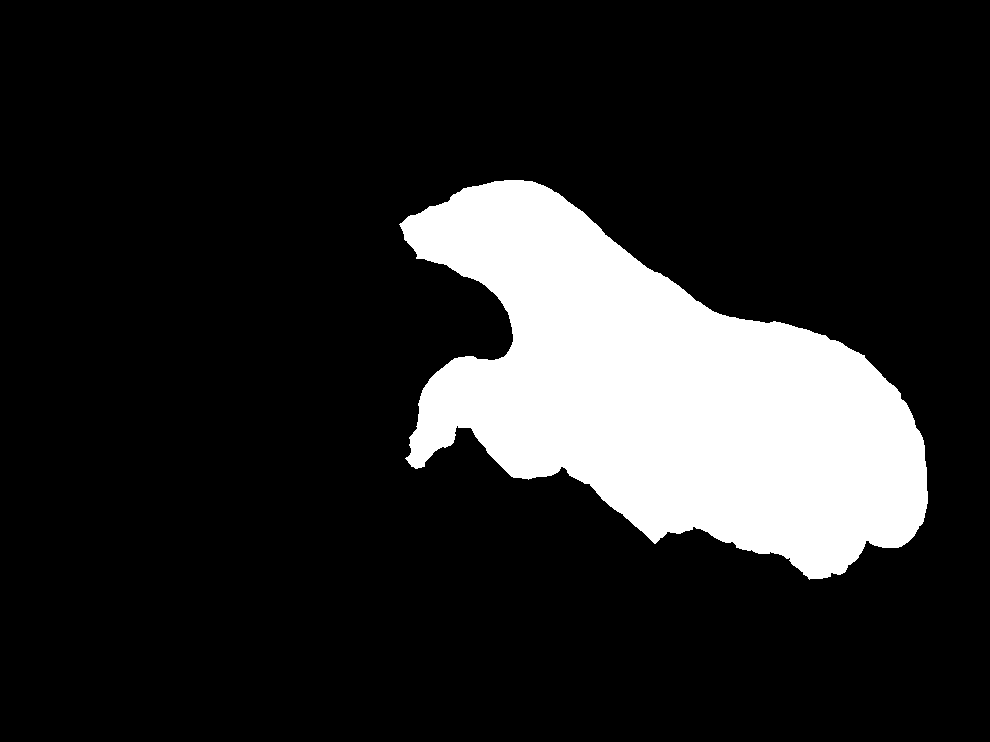}
    \includegraphics[width=0.1\textwidth]{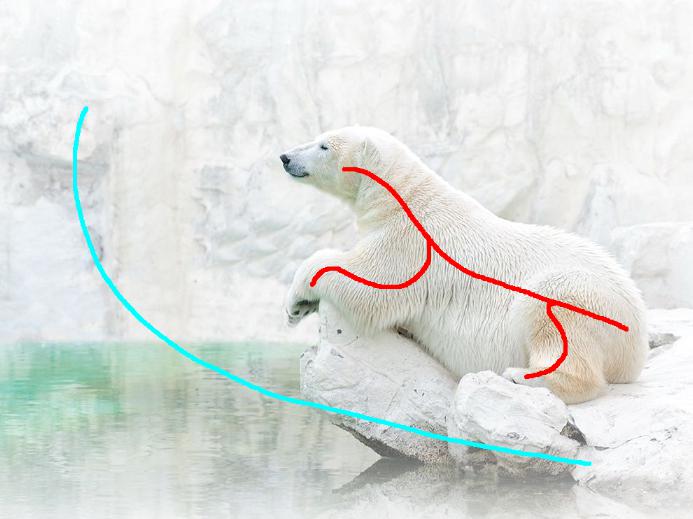}
    \hspace{0.01\textwidth}
    \includegraphics[width=0.1\textwidth]{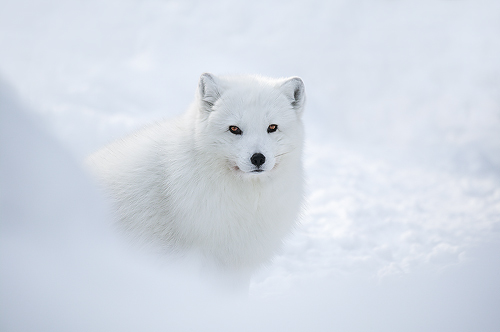}
    \includegraphics[width=0.1\textwidth]{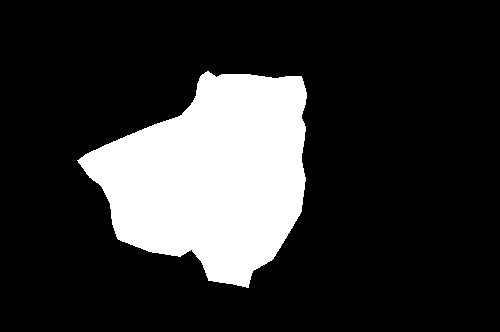}
    \includegraphics[width=0.1\textwidth]{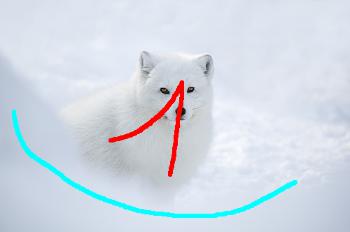}
    \\
    
    \includegraphics[width=0.1\textwidth]{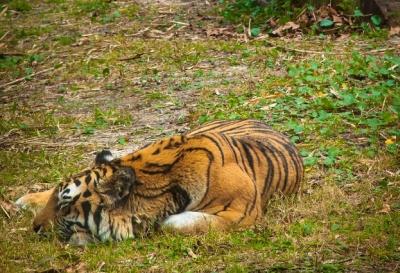}
    \includegraphics[width=0.1\textwidth]{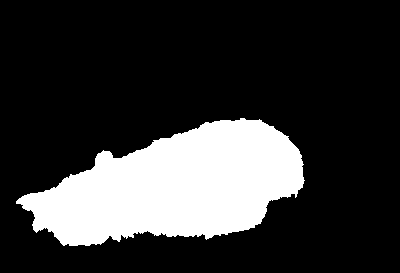}
    \includegraphics[width=0.1\textwidth]{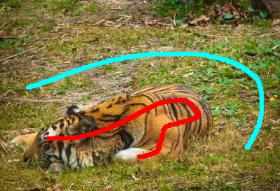}    
    \hspace{0.01\textwidth}
    \includegraphics[width=0.1\textwidth]{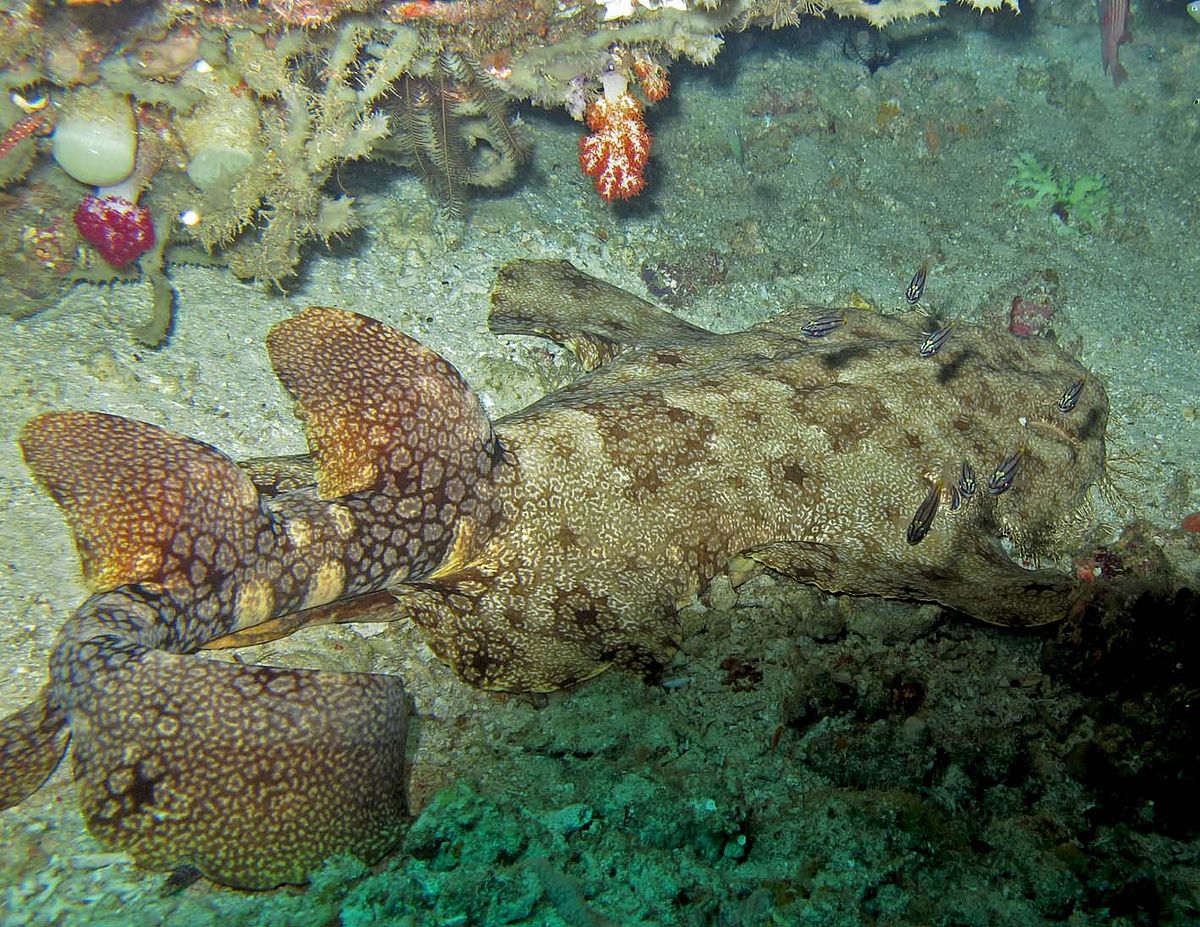}
    \includegraphics[width=0.1\textwidth]{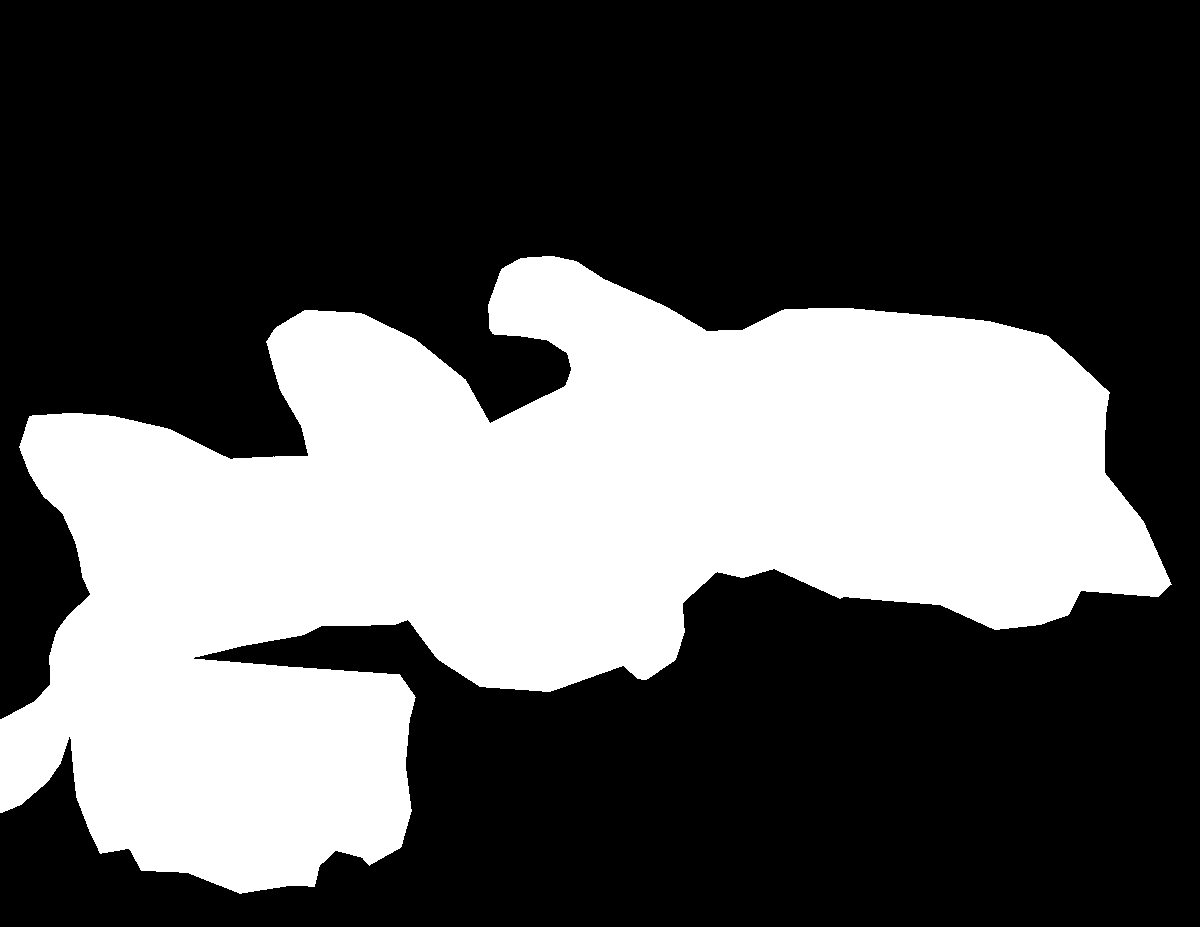}
    \includegraphics[width=0.1\textwidth]{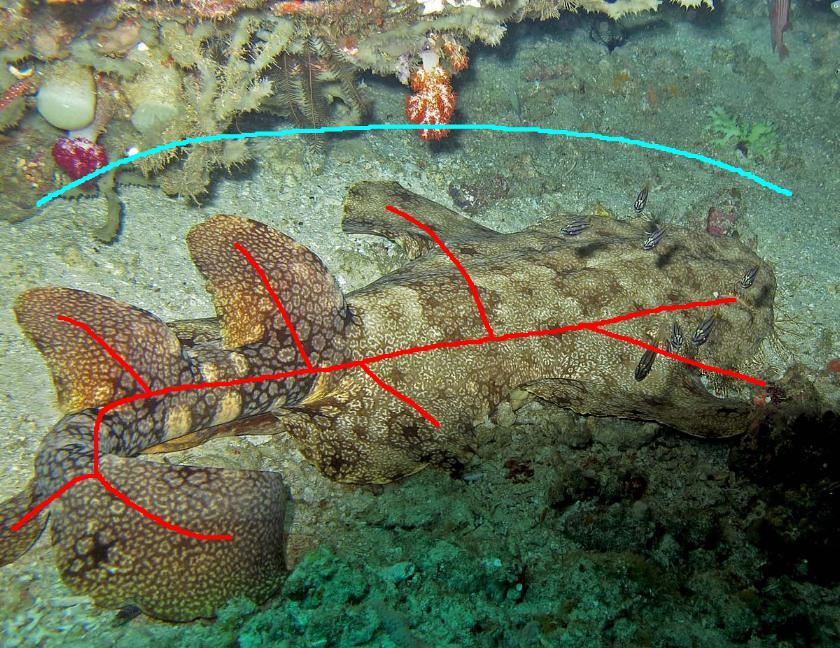}
    \hspace{0.01\textwidth}
    \includegraphics[width=0.1\textwidth]{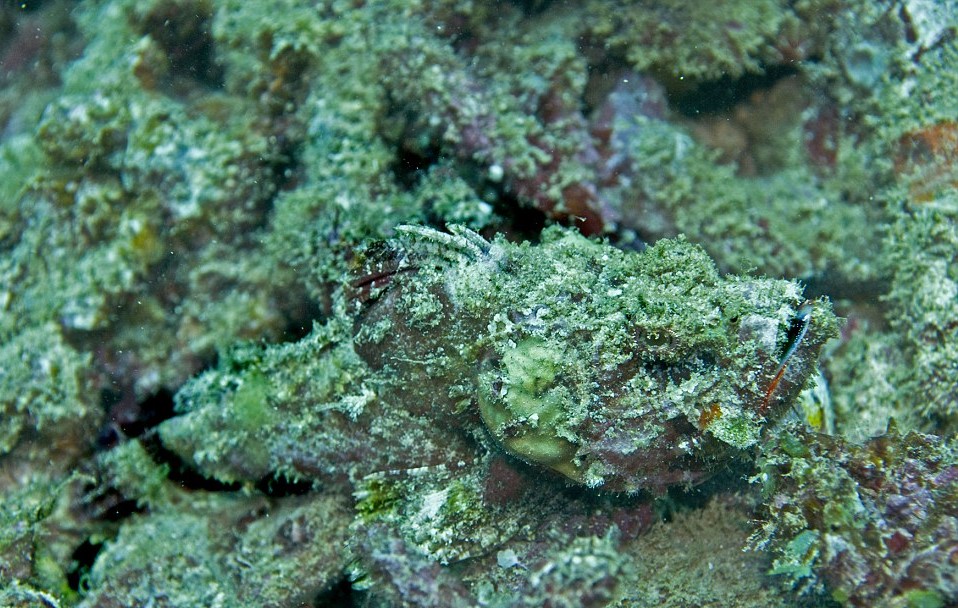}
    \includegraphics[width=0.1\textwidth]{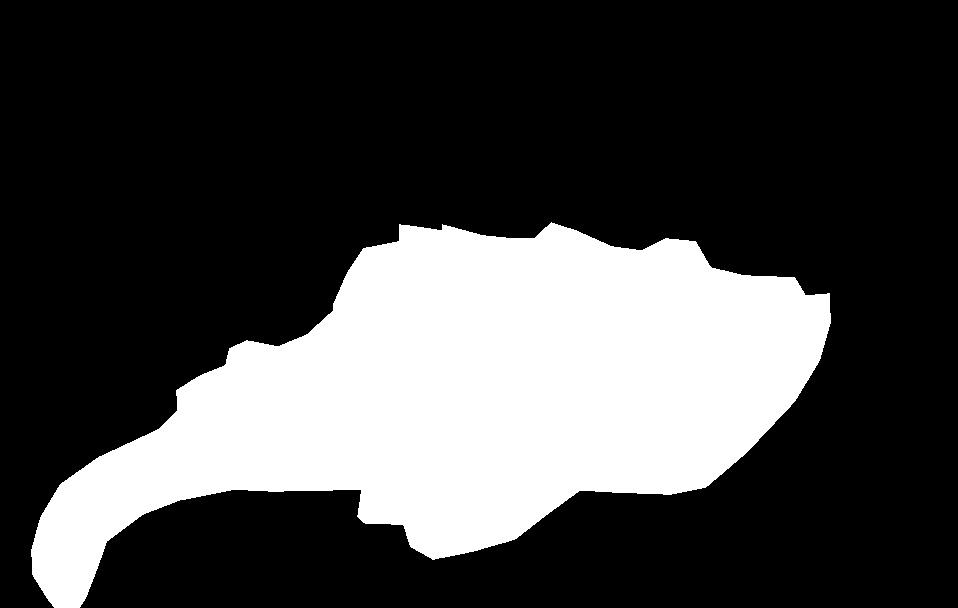}
    \includegraphics[width=0.1\textwidth]{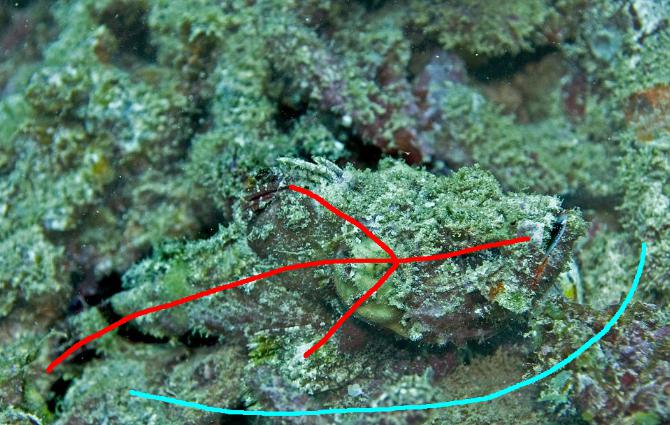}
    \hspace{0.01\textwidth}
    \\
    
    \includegraphics[width=0.1\textwidth]{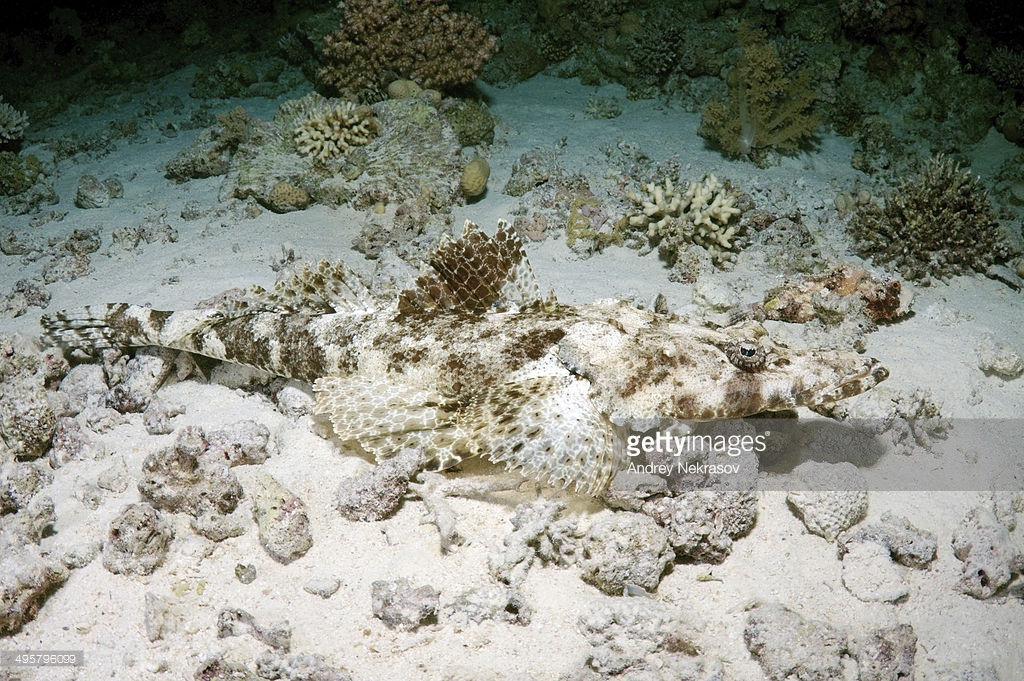}
    \includegraphics[width=0.1\textwidth]{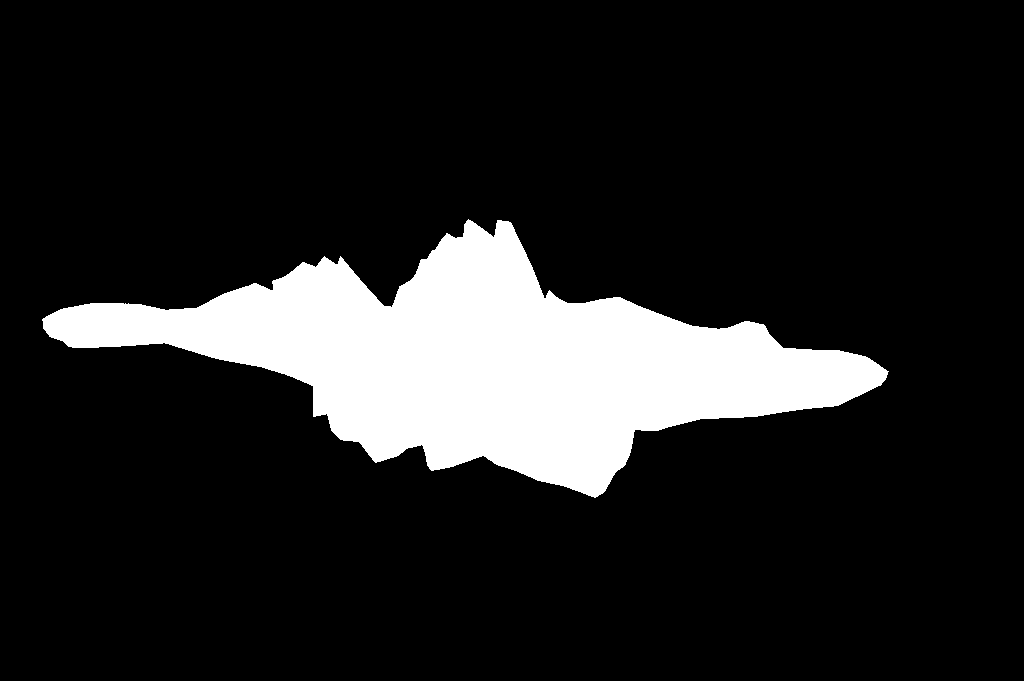}
    \includegraphics[width=0.1\textwidth]{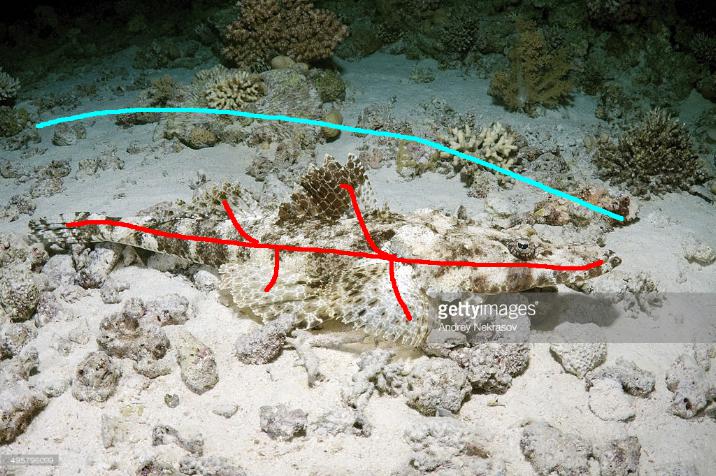}
    \hspace{0.01\textwidth}
    \includegraphics[width=0.1\textwidth]{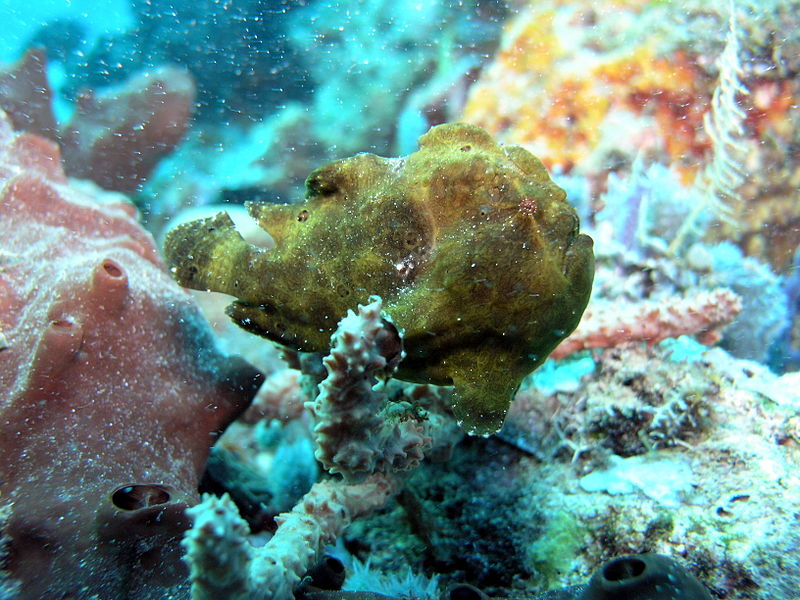}
    \includegraphics[width=0.1\textwidth]{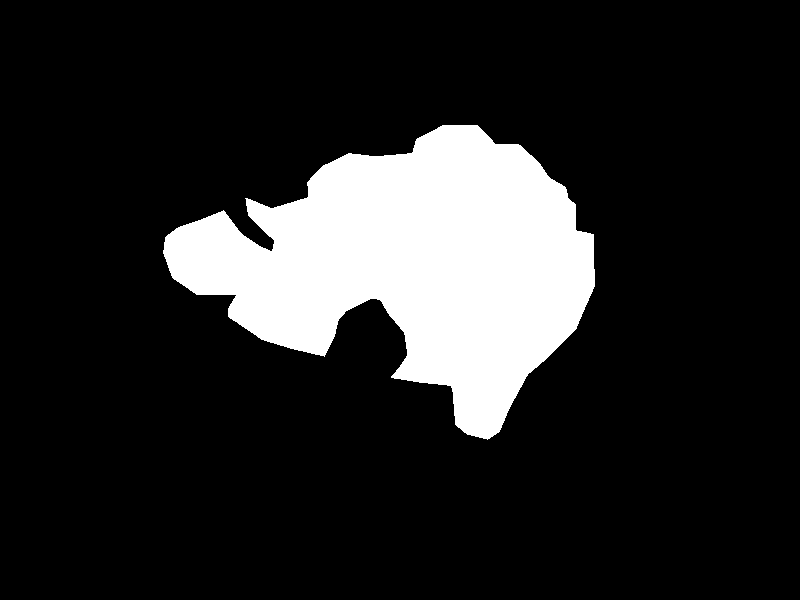}
    \includegraphics[width=0.1\textwidth]{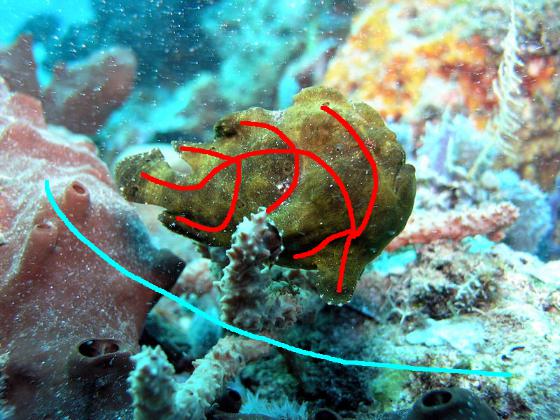}    
    \hspace{0.01\textwidth}
    \includegraphics[width=0.1\textwidth]{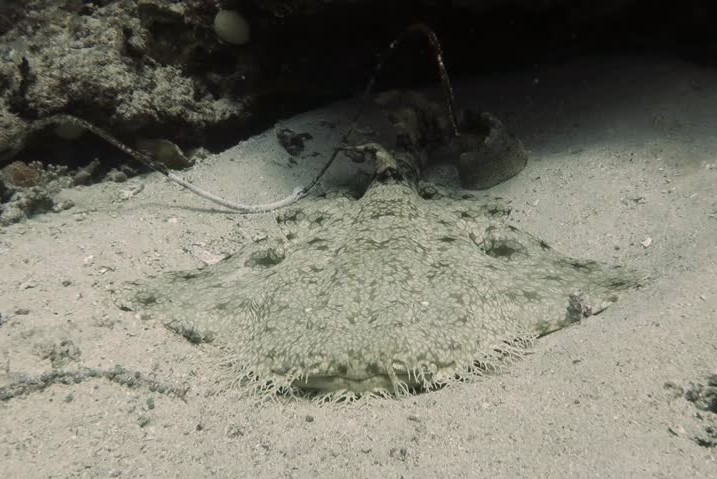}
    \includegraphics[width=0.1\textwidth]{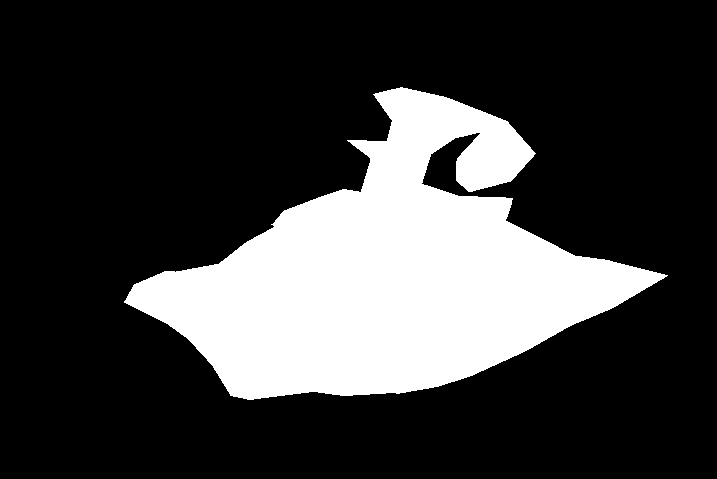}
    \includegraphics[width=0.1\textwidth]{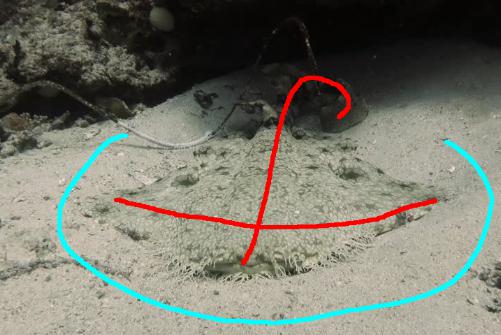}
    \\
    
    \includegraphics[width=0.1\textwidth]{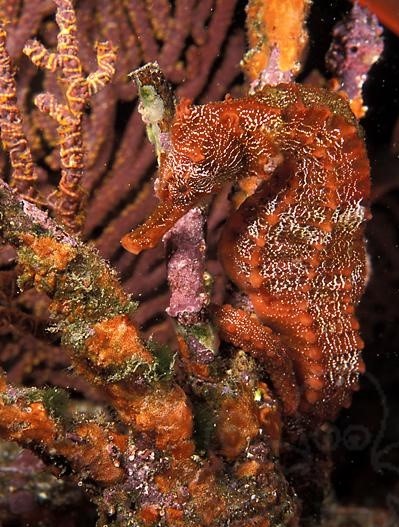}
    \includegraphics[width=0.1\textwidth]{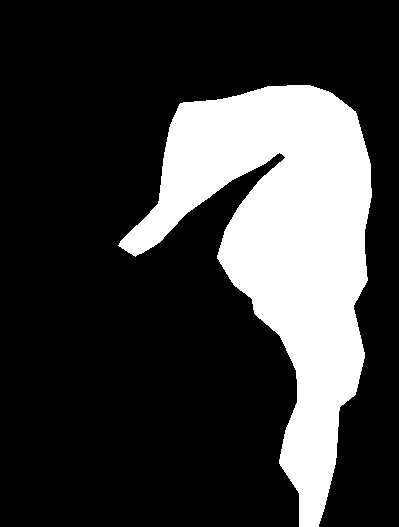}
    \includegraphics[width=0.1\textwidth]{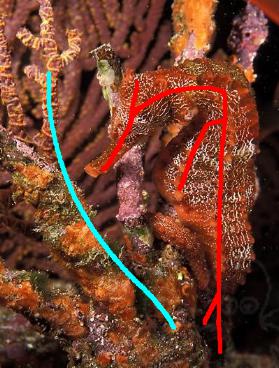}    
    \hspace{0.01\textwidth}
    \includegraphics[width=0.1\textwidth]{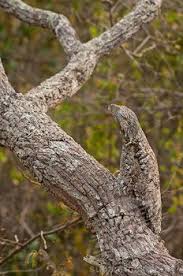}
    \includegraphics[width=0.1\textwidth]{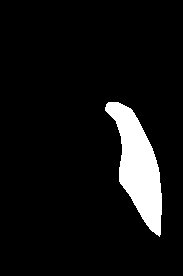}
    \includegraphics[width=0.1\textwidth]{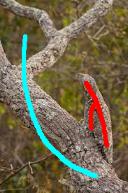}
    \hspace{0.01\textwidth}   
    \includegraphics[width=0.1\textwidth]{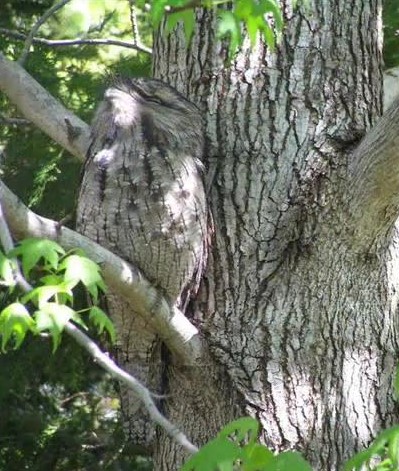}
    \includegraphics[width=0.1\textwidth]{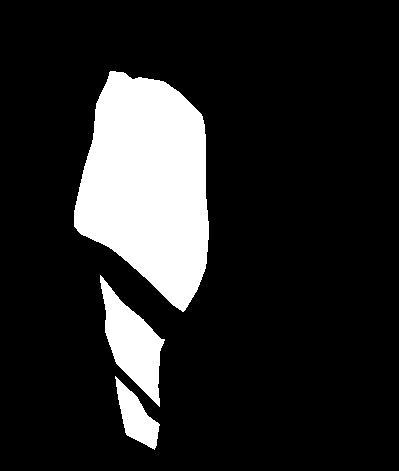}
    \includegraphics[width=0.1\textwidth]{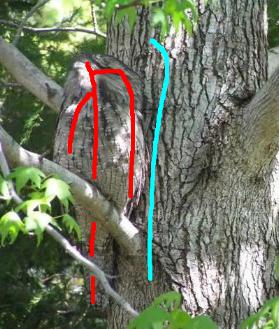} 
    \hspace{0.01\textwidth}
    \\
    
    \includegraphics[width=0.1\textwidth]{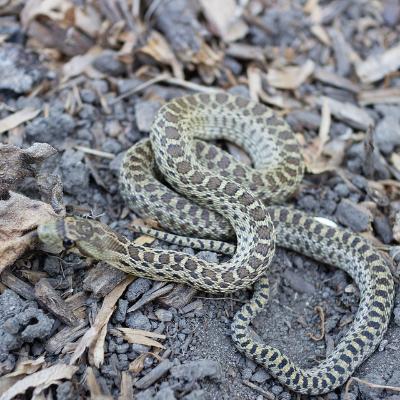}
    \includegraphics[width=0.1\textwidth]{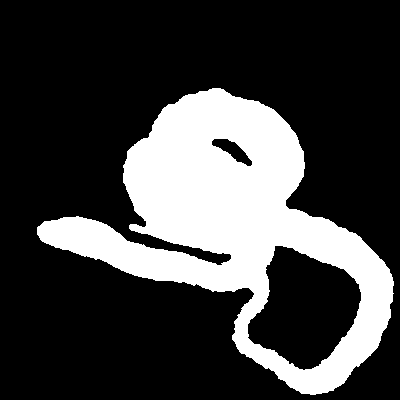}
    \includegraphics[width=0.1\textwidth]{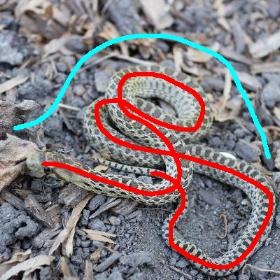}
    \hspace{0.01\textwidth}
    \includegraphics[width=0.1\textwidth]{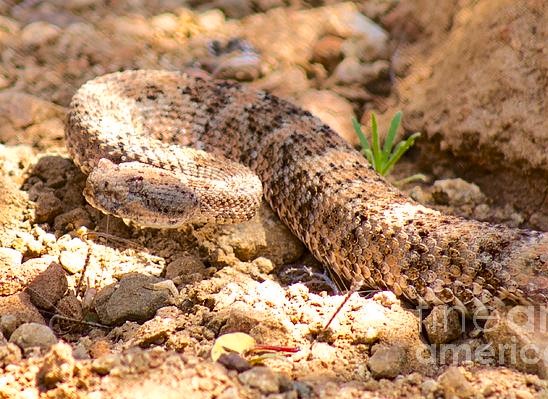}
    \includegraphics[width=0.1\textwidth]{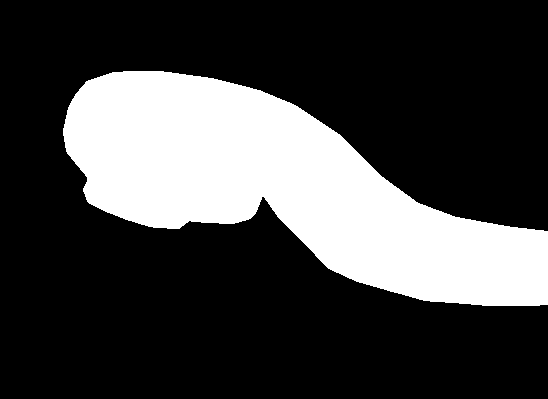}
    \includegraphics[width=0.1\textwidth]{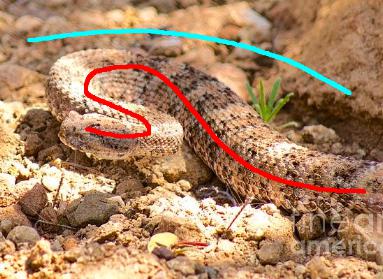}
    \hspace{0.01\textwidth}
    \includegraphics[width=0.1\textwidth]{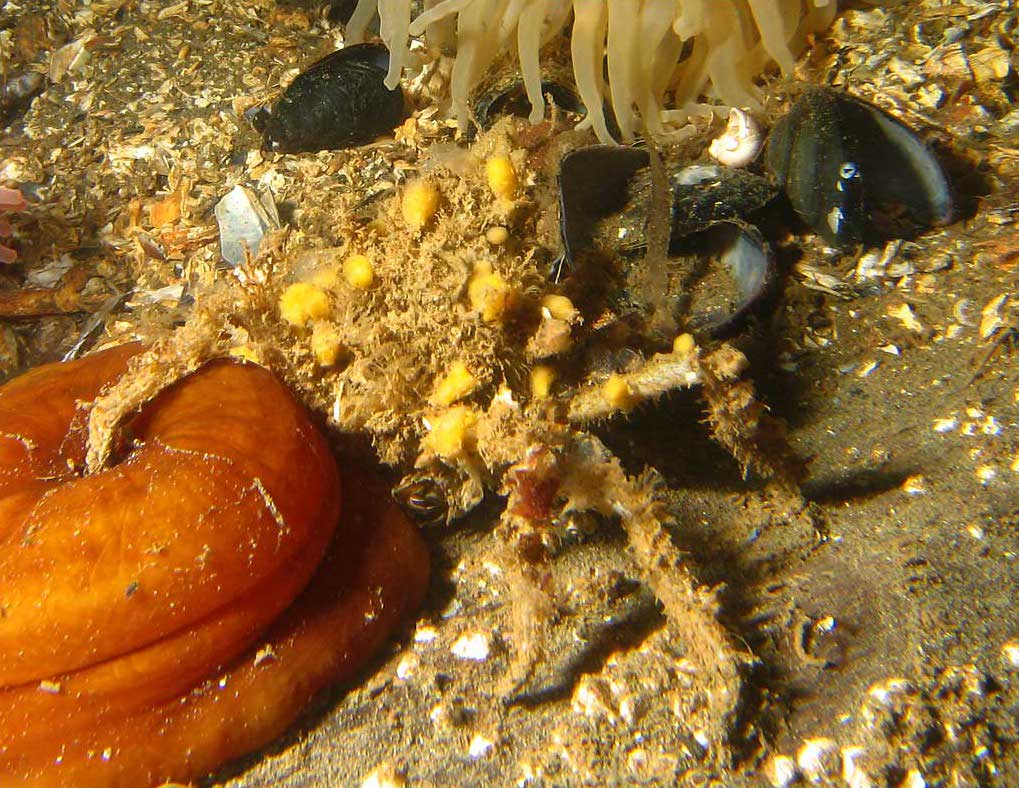}
    \includegraphics[width=0.1\textwidth]{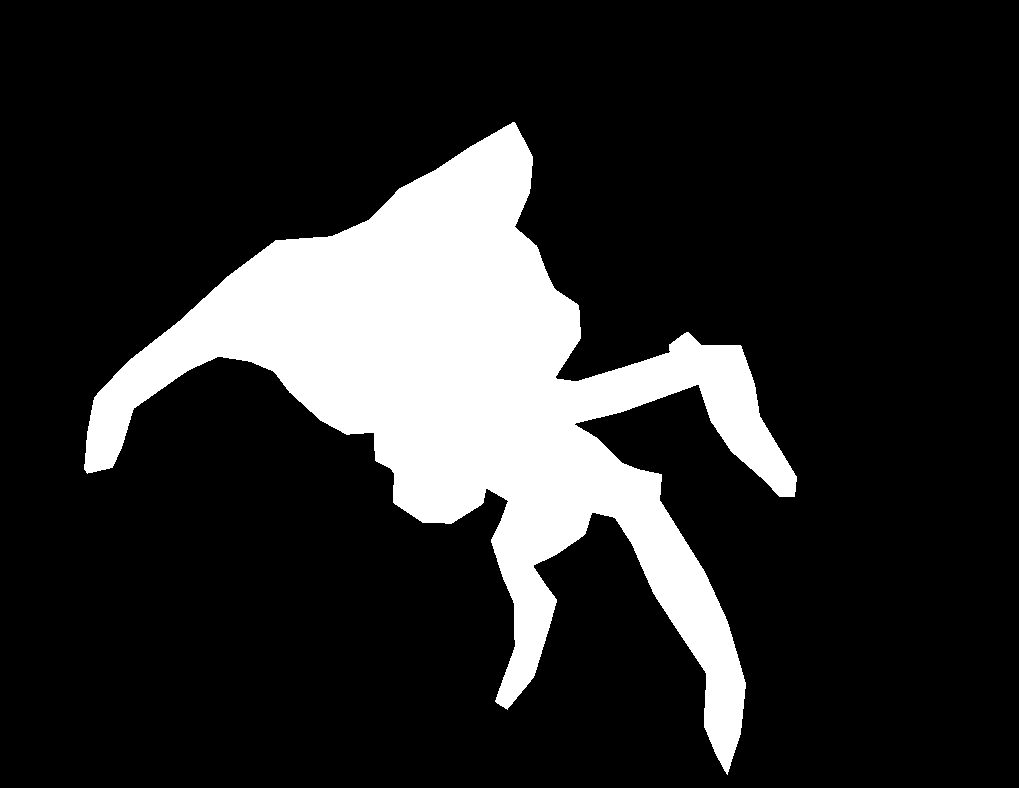}
    \includegraphics[width=0.1\textwidth]{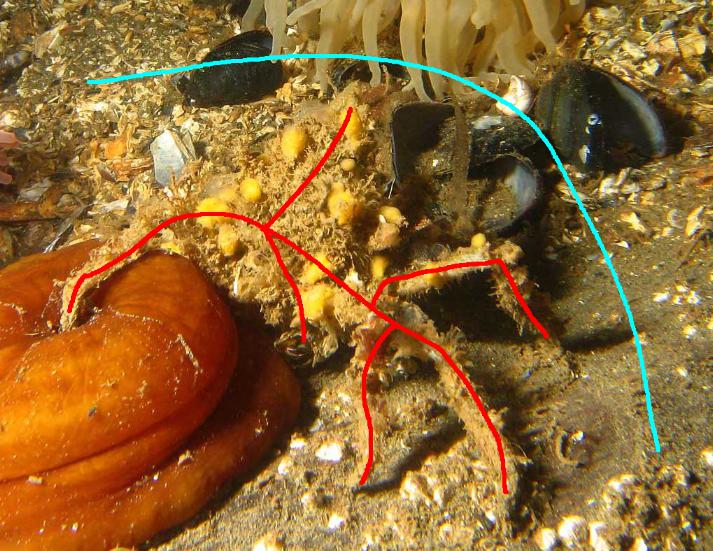}
    \\

    \includegraphics[width=0.1\textwidth]{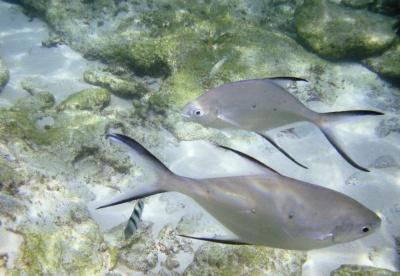}
    \includegraphics[width=0.1\textwidth]{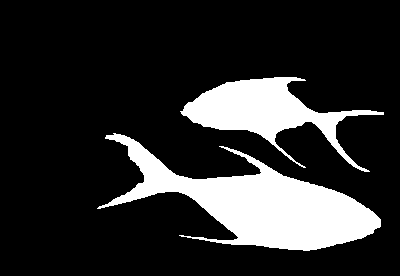}
    \includegraphics[width=0.1\textwidth]{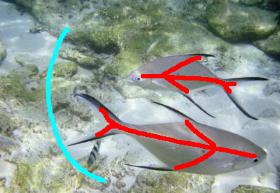}
    \hspace{0.01\textwidth}
    \includegraphics[width=0.1\textwidth]{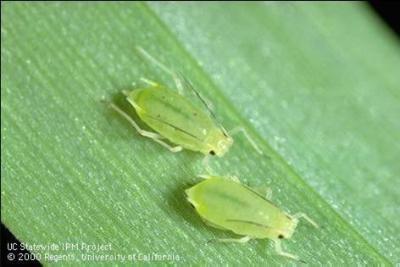}
    \includegraphics[width=0.1\textwidth]{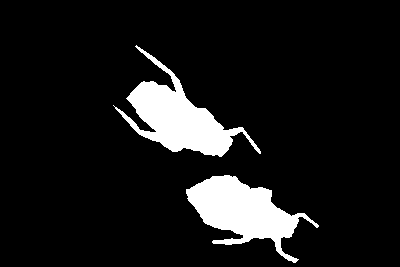}
    \includegraphics[width=0.1\textwidth]{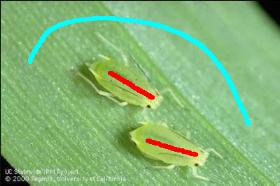}
    \hspace{0.01\textwidth}
    \includegraphics[width=0.1\textwidth]{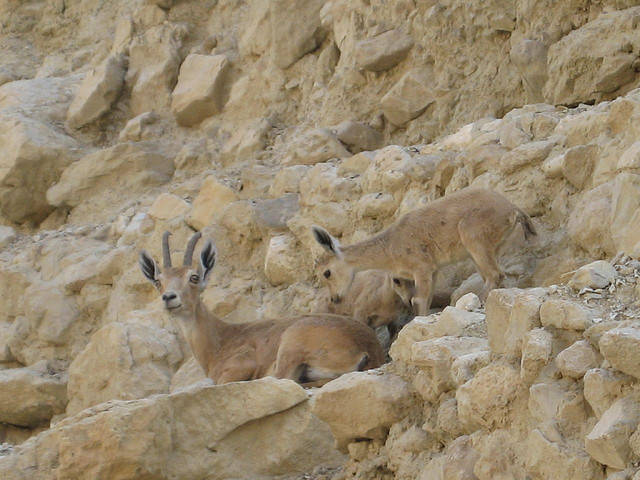}
    \includegraphics[width=0.1\textwidth]{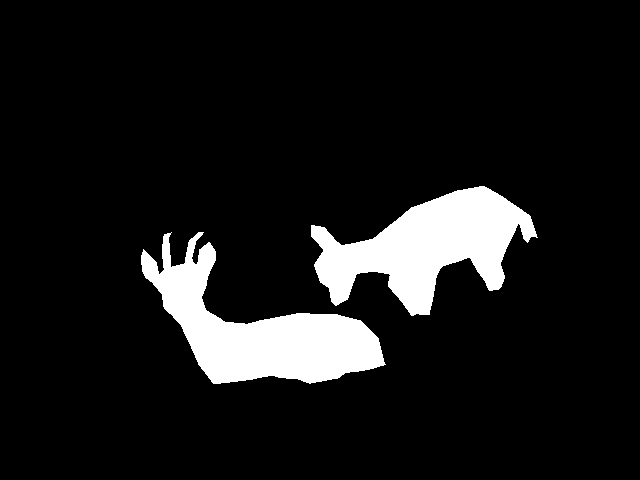}
    \includegraphics[width=0.1\textwidth]{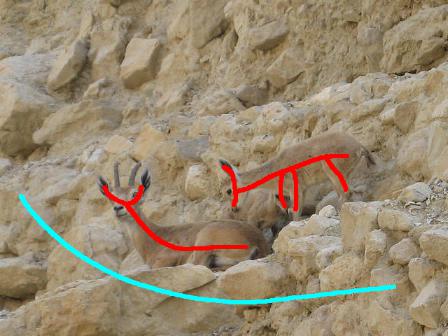}
    \hspace{0.01\textwidth}
    \\
    \includegraphics[width=0.1\textwidth]{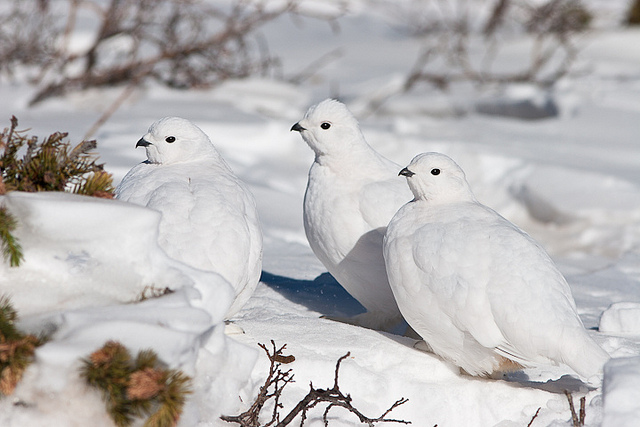}
    \includegraphics[width=0.1\textwidth]{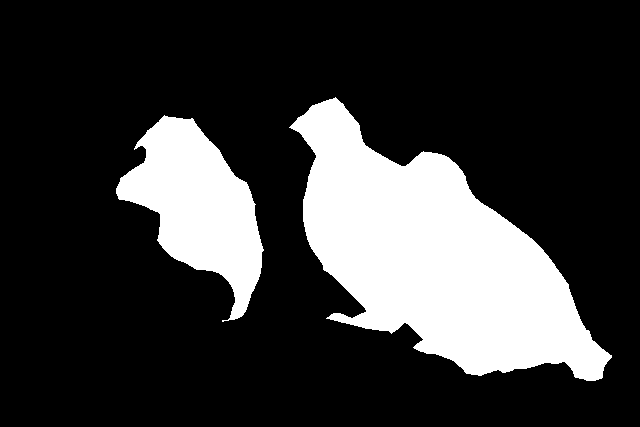}
    \includegraphics[width=0.1\textwidth]{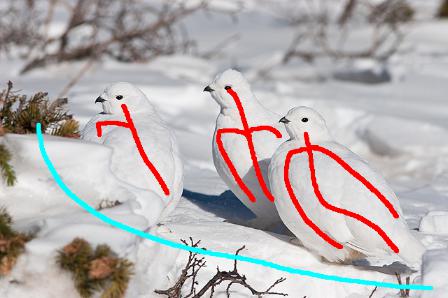}
    \hspace{0.01\textwidth}
    \includegraphics[width=0.1\textwidth]{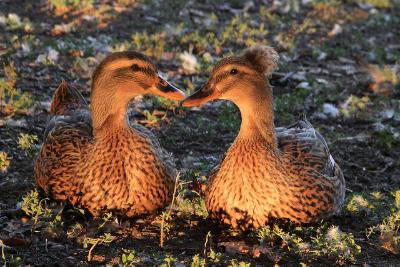}
    \includegraphics[width=0.1\textwidth]{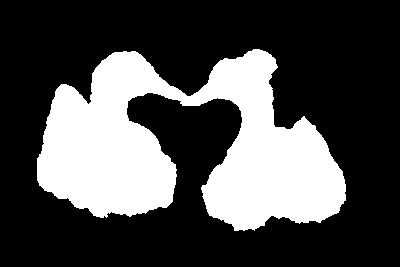}
    \includegraphics[width=0.1\textwidth]{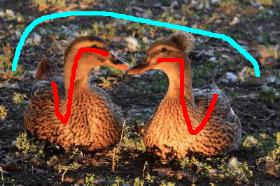}
    \hspace{0.01\textwidth}
    \includegraphics[width=0.1\textwidth]{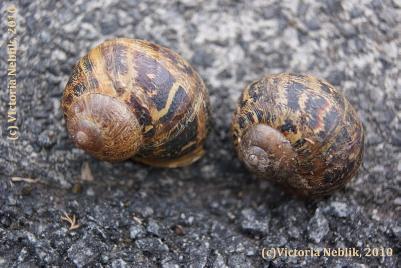}
    \includegraphics[width=0.1\textwidth]{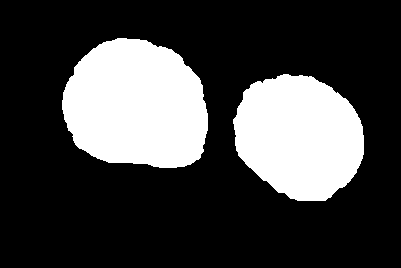}
    \includegraphics[width=0.1\textwidth]{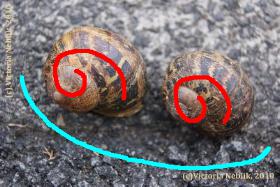}
    \\

    \makebox[0.1\textwidth]{\scriptsize Image}
    \makebox[0.1\textwidth]{\scriptsize GT (pixel-wise)}
    \makebox[0.1\textwidth]{\scriptsize Scribble}
    \hspace{0.01\textwidth}
    \makebox[0.1\textwidth]{\scriptsize Image}
    \makebox[0.1\textwidth]{\scriptsize GT (pixel-wise)}
    \makebox[0.1\textwidth]{\scriptsize Scribble}
    \hspace{0.01\textwidth}
    \makebox[0.1\textwidth]{\scriptsize Image}
    \makebox[0.1\textwidth]{\scriptsize GT (pixel-wise)}
    \makebox[0.1\textwidth]{\scriptsize Scribble}
    \\
    \caption{Examples of S-COD dataset. It includes many categories of animals in challenging scenarios.}
    \label{fig:dataset}
\end{figure*}

The pixel-wise annotation is time-consuming and labor-intensive. Also, it assigns equal significance to every pixel, nor is it the primary structure or the details of foreground objects. To overcome these limitations, we propose a new dataset S-COD with scribble annotations. Annotators are asked to describe objects' primary structures in scribbles according to their first impression without knowing the ground-truth. S-COD includes 4,040 images about diverse objects (\textit{e.g., } Chondrichthyes, Amphibia, Reptilia, Aves, and Mammalia) in challenging scenarios (\textit{e.g., }similar appearance, partial occlusions, and multiple objects). See Figure~\ref{fig:dataset} for examples in our dataset.

\section{Module Details}
\subsection{Local-Context Contrasted (LCC) Module}
The local-context contrasted (LCC) module first reduces $F_{in}$'s first goes through a 1$\times$1 convolutional layer with batch normalization and ReLU, and then takes the obtained $F_{low} \in 64 \times H \times W$ to two low-level contrasted extractors (LCE) focus on different sizes of fields. An LCE consists of a local receptor (LR), a context receptor (CR), and two low-level feature extractors (LFE). The $F_{low}$ is fed into an LR which is a 3$\times$3 convolutional layer with 1 dilation rate and an LFE to obtain the $f_{local}$. Meanwhile, $f_{low}$ is also extracted by a CR which is a 3$\times$3 convolutional layer with dilation $d_{context}$ and further by an LFE for $f_{context}$. We take the subtraction of $f_{local}$ and $f_{context}$ into batch normalization and ReLU to get one level contrasted feature $f_{contrast}$. We set the $d_{context}$ to 4 and 8 for two level of LCE, extracting low-level contrasted features $f_{contrast}^{1}$ and $f_{contrast}^{2}$ concentrating on different sizes of receptive fields. The final output is a concatenation of $f_{contrast}^{1}$ and $f_{contrast}^{2}$.

\begin{align*}
& F_{low} = \mathcal{R}(\mathcal{N}(f_{1}(F_{in},\theta_{1}))) \\
& F_{c}^{0} = \mathcal{R}(\mathcal{N}(LFE(f_{l}^{0}(F_{low}, \theta_{l}^{0}), \alpha_{0}^{0})-\\
&    \quad \quad \quad   LFE(f_{c}^{0}(F_{low},\theta_{c}^{0}), \alpha_{0}^{1}, d_{context}=4))) \\
& F_{c}^{1} = \mathcal{R}(\mathcal{N}(LFE(f_{l}^{1}(F_{low}, \theta_{l}^{1}), \alpha_{1}^{0})-\\
&  \quad \quad \quad     LFE(f_{c}^{1}(F_{low},\theta_{c}^{1}), \alpha_{1}^{1}, d_{context}=8))) \\
& F_{c}^{out} = F_{c}^{0} \copyright F_{c}^{1}
\end{align*}

Inspired by the cross-channel direction and position sensitive strategy proposed by Hou \textit{et al.} \cite{hou2021coordinate}, we adopt a crossing spatial and channel attention mechanism in LFE to extract low-level features (\textit{e.g.,} texture, intensity, color). The input feature $F_{in}$ is firstly taken into a 1 dimensional horizontal and vertical global pooling separately. We use a 1$\times$1 convolutional layer, batch normalization and Swish function to process the concatenation of two pooling results. Then the processed feature $F_{mid}$ is split to $F_{mid}^{h} \in C \times H \times 1$ and $F_{mid}^{w} \in C \times 1 \times W$ with permutation. $F_{mid}^{h}$ and $F_{mid}^{w}$ go through a 1$\times$1 convolutional layer and Sigmoid, before multiplying together with $F_{in}$ as an output.

\begin{align*}
& F_{mid} = Swish(\mathcal{N}(f_{1}(\mathcal{P}_{h}(F_{in}) \copyright  permute(\mathcal{P}_{w}(F_{in})), \theta_{1}))) \\
& F_{mid}^{h}, F_{mid}^{w} = split(F_{mid}) \\
& F_{out} = F_{in} \odot Sigmoid(f_{h}(F_{mid}^{h},\theta_{h})) \odot \\
& \quad \quad \quad Sigmoid(f_{w}(permute(F_{mid}^{w}),\theta_{w}))
\end{align*}

\subsection{Logical Semantic Relation (LSR) Module}
To effectively analyze objects' relations, LSR learns logical semantic information in 4 branches. The first branch is a single 1$\times$1 convolutional layer. The second branch includes the sequence of a 1$\times$1 convolutional layer, a 7$\times$7 convolutional layer with 3 paddings, and a 3$\times$3 convolutional layer with 7 padding, 7 dilation rate. The last two branches consist of a 1$\times$1 convolutional layer, two 7$\times$7 convolutional layers with 3 paddings, and a 3$\times$3 convolutional layer with 7 padding, 7 dilation rate. To be clarified, all convolutional layers are followed by a batch normalization. In addition, a GeLU is used after all convolutional layers, except for the last layers in each branch. We then use a 3$\times$3 convolutional layer for concatenated outputs of four branches, and add it to the input feature after a 1$\times$1 convolution. The final output is processed further by a GeLU function.

\begin{align*}
& F_{b}^{0} = \mathcal{N}(f_{0}^{0}(F_{in}, \theta_{0}^{0})) \\
& F_{b}^{1} = \mathcal{N}(f_{1}^{2}(\mathcal{G}(\mathcal{N}(f_{1}^{1}(\mathcal{G} (\mathcal{N} (f_{1}^{0}(F_{in}, \theta_{1}^{0}))), \theta_{1}^{1}))),\theta_{1}^{2})) \\
& F_{b}^{2} = \mathcal{N}(f_{2}^{3}(\mathcal{G}(\mathcal{N}(f_{2}^{2}(\mathcal{G}(\mathcal{N}(f_{2}^{1}(\mathcal{G}(\mathcal{N}(f_{2}^{0}(F_{in}, \theta_{2}^{0}))),\theta_{2}^{1}))),\theta_{2}^{2}))),\theta_{2}^{3})) \\
& F_{b}^{3} = \mathcal{N}(f_{3}^{3}(\mathcal{G}(\mathcal{N}(f_{3}^{2}(\mathcal{G}(\mathcal{N}(f_{3}^{1}(\mathcal{G}(\mathcal{N}(f_{3}^{0}(F_{in}, \theta_{3}^{0}))),\theta_{3}^{1}))),\theta_{3}^{2}))),\theta_{3}^{3})) \\
& F_{logic}^{mid} = \mathcal{N}(f_{c}(F_{b}^{0} \copyright F_{b}^{1} \copyright F_{b}^{2} \copyright F_{b}^{3}, \theta_{c})) \\
& F_{logic}^{out} = \mathcal{R}(F_{logic}^{mid} \oplus f_{res}(F_{in},\theta_{res}))
\end{align*}

\section{Limitations and Future Work}
\begin{figure}[t] \centering
    \includegraphics[width=0.1\textwidth]{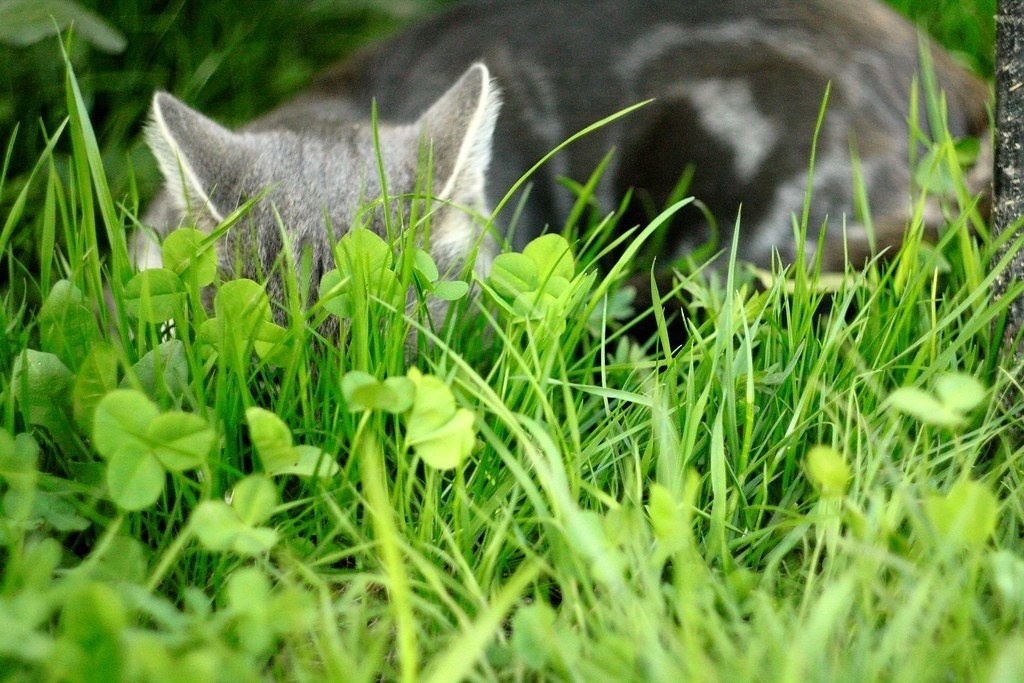}
    \includegraphics[width=0.1\textwidth]{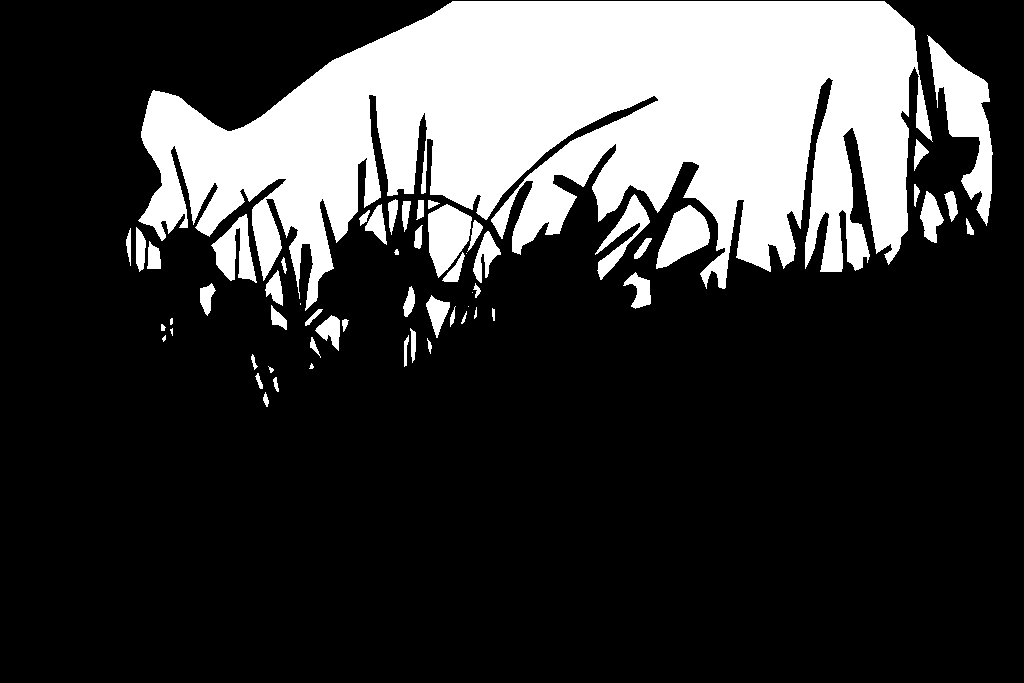}
    \includegraphics[width=0.1\textwidth]{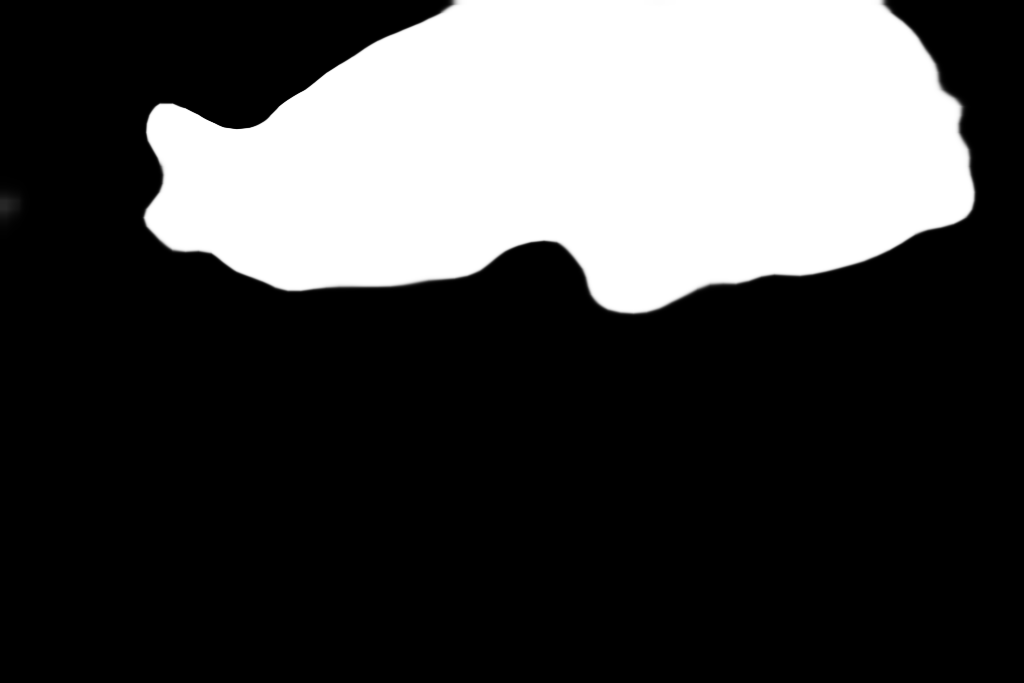}
    \\
    
    \includegraphics[width=0.1\textwidth]{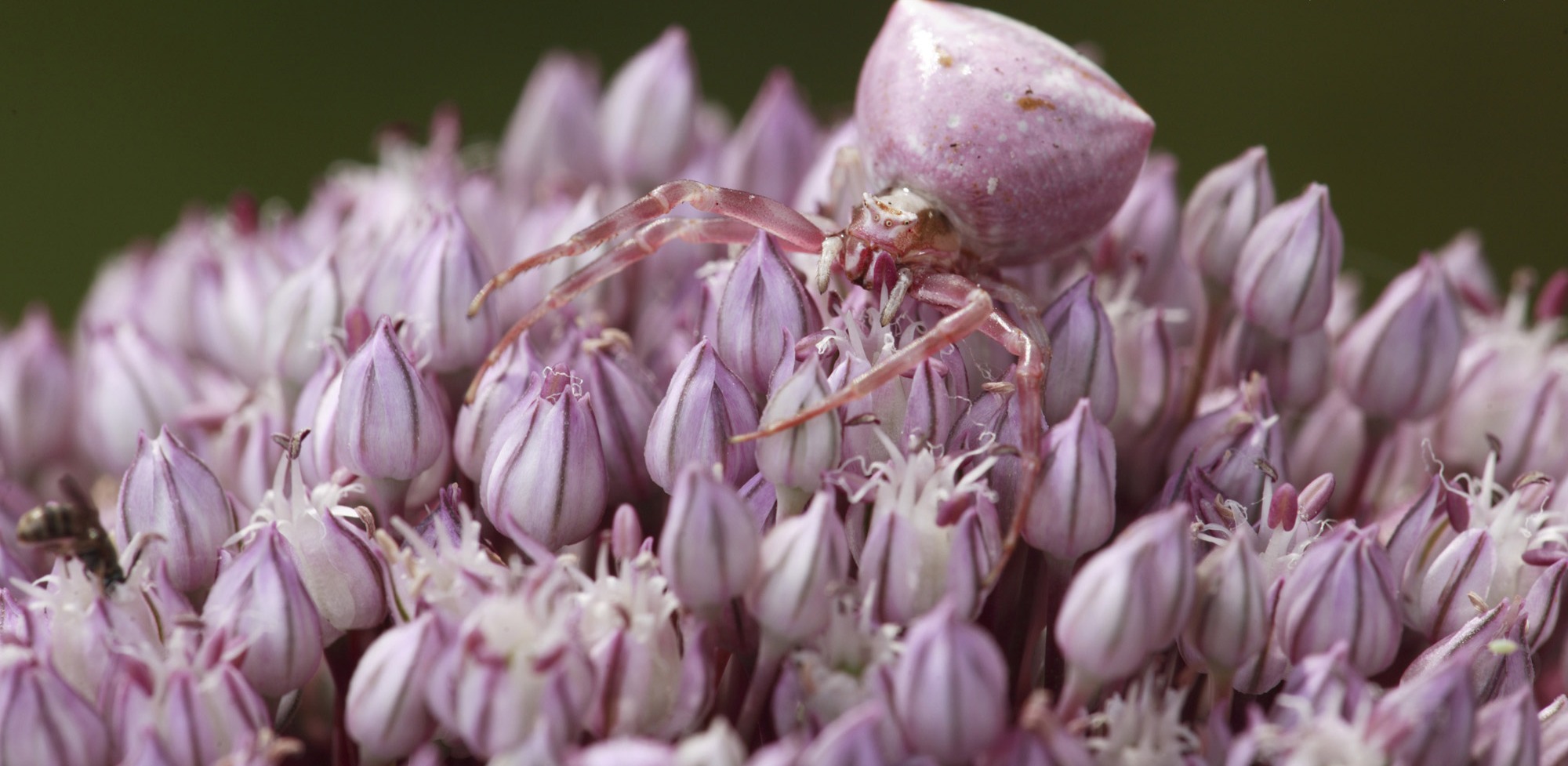}
    \includegraphics[width=0.1\textwidth]{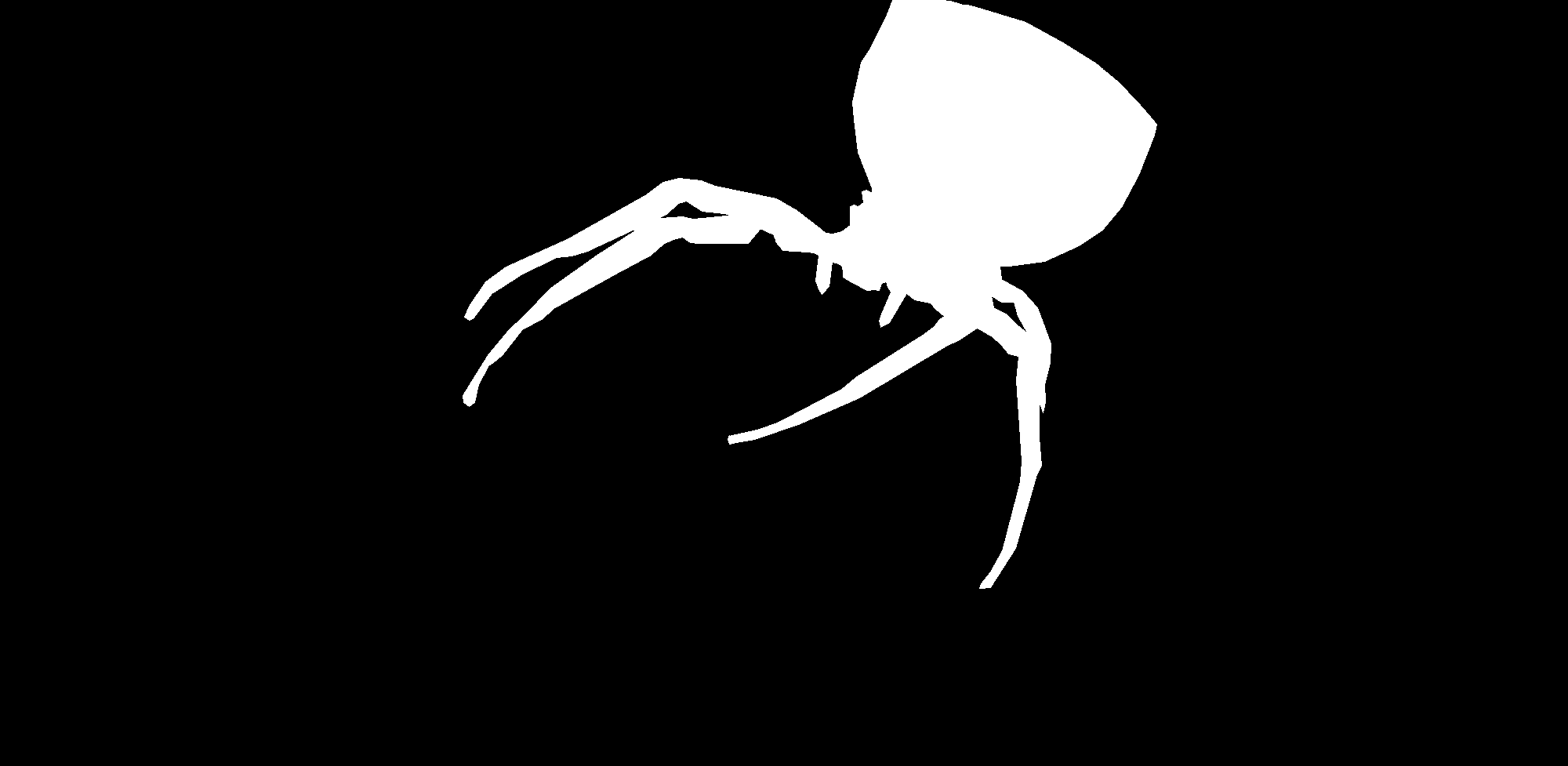}
    \includegraphics[width=0.1\textwidth]{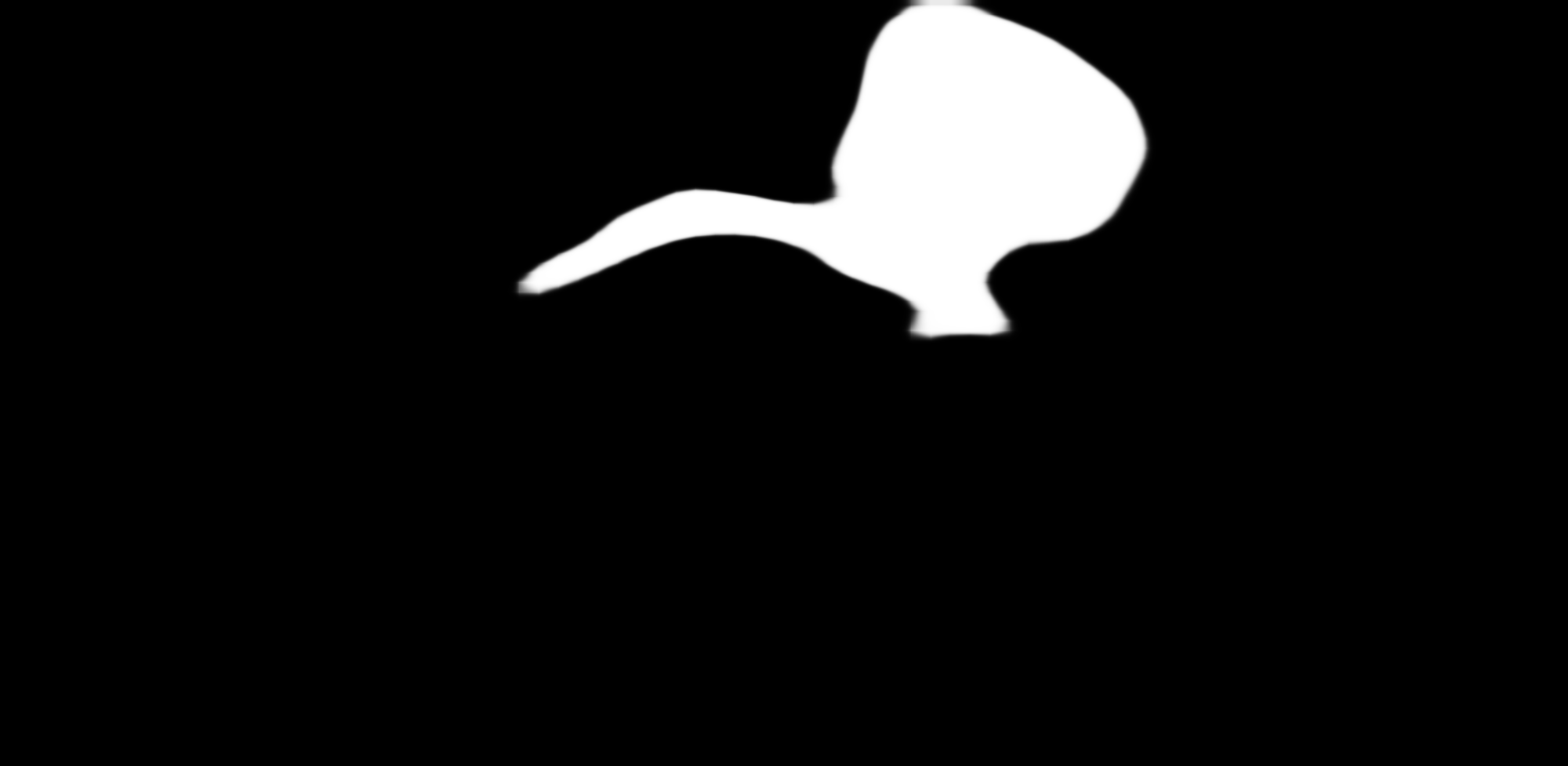}
    \\

    \makebox[0.1\textwidth]{\scriptsize Input}
    \makebox[0.1\textwidth]{\scriptsize GT}
    \makebox[0.1\textwidth]{\scriptsize Ours}
    \\
    \caption{Two examples of our method's failures that inaccurately predict complex boundaries.}
    \label{fig:limitation}
\end{figure}

Our method is effective, but it has limitations. Due to scribble annotations in objects' main bodies, it performs under expectations in scenarios with intricate details in boundaries (\textit{e.g., }tiny intensive occlusions, slender tentacles). Figure \ref{fig:limitation} shows two examples of our failures. The first row is a cat occluded by tiny intensive grasses. Although our method successfully predicts the cat's primary structure, it does not segment grasses. The second row is an animal with slender tentacles. Although our method predicts its main body and partial tentacles, which are closed to the body, it cannot detect entire tentacles. In the future, we will pay more attention to local details, especially intricate boundaries, to segment more precise camouflaged objects under weakly-supervised learning.

\bibliography{egbib}